\documentclass[aoas]{imsart}

\RequirePackage{amsthm,amsmath,amsfonts,amssymb}
\RequirePackage[authoryear]{natbib}
\RequirePackage[colorlinks,citecolor=blue,urlcolor=blue]{hyperref}%% uncomment this for coloring bibliography citations and linked URLs
\RequirePackage{graphicx}%% uncomment this for including figures

\usepackage[english]{babel}
\usepackage{mathtools}
\usepackage{bm}
\usepackage[table]{xcolor}
\usepackage{color}
\usepackage{graphicx}
\usepackage{datetime}
\usepackage{booktabs}
\usepackage{tabularx}
\usepackage{makecell}
\usepackage{float}
\usepackage{xspace}
\usepackage{enumitem}
\usepackage{soul}
\usepackage{rotating}
\usepackage{subfig}
\usepackage{adjustbox}
\usepackage{multirow}
\usepackage{algorithm}
\usepackage{algpseudocode}
\usepackage{pdfpages}

\startlocaldefs
\theoremstyle{plain}
\newtheorem{theorem}{Theorem}

\newtheorem{proposition}[theorem]{Proposition}
\theoremstyle{remark}

\newtheorem{remark}{Remark}
\newcommand{\pkg}[1]{\textsf{#1}\xspace}
\newcommand{\methodset}[1]{\MakeLowercase{\textsc{#1}}\xspace}

\newcommand*{\eg}{e.g.\@\xspace}
\newcommand*{\ie}{i.e.\@\xspace}
\newcommand*{\vs}{vs.\@\xspace}
\makeatletter
\newcommand*{\etc}{%
\@ifnextchar{.}%
{etc}%
{etc.\@\xspace}%
}
\makeatother

\newcommand{\siref}{Supplementary~Material~\citep{smf}}
\newcommand{\sisec}[1]{Supplementary~Material,~Section~S#1~\citep{smf}}

\newcommand{\secref}[1]{Section~\ref{#1}}

\newcommand{\figref}[1]{Figure~\ref{#1}}
\newcommand{\tabref}[1]{Table~\ref{#1}}
\newcommand{\algoref}[1]{Algorithm~\ref{#1}}

\newcommand{\nth}{-\emph{th}\xspace}

\newcommand{\eps}{\varepsilon}

\newcommand{\norm}[1]{\left\lVert #1 \right\rVert}

\newcommand{\ind}{\mathbb{I}}

\newcommand{\set}[1]{\left\{{#1}\right\}}
\newcommand{\setR}{\mathbb{R}}

\newcommand{\inv}[1]{{#1}^{\scriptsize -1}}
\newcommand{\tra}[1]{{#1}^\mathsf{T}}

\DeclareMathOperator*{\argmax}{arg\,max}
\DeclareMathOperator*{\argmin}{arg\,min}

\DeclareMathOperator{\oPr}{Pr}
\newcommand{\prob}[2][]{{\oPr_{#1}}{\left\{#2\right\}}}
\DeclareMathOperator{\oE}{E}
\newcommand{\E}[2][]{{\oE_{#1}}{\left[#2\right]}}

\newcommand{\given}{\,{\,\vert\,}\,}

\newcommand{\iidsim}{\,\overset{\text{iid}}{\scalebox{1.25}{$\sim$}}\,}

\newcommand{\cndGaussian}{{\normalfont (C1)}\xspace}
\newcommand{\cndEqualPrecision}{{\normalfont (C2)}\xspace}
\newcommand{\cndInfDiffLoglik}{{\normalfont (C3)}\xspace}

\newcommand{\bSigma}{\bm{\Sigma}}

\newcommand{\bx}{\bm{x}}
\newcommand{\by}{\bm{y}}
\newcommand{\bmu}{\bm{\mu}}
\newcommand{\btheta}{\bm{\theta}}

\newcommand{\otn}{1,\allowbreak 2, \allowbreak \ldots, \allowbreak n}
\newcommand{\otK}{1,\allowbreak 2, \allowbreak \ldots, \allowbreak K}

\DeclareMathOperator{\qs}{qs}

\DeclareMathOperator{\ent}{ent}

\DeclareMathOperator{\dkl}{d_{KL}}

\newcommand{\bthetam}{\btheta^{(m)}}
\newcommand{\hbtheta}{\hat{\btheta}_{n}}
\newcommand{\hbthetam}{\hat{\btheta}^{(m)}_n}
\newcommand{\pim}{\pi^{(m)}}
\newcommand{\bmum}{\bmu^{(m)}}
\newcommand{\bSigmam}{\bSigma^{(m)}}
\newcommand{\Hd}{H(\cdot)}
\newcommand{\Td}{T(\cdot)}

\endlocaldefs

\usepackage{mdframed}
\newmdenv[
  backgroundcolor=gray!20,
  frametitle=,
  skipabove=\topsep,
  skipbelow=\topsep,
]{reminder}

\begin{document}

\begin{reminder}
{\centering \color{red} \textbf{\textsf{This is a preprint. The revised version of this paper is published as}}\\}
\vspace{5pt}
\noindent Coraggio, L., and Coretto, P. (2023). %
``Selecting the number of clusters, clustering models, and algorithms. A unifying approach based on the quadratic discriminant score''. %
\textit{Journal of Multivariate Analysis}, Vol. 196, p. 105181
(\href{https://doi.org/10.1016/j.jmva.2023.105181}{\sf doi: 10.1016/j.jmva.2023.105181}).
\end{reminder}
{\let\newpage\relax\maketitle}

\begin{frontmatter}
\title{Selecting the number of clusters, clustering models, and algorithms. A unifying approach based on the quadratic discriminant score}
%\title{A sample article title with some additional note\thanksref{T1}}
\runtitle{Selecting the number of clusters, clustering models, and algorithms.}
%\thankstext{T1}{A sample of additional note to the title.}

\begin{aug}
%%%%%%%%%%%%%%%%%%%%%%%%%%%%%%%%%%%%%%%%%%%%%%%
%% Only one address is permitted per author. %%
%% Only division, organization and e-mail is %%
%% included in the address.                  %%
%% Additional information can be included in %%
\author[A]{\fnms{Luca} \snm{Coraggio}\ead[label=e1]{luca.coraggio@unina.it}}
\and
\author[B]{\fnms{Pietro} \snm{Coretto}\ead[label=e2]{pcoretto@unisa.it}}
%%%%%%%%%%%%%%%%%%%%%%%%%%%%%%%%%%%%%%%%%%%%%%
%% Addresses                                %%
%%%%%%%%%%%%%%%%%%%%%%%%%%%%%%%%%%%%%%%%%%%%%%
\address[A]{Department of Economics and Statistics, University of Naples Federico II (Italy), \printead{e1}}

\address[B]{Department of Economics and Statistics, University of Salerno (Italy), \printead{e2}}
\end{aug}

\begin{abstract}
Cluster analysis requires  fixing the number of clusters and often many other hyper-parameters.
In practice, one produces several partitions, and a final one is chosen based on validation or selection criteria.
There exist an abundance of validation methods that, implicitly or explicitly, assume a certain clustering notion.
In this paper, we focus on groups that can be well separated
by quadratic or linear boundaries.
The reference cluster concept is defined through the quadratic discriminant function and parameters describing clusters' size, center and scatter.
We develop two cluster-quality criteria that are consistent with groups generated from a class of elliptic-symmetric distributions. 
Using the bootstrap resampling of the proposed criteria, we propose a selection rule that allows choosing among many clustering solutions, eventually obtained from different methods.
Extensive experimental analysis shows that the proposed methodology achieves a better overall performance compared to established alternatives from the literature.
\end{abstract}


\end{frontmatter}
%%%%%%%%%%%%%%%%%%%%%%%%%%%%%%%%%%%%%%%%%%%%%%
%% Please use \tableofcontents for articles %%
%% with 50 pages and more                   %%
%%%%%%%%%%%%%%%%%%%%%%%%%%%%%%%%%%%%%%%%%%%%%%
%\tableofcontents

%%%%%%%%%%%%%%%%%%%%%%%%%%%%%%%%%%%%%%%%%%%%%%
%%%% Main text entry area:

\section{Introduction}\label{sec:intro}
The typical workflow in cluster analysis is to run one or more algorithms with various settings producing several partitions, among which a researcher needs to choose a final one. There may be multiple partitions that describe the data well according to different clusters' concepts \citep{von2012clustering}. Because of the intrinsic unsupervised nature of the clustering problem, the selection of the desired cluster solution remains a long-standing and open problem \citep{hennig2015clusteringstrategy}. The most significant common issue to all methods and algorithms is choosing an appropriate number of groups, $K$. However, $K$ is not the only relevant decision: many clustering methods and algorithms also require hyper-parameters that control the complexity level at which the data structure is represented.
Similar methods with different hyper-parameters may discover different partitions of a given data set, even at the fixed ``true'' $K$. 
In \sisec{2}, we provide an example on the well-known Iris data set \citep{edgaranderson1936theiris}.
There is a vast catalog of methods proposed to solve the selection problem; for a recent comprehensive overview, see \cite{halkidi2015method}.
Traditionally, in cluster analysis, the selection of the desired partition has been treated as a validation problem rather than a model selection problem.
This is probably because many clustering methods are not derived from stochastic models, although most are built around at least some implicit model assumptions.
Recently, \cite{ullmann2021validation} attempted to categorize different types of validation approaches. Our proposal contributes to the literature on \emph{internal validation methods}, which are methods using the same data used to fit the clusters.
The advantage of internal methods is that they do not require additional information that is sometimes expensive to collect.
Typically, new proposals are advertised claiming their universal ability to discover the data's ``true'' groups.
However, in pure unsupervised contexts, true groups do not exist.
Furthermore, it is often overlooked that each method pursues a specific notion of clusters, which implicitly or explicitly assumes the existence of certain structures in the data. As noted in \cite{akhanli2020comparing}, one needs to choose the validation approach that is consistent with the primary goal of the analysis.
There are \emph{method-dependent} validation methods, specifically designed to evaluate the output of a specific clustering method and \emph{method-independent} methods that can potentially evaluate the output of any clustering methodology.
However, even method-independent validation approaches privilege a certain idea of clusters.

In this paper, we take a different approach: we first define a notion of clusters that different clustering methods may retrieve and then propose a method-independent validation criterion to measure the quality of such clusters.
Specifically, we look for clusters that can be well separated by quadratic boundaries or linear boundaries as a special case.
These clusters are consistent with a class of elliptically-symmetric distributions (ESD), where the within-group dependence structure of the features is mainly driven by correlation.
The quest for clusters of this type is rather common in applications \citep{fraley1998many}.

\paragraph*{Related literature}%
Model-based clustering (MBC) methods, based on ML estimation of finite mixture models of ESDs, are strong candidates for capturing the clusters mentioned above. 
Assuming that each of the $K$ mixture components generates a group, the selection of the desired clustering solution is translated into a model-selection problem where, in practice, the likelihood fit is contrasted with a penalty accounting for model complexity.
The most popular selection strategy is to use information criteria such as the BIC and the AIC \citep{mclachlan2000finite, bouveyron2019modelbased}.

Although information-type criteria are based on a solid theoretical background, there are some issues with their application to cluster selection.
\cite{keribin1998consistent} showed that the BIC is consistent for the number of mixture components under somewhat restrictive assumptions, but practitioners tend to believe that this result is more general, causing some faith in it.
The consistency notion of \cite{keribin1998consistent} is for the recovery of the underlying data distribution and not for clusters.
Paradoxically, these consistency results are problematic for cases where the mixture model is not meant to capture the ``true'' underlying distribution but rather for approximating the density regions formed by the clusters.
Finite Gaussian mixtures can approximate a large class of distributions \citep{nguyen2020approximation}, implying that consistent criteria like the BIC will include additional components inflating $K$ if, for example, a group that is only approximately normal is better fitted by more than one Gaussian component.
The \emph{Integrated Complete-data Likelihood} (ICL) criterion of \cite{biernacki2000assessing} \citep[see also][]{baudry2015estimation} modifies the BIC, adapting it to solve the clustering problem.
Another model-selection approach, based on likelihood-type criteria derived from mixture models, is the cross-validation method proposed in \cite{smyth2000model}.
An additional drawback of information-type indexes is that they are method-dependent: they only allow to compare solutions from MBC methods because their calculation is based on likelihood quantities and models' degrees of freedom.
A further issue is that, in some cases, these criteria can not be calculated for MBC methods when the effective degrees of freedom of the underlying model can not be derived (see the case of ML for Gaussian mixtures with the eigen-ratio regularization treated in Section \ref{sec:empirical_analysis}).

Outside the MBC context, there are many method-independent internal validation indexes that mostly measure the within-cluster homogeneity contrasted to a measure of between-clusters heterogeneity.
Notable examples are the popular CH index of \cite{calinski1974dendrite} and the Average Silhouette Width criteria (ASW) of \cite{rousseeuw1990finding}.
These are not genuinely \emph{model-free} indexes because they purse cluster shapes that depend on the underlying dissimilarity notion.
These indexes have in common with the BIC-type criteria that they implicitly attempt to contrast the cluster fit \vs the increased complexity caused by the increase in $K$.

Another idea from the literature that inspired some aspects of the present contribution is that of \emph{stability selection} \citep{bendavid2006asober}.
The idea taken from this literature is not the notion of stability, which is about finding similar clusterings on similar data sets \citep{hennig2007clusterwise,fang2012selection}, but the idea of exploring variations of the clusterings based on perturbations of the data set obtained by bootstrap resampling of the original data.

\paragraph*{Contribution and organization of the paper}%%
We develop a framework where each cluster is represented by a triplet of parameters representing the notions of size, location and scatter.
This allows to map clusterings obtained with different methods in a form that is consistent with the notion previously discussed. The method-independent nature of our proposal is a major advantage over competitors from the MBC literature.
These clusters' parameters and the quadratic score function, at the heart the Quadratic Discriminant Analysis (QDA), are used to develop two cluster quality criteria called \emph{quadratic scores}.
These criteria are shown to be consistent with clusters generated from a restricted class of ESDs, including the popular Gaussian model (\secref{sec:qs:reference_cluster}).
We show connections between the proposed criteria and likelihood-type quantities related to finite mixtures of ESDs and, in particular, Gaussian mixtures (see \secref{sec:qs:Hn_Sn}).
In the same spirit of the pioneering work of \cite{akaike1973information} on model-selection, we propose to select a clustering solution produced by a method that achieves the largest expected score across all possible partitions of data sets sampled from the data distribution.
The expected score and its confidence interval are approximated via empirical bootstrap in \secref{sec:resampling}.
Finally, in \secref{sec:empirical_analysis}, we propose an extensive numerical analysis where the proposed method is compared against some alternatives on both real and artificial data sets.
Overall, the proposed methodology shows a superior performance and proves to be able to retrieve interesting clustering solutions even in adverse circumstances.
Proofs of the statements are given in Appendix~\ref{sec:proofs}; additional examples and details are given in \siref.

\section{Quadratic scoring}\label{sec:qs}
We fix some general notation used throughout the rest of the paper.
The general clustering problem is to construct a partition $\mathcal{G}_K \allowbreak = \allowbreak \set{\allowbreak G_k,\allowbreak \; k=\otK}$ allocating the objects $\{\otn\}$ into $K$ groups, where $K$ is generally unknown.
Let $\mathbb{X}_n=\set{\bx_i, \; i=\otn}$ be an observed sample of $p$-dimensional feature vectors $\bx_i\in \setR^p$; $\mathbb{X}_n$ is the observed version of a random sample $\mathcal{X}_n=\set{X_i, \; i=\otn}$, where $X_i\in \setR^p$ is the $p$-dimensional random vector of features representing the $i$\nth unit.
In clustering, a typically unsupervised task, we observe the features, but we do not observe the group memberships that we want to discover.
Group memberships are introduced through the random vector of 0-1 variables $Z=\tra{(Z_{1}, \allowbreak Z_{2}, \allowbreak \ldots, \allowbreak Z_K)}$, where $Z_k=1$ denotes membership to the $k$\nth group.
For the $i$-th sample point we define the group memberships as $Z_{ik}=\ind\set{i \in G_k}$, where $\ind\set{\cdot}$ is the usual indicator function.

%% In some cases $\hbtheta$ corresponds to an estimate of a probability model, \eg in the case of MBC algorithms.
%In other situations, $\hbtheta$ maps the output of an algorithm that, although does not pursue the fitting of a population model, can still effectively retrieve the clustered regions produced by populations like those assumed in Proposition~\ref{stm:coherence_qsp}.
%In general we can map a clustering output into a $\hbthetam$ by taking $\hat{\pi}_{n,k}$ equal the fraction of points in the $k$\nth cluster, and $(\hat{\bmu}_{n,k}, \hat{\bSigma}_{n,k})$ equal to the sample mean vector and covariance matrix of points assigned to the $k$\nth cluster.

\subsection{The reference cluster concept}\label{sec:qs:reference_cluster}
Assume that $X \sim F$, where $F$ is the population distribution function producing $K$ clustered regions of points.
We assume that each cluster $k = \otK$ is meaningfully described by the triplet of parameters $\btheta_k = \set{\pi_k, \bmu_k, \bSigma_k}$ formalizing the notions of size, center and scatter.
For the $k$\nth cluster, $\pi_k$ is the expected fraction of points belonging to the $k$\nth group, $\bmu_k \in \setR^p$ is the vector of centers and $\bSigma_k \in \setR^{p\times p}$ is a positive definite scatter matrix that either coincides with or is proportional to the group's covariance matrix.
A cluster configuration $m$ of $K$ groups is represented by the parameter vector $\bthetam$ including all unique elements of the objects $\{(\pi_k^{(m)}, \bmu_k^{(m)}, \bSigma_k^{(m)}),\; k = \otK \}$.
Since different $\bthetam$ may refer to cluster configurations with a different number of groups, depending on the context, we will often use the notation $K(\bthetam)$, or $K(m)$, to denote the number of groups described by $\bthetam$.
The set of possible configurations is denoted with $\mathcal{M}$.
The superscript $(m)$ is dropped if it is unnecessary to index more than one cluster configuration, $m \in \mathcal{M}$.
Note that $\btheta$ is a \emph{parameter} serving as a general description of the clustered region but, in general, we do not presume that $F$ is necessarily a function of $\btheta$.
Given a configuration $\btheta$, we look for clusters that form a partition of the data space into $K$ disjoint subsets $\mathcal{Q}(\btheta)= \set{Q_k(\btheta),\; k=\otK}$,
\begin{equation}\label{eq:qs_partition}
Q_k(\btheta) \coloneqq \set{\bx \in \setR^p : \; \qs(\bx, \btheta_k) = \max_{1 \leq j \leq K} \qs(\bx,\btheta_j) },
\end{equation}
where $\qs(\bx, \btheta_k)$ is the quadratic score function at $\bx$ according to $\btheta_k$, that is
\begin{equation}\label{eq:quadratic}
\qs(\bx,\btheta_k) \coloneqq %%
\log(\pi_k) %%
- \frac{1}{2} \log\left(\det(\bSigma_k)\right) %%
- \frac{1}{2} \tra{(\bx-\bmu_k)}\inv{\bSigma_k}(\bx-\bmu_k).
\end{equation}
From now onward, we call $\mathcal{Q}(\btheta)$ the \emph{quadratic partition}.
A point $\bx$ is defined to belong to the group for which the quadratic score is maximized. Hence, $\qs(\bx,\btheta_k)$ can generally be interpreted as a measure of the fit of $\bx$ into the $k$\nth cluster according to $\btheta_k$.
Note that $\exp(\qs(\bx; \bmu_k, \bSigma_k)) \propto \pi_k\phi(\bx; \bmu_k, \bSigma_k)$, where $\phi(\cdot, \bmu_k, \bSigma_k)$ is the multivariate normal density function with mean $\bmu_k$ and covariance $\bSigma_k$.
The classical interpretation of \eqref{eq:qs_partition} is that it represents the optimal classification boundaries under the Gaussian assumption.
As noted in \cite{hastie2001discussion}, in practice, the quadratic score can effectively describe partitions well beyond Gaussianity whenever quadratic and linear boundaries can adequately separate clustered regions.
The following result states that the partition in \eqref{eq:qs_partition} is consistent with a class of elliptic-symmetric models that includes the Gaussian.
\begin{proposition}\label{stm:coherence_qsp}
Assume $\prob{Z_k = 1 } = \pi_k$ and that for all $k= \otK$ the group-conditional distribution, \ie the distribution of $X \given Z_k=1$, has density function
\begin{equation}\label{eq:familyQDA}
f(\bx; \bmu_k, \bSigma_k) = %%
 \det(\bSigma_k)^{-\frac{1}{2}} %%
 g \left( \tra{(\bx-\bmu_k)}\inv{\bSigma_k}(\bx-\bmu_k)\right),
\end{equation}
where $g(\cdot)$ is a strictly decreasing function on $[0,+\infty)$, $\bmu_k \in \setR^p$ is the centrality parameter and $\bSigma_k\in \mathbb{R}^{p\times p}$ is a positive definite scatter matrix.
Assume at least one of the following:
\begin{description}
\item [\cndGaussian] 
$f(\cdot)$ is the Gaussian density function (for an appropriate choice of $g(\cdot)$);

\item [\cndEqualPrecision] 
$\pi_i\det(\bSigma_i)^{-\frac{1}{2}}=\pi_j\det(\bSigma_j)^{-\frac{1}{2}}$, $i\neq j$, $i,j = \otK$.
\end{description}
Then, for any partition of the feature space $\set{A_k,\; k=\otK}$,
\begin{equation}\label{eq:pr_best_partition}
\prob{\bigcup_{k=1}^K \set{Z_k=1 \cap X \in A_k}} %%
\; \leq \; %%
\prob{\bigcup_{k=1}^K \set{Z_k=1 \cap X \in Q_k(\btheta)}},
\end{equation}
where $Q_k(\btheta) \in \mathcal{Q}(\btheta)$ is defined in \eqref{eq:qs_partition}.
\end{proposition}
The previous result connects and develops ideas from linear classification and its connections to elliptically-symmetric families investigated in \cite{velilla2005consistency}.

\begin{remark}\label{rmk:esd_conditional_model}
The quadratic partition achieves the largest probability that its members contain points generated from the $K$ sub-populations.
The group-conditional model \eqref{eq:familyQDA} includes popular unimodal models like the Gaussian, the Student-t, the Laplace, the multivariate logistic, \etc.
These models generate groups of points lying in regions that are intersections of ellipsoids described by the pairs $(\bmu_k, \bSigma_k)$ and, within each group, the features are connected via their joint linear dependence.
The generating mechanism assumed in Proposition~\ref{stm:coherence_qsp} is consistent with data generated from finite mixtures of such elliptically-symmetric families.
Outside the Gaussian case {\cndGaussian}, Proposition~\ref{stm:coherence_qsp} is restricted to the cases where groups have a comparable square root of the generalized precision, $\det(\bSigma_k)^{-\frac{1}{2}}$, after weighting by the cluster size $\pi_k$.
A special case of \cndEqualPrecision is when groups are balanced (equal sizes $\pi_k$) and homoscedastic (equal dispersions $\bSigma_k$).
\end{remark}

%% Imagine a machine that randomly shoots balls onto the tennis field so that these balls tempt to concentrate in certain areas of the field.
%We are given a list of possible ways to position a number of boxes onto the tennis field.
%At the end of the game we want to arrange the boxes so that the probability that the boxes are just right to catch all the balls is large.
%The parallel here is that

\subsection{Scoring cluster configurations}\label{sec:qs:Hn_Sn}
Given $\mathbb{X}_n$, we want to measure how well a cluster configuration $\bthetam$ organizes these points within the quadratic partition.
 We want to select the ``{boxes}'' $\set{Q_k(\btheta),\; k=\otK}$ that best represents the clustered points.
 Let $B_\eps^k(\bx_i; \btheta)$ be a ball of radius $\eps>0$, centered at $\bx_i$, such that $B_\epsilon^k(\bx_i; \btheta) \subset Q_k(\btheta)$, \ie $B_\eps^k(\bx; \btheta) \coloneqq \set{\by \in \setR^p: \norm{\by-\bx} < \eps \cap \by \in Q_k(\btheta) }.$
For $\eps$ sufficiently small, the joint probability that all points in $\mathbb{X}_n$ are accommodated in the quadratic partition consistently with the underlying group memberships is
\begin{equation}\label{eq:prod_joint_prob_clust_conf}
\prod_{i=1}^{n} \prob{Z_k=1 \cap X_i \in B_\eps^k(\bx_i; \btheta)} %
= %
\prod_{i=1}^{n} \prob{Z_k=1} \prob{X_i \in B_\eps^k(\bx_i; \btheta) \given Z_k=1}.
\end{equation}
Under the generating process of Proposition~\ref{stm:coherence_qsp}, taking $\eps \to 0$, the probability law \eqref{eq:prod_joint_prob_clust_conf} is transformed into its density representation
\begin{equation}\label{eq:hard_score_L}
\mathcal{L}_n(\btheta) \coloneqq \prod_{i=1}^{n} \prod_{k=1}^{K(\btheta)} %
\left( \pi_k f(\bx_i; \bmu_k, \bSigma_k) \right)^{\ind\set{\bx_i \in Q_k(\btheta)}},
\end{equation}
where $\ind\set{\cdot}$ is the usual indicator function.
\eqref{eq:hard_score_L} closely resembles the likelihood function for a partition model \cite[see][Ch.
7]{fruhwirth2019handbook}. 
However, this is not exactly the case: for a partition model, we would have had class membership indicators replacing $\ind\set{\bx_i \in Q_k(\btheta)}$ in \eqref{eq:hard_score_L}.
Taking the logarithm of \eqref{eq:hard_score_L}, we would like to achieve the largest 
\begin{equation}\label{eq:Ln}
L_n(\btheta) = \frac{1}{n}\sum_{i=1}^{n} \sum_{k=1}^{K(\btheta)} %
{\ind\set{\bx_i \in Q_k(\btheta)}} \log(\pi_k f(\bx_i; \bmu_k, \bSigma_k)).
\end{equation}
Evaluation of \eqref{eq:Ln} requires the knowledge of the specific group-conditional model $f(\cdot)$.
However, we want to evaluate the quality of the partition even when the group-conditional distribution is not precisely known.
Proposition~\ref{stm:coherence_qsp} states that, for certain group-conditional distributions, point-wise maximization of the quadratic score in the feature space well captures the main clustered regions.
We propose to rank cluster configurations based on the following \emph{hard score criterion}:
\begin{equation}\label{eq:Hn}
H_n(\btheta) = \frac{1}{n}\sum_{i=1}^{n} \sum_{k=1}^{K(\btheta)} %
{\ind\set{\bx_i \in Q_k(\btheta)}} \qs(\bx_i; \btheta_k).
\end{equation}
We call it \emph{hard} because $H_n(\cdot)$ is a weighted average of the points score with the 0--1 ``hard'' weights $\ind\set{\bx_i \in Q_k(\btheta)}$.
Interpreting $\qs(\bx_i; \btheta_k)$ as the strength at which the object $i$ is assigned to the $k$-th group, \eqref{eq:Hn} is the average strength achieved by a cluster configuration.
Despite this qualitative interpretation of $H_n(\cdot)$, there is a connection between \eqref{eq:Ln} and \eqref{eq:Hn} at the population level, based on the fact that $\qs(\bx_i; \btheta_k)$ contains the kernel of the Gaussian density.
Under regularity conditions, both sample averages \eqref{eq:Ln} and \eqref{eq:Hn} will asymptotically approach their population counterparts
\begin{equation}\label{eq:LH}
L(\btheta) = \sum_{k=1}^{K(\btheta)} \int_{Q_k(\btheta)} \log(\pi_k f(\bx; \btheta_k)) dF \quad\text{and}\quad H(\btheta) = \sum_{k=1}^{K(\btheta)} \int_{Q_k(\btheta)} \qs(\bx; \btheta_k) dF,
\end{equation}
respectively.
 The following proposition clarifies the relationship between $\Hd$ and $L(\cdot)$.
\begin{proposition}\label{stm:H_vs_L}
Assume that the following integrals exist and that
\begin{description}
\item[\cndInfDiffLoglik]
$\inf_{\bthetam \in \Theta_M} \set{\int_{Q_k(\bthetam)} \log f(\bx;\bmu_k^{(m)}, \bSigma_k^{(m)})dF - \int_{Q_k(\bthetam)} \log \phi(\bx; \bmu_k^{(m)}, \bSigma_kk)dF} \geq 0$ for all $k = 1,2,\ldots, K(\bthetam).$
\end{description}
Then
\begin{equation}\label{eq:H_L_penalty}
H(\bthetam) = c + L(\bthetam) - \Lambda(\bthetam),
\end{equation}
where $c$ is a positive constant, and
\[
\Lambda(\bthetam) = \sum_{k=1}^{K} \int_{Q_k(\bthetam)} %
\log \left( \frac{f(\bx; \bmu_k^{(m)}, \bSigma_k^{(m)})}{\phi(\bx; \bmu_k^{(m)}, \bSigma_k^{(m)})}
\right) dF \geq 0.
\]
\end{proposition}

At the population level, the hard score criterion can be interpreted as the quality of the fitting of the partition, expressed by $L(\cdot)$, minus a penalty term, $\Lambda(\bthetam) \geq 0$, that measures the departure from the Gaussian clusters' prototype model embedded into the quadratic score function.
When clusters are truly Gaussian, \ie $f(\cdot) = \phi(\cdot),$ then $\Lambda(\bthetam) = 0$ and $H(\bthetam) \propto L(\bthetam)$.
 Condition \cndInfDiffLoglik is needed to interpret the criterion: it ensures that $\Lambda(\bthetam) \geq 0$ for any possible cluster configuration $\bthetam$ under comparison so that it works as a penalty.
{\cndInfDiffLoglik} obeys to \emph{the natural principle} that, whenever we pick a configuration $\btheta$, the approximating Gaussian model underlying $\qs(\cdot)$ can not fit the quadratic regions better than the underlying true generating model $f(\cdot)$.
Indeed, {\cndInfDiffLoglik} is violated if there exists a configuration $\bthetam$ for which $\int_{Q_k(\bthetam)} \log f(X;\bmu_k^{(m)}, \bSigma_k^{(m)})dF < \int_{Q_k(\bthetam)} \log \phi(X; \bmu_k^{(m)}, \bSigma_k^{(m)})dF$, where these integrals can be seen as the expected log-likelihood contribution over the $k$-th members of the quadratic partition under $f(\cdot)$ and $\phi(\cdot)$, respectively.
From Proposition~\ref{stm:H_vs_L}, it immediately follows that
\[ \argmax_{m\in\mathcal{M}} H(\bthetam) = \argmax_{m\in\mathcal{M}}\set{L(\bthetam) - \Lambda(\bthetam)}. \]

Since $\qs(\cdot)$ measures the strength at which a point is assigned to a cluster, a smooth weighting is obtained by normalizing the quadratic scores.
We propose to use the softmax transformation, that is the $i$-th point's weight into the $k$-th cluster is
\begin{equation}\label{eq:tau_def}
\tau_k(\bx_i; \btheta) = \frac{\exp(\qs(\bx_i;\btheta_k))}{\sum_{i=1}^n \exp(\qs(\bx_i;\btheta_k))}.
\end{equation}
The corresponding \emph{smooth score criterion} is defined as
\begin{equation}\label{eq:Tn}
T_n(\btheta) = \frac{1}{n} \sum_{i=1}^{n} \sum_{k=1}^{K(\btheta)} %
\tau_k(\bx_i; \btheta) \qs(\bx_i;\btheta_k).
\end{equation}
Other weighting schemes are possible, but the choice of the softmax transformation is because it guarantees some form of optimality for Gaussian clusters (see the following proposition).
Under regularity conditions, for sufficiently large $n$, \eqref{eq:Tn} will approach its population counterpart
\begin{equation}\label{eq:T}
T(\btheta) = \sum_{k=1}^{K(\btheta)} \int \tau_k(\bx; \btheta) \qs(\bx;\btheta_k)dF.
\end{equation}

Under the generating mechanism of Proposition~\ref{stm:coherence_qsp}, the unconditional distribution of $X$ has the finite mixture density
\begin{equation}\label{eq:psi_dens_f}
\psi_f(\bx;\btheta):= \sum_{k=1}^{K(\btheta)} \pi_k f(\bx; \bmu_k, \bSigma_k).
\end{equation}
For a sample point $\bx_i \in \mathbb{X}_n$, under \eqref{eq:psi_dens_f} define the \emph{posterior weights}
\begin{equation}\label{eq:omega_posteriors}
 \omega_{f,k}(\bx_i; \btheta) %
= \prob{Z_{ik} = 1 \given \mathbb{X}_n} %
= \frac{\pi_k f(\bx_i; \bmu_k, \bSigma_k)}{\psi_f(\bx_i; \btheta)}.
\end{equation}
The ratios defined in \eqref{eq:omega_posteriors} are central in MBC methods where the $i$\nth object is assigned to the $k$-th component by the following rule 
\begin{equation}\label{eq:map}
\hat z_k(\bx_i; \btheta) = %%
\ind \set{k = \argmax_{1 \leq j \leq K(\btheta)} \; \omega_{f,k}(\bx_i; \btheta) };
\end{equation}
in practice, $\btheta$ is replaced with an estimate.
Typically, $\btheta$ is fitted based on an ML-type estimator, numerically approximated with the EM-algorithm \citep{mclachlan2000finite}.
 The rule \eqref{eq:map}, called MAP, retrieves the unobservable membership variables $\set{Z_{ik}}$ and coincides with the optimal Bayes classifier if the group-conditional model holds.
The MAP rule produces a hard assignment from the smooth (also called fuzzy) membership weights in \eqref{eq:omega_posteriors}.
The overall uncertainty of the assignment \eqref{eq:map} reflecting \eqref{eq:omega_posteriors} is captured by
\begin{equation}\label{eq:ent_Z_conditional_X}
 \ent(X; \btheta) = -\sum_{k=1}^{K(\btheta)} \omega_{f,k}(X; \btheta) \log \omega_{f,k}(X; \btheta),
\end{equation}
which is the entropy of the conditional distribution of $Z \given X$.
In situations where clusters are strongly separated the posteriors weights \eqref{eq:omega_posteriors} will be close to either 1 or 0 for most points, and the MAP assignment will produce ``clear clusters'', reflecting the low entropy of $Z \given X$.
On the other hand, cluster configurations with substantial overlap will exhibit large entropy.
Let $\psi_{\phi}$ be the mixture model \eqref{eq:psi_dens_f} when the group-conditional model is the Gaussian density $\phi(\cdot)$, and let $\ent_\phi(\cdot)$ be the corresponding entropy.
Moreover, let $\dkl(f_0||g)$ the Kullback-Leibler discrepancy from the approximating model $g$ to the ``true'' model $f_0$.

\begin{proposition}\label{stm:T_vs_dkl_plus_ent}
Let $f_0$ be the density function corresponding to the ``true'' underlying population distribution function $F$.
Then
\begin{equation}\label{eq:T_vs_dkl_entropy}
\argmax_{m \in \mathcal{M}} \; T(\bthetam) %
= %
\argmin_{m \in \mathcal{M}} \; \set{\dkl(f_0 \,||\, \psi_{\phi}(\cdot; \bthetam)) + \E[F]{\ent_{\phi}(X;\bthetam)}},
\end{equation}
where all expectations are assumed to exist and $\E[F]{\cdot}$ denotes the expectation under $F$.
\end{proposition}

Proposition~\ref{stm:T_vs_dkl_plus_ent} clarifies that $\Td$ looks for a compromise between the best approximation of $f_0$, in the sense of $\psi_\phi$, and the lowest entropy of the resulting assignment under the Gaussian prototype model.
The entropy term discourages the criteria from focusing on too complex clustering structures.
The term $\dkl(f_0 \,||\, \psi_{\phi}(\cdot; \bthetam))$ can be made arbitrarily small if $\bthetam$ is an overly rich description of the density regions produced by $F$.
Indeed, finite Gaussian mixtures can approximate any continuous distribution in a nonparametric sense \citep{nguyen2020approximation}.
However, an overly complex $\bthetam$ (\eg $K(\bthetam)$ is large) that describes the density regions too locally would imply a strong overlap and therefore a large $\ent_{\phi}(\cdot)$.

Propositions~\ref{stm:H_vs_L} and \ref{stm:T_vs_dkl_plus_ent} clarify the type of model reference-concept driving the proposed score selection.
\cite{baudry2015estimation} formulated a parameter estimation criterion based on the right-hand side of \eqref{eq:T_vs_dkl_entropy} to perform MBC.
In contrast, here $\Hd$ and $\Td$ are not meant to be parameter estimation criteria, as the ``true'' underlying generating model $F$ may well be not a function of the $\bthetam$ for  $m \in \mathcal{M}$.
This will become clearer in the examples of \secref{sec:qs:cluster_boundaries}, where we show an example where the maximum score can not identify the true underlying distribution even in the Gaussian case.

Under the Gaussian assumption, there is a further connection between the sample scores $H_n(\cdot)$ and $T_n(\cdot)$ and what is called observed \emph{complete data log-likelihood} into the MBC literature.
For details, we refer the reader to \sisec{3}.

\subsection{Clusters' boundaries.}\label{sec:qs:cluster_boundaries}

To see how $\Hd$ and $\Td$  define the clusters' boundary in Gaussian and non-Gaussian settings, consider the following examples. We define two data generating processes (dgp):
\begin{description}[leftmargin=!, labelwidth=\widthof{\textsc{dgpG~~}}]%
\item [dgpG] %
$F$ is a mixture of two spherical Gaussians in dimension $p=2$ with equal sizes $\pi_1=\pi_2=0.5$ and equal identity covariance matrix.
The first Gaussian component is centered at $\bmu_1 = \tra{(0,0)}$, while the second component has mean $\bmu_2=\tra{(d,0)}$, for some fixed $d>0$.

\item [dgpU] %
$F$ is a mixture of two uniform distributions with equal volume in dimension $p=2$ and $\pi_1=\pi_2=0.5$.
The first uniform distribution has support on the square $[-1,1]^2$ with center at $\bmu_1 = \tra{(0,0)}$.
The second uniform distribution takes value on the square $[d-1, d+1]\times[-1,1]$ with center at $\bmu_2 =\tra{(d,0)}$, for some fixed $d>0$.
\end{description}
In both cases, $d$ is the Euclidean distance between the clusters' centers.
For $d \in [0,10]$ we have different data generating processes.
For each value of $d$, we have a different generating distribution function, $F_d$, that is a mixture of: two Gaussian components in dgpG; two uniform components in dgpU.
The dgpU is introduced as a substantial departure from the elliptic assumption of Proposition~\ref{stm:coherence_qsp}.

We recall that the cluster configuration parameter $\bthetam$ represents the $m$-th configuration collecting the triplets $(\pim_k, \bmum_k, \bSigmam_k)$ representing the $k$-th cluster size, center and scatter.
At each $d$, we want to compare the population version of the score for two alternative cluster configurations $\{\btheta^{(1)}, \btheta^{(2)}\}$, where $K(\btheta^{(1)})=1$ and $K(\btheta^{(2)})=2$.
The number of possible choices of such configurations is infinite. Hence, we compare two possible specifications, $\btheta^{(1)}$ and $\btheta^{(2)}$, that try to reflect the group-conditional distributions corresponding to $F_d$.
The problem here is that the two types of $F_d$ considered in the example are not always a function of  cluster configuration parameters. 
In the dgpG case with $K=2$, the generating distribution $F_d$ is exactly specified in terms of proportion, mean and covariance parameters of the two Gaussian components.
However, for all the remaining cases, this is not true.
For example, in the dgpG case with $K=1$, we need to define $\btheta^{(1)} = (\pi^{(m)}, \bmu^{(m)}, \bSigma^{(m)})$ that does not coincide with the parameters of the corresponding $F_d$.
In each case, we defined competing cluster configuration parameters that are \emph{natural} descriptions of the group-conditional distributions. We have three different cases.
\begin{itemize}
\item \textbf{dgpG} and \textbf{dgpU with $K=1$.~} We set $\btheta^{(1)}=\left(\pi_1^{(1)}, \bmu^{(1)}, \bSigma^{(1)}\right)$ as follows
\begin{equation}\label{eq:dgpG_dgpU_k1}
\pi_1^{(1)} = 1, \qquad
\bmu^{(1)} = \int \bx dF_d,\qquad
\bSigma^{(1)} = \int \left( \bx-\bmu^{(1)}\right)\tra{\left( \bx-\bmu^{(1)}\right)}dF_d.
\end{equation}

\item \textbf{dgpG with $K=2$.~} %
This is the easiest case, because as previously noted, the parameters of $F_d$ coincides with the parameters of the two groups.
In this case, $\btheta^{(2)}$ is defined as follows
\begin{equation}\label{eq:dgpG_k2}
\pi_1^{(2)} = \pi_2^{(2)} = 0.5, \quad
\bmu^{(2)}_{1} = \tra{(0,0)}, \;
\bmu^{(2)}_{2} = \tra{(d,0)}, \quad
\bSigma^{(2)}_{1} = \bSigma^{(2)}_{2} =
\begin{pmatrix}
1 & 0\\
0 & 1
\end{pmatrix}.
\end{equation}

\item \textbf{dgpU with $K=2$.~} %
The main problem for this case is that a uniform distribution is not a function of a scatter parameter.
Both uniform components in dgpU have the same volume and, apart from their center, they would produce the same scatter of points.
First, we computed
\[
\bm{V}_U = \int \bx\tra{\bx}dU,
\]
where $U$ is the distribution function of a random variable $X$ uniformly distributed on the square $[-1,1]^2$.
$\bm{V}_U$ would be the covariance of such $X$.
The parameter $\btheta^{(2)}$ is set as follows
\begin{equation}\label{eq:dgpU_k2}
\pi_1^{(2)} = \pi_2^{(2)} = 0.5, \quad
\bmu^{(2)}_{1} = \tra{(0,0)}, \;
\bmu^{(2)}_{2} = \tra{(d,0)}, \quad
\bSigma^{(2)}_{1} = \bSigma^{(2)}_{2} = \bm{V}_U.
\end{equation}
\end{itemize}
Since some of the previous integrals, including those defining $H(\cdot)$ and $T(\cdot)$, can not be calculated analytically we computed their approximation (for each value of $d$) using Monte Carlo integration; all the integrals involved in the example are computed on completely independent experiments with $10^6$ random draws.
Each integral has been computed 100 times, and the results were averaged to obtain a Monte Carlo standard error consistently below $10^{-5}$.

\begin{figure}[!t]
\centering
\subfloat[Plots of $\Hd$\label{fig:fusion_Hard}]{%%
\includegraphics[height=0.15\textheight,keepaspectratio]%
{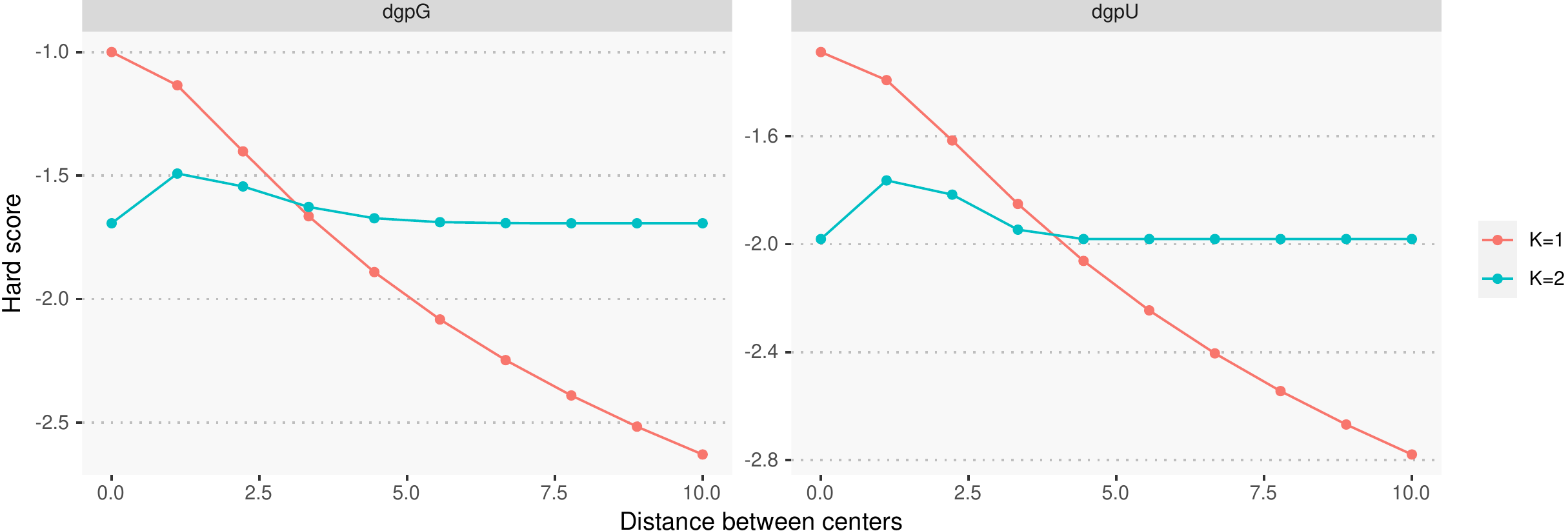}
}\\
\subfloat[Plots of $\Td$\label{fig:fusion_Smooth}]{%
\includegraphics[height=0.15\textheight,keepaspectratio]%
{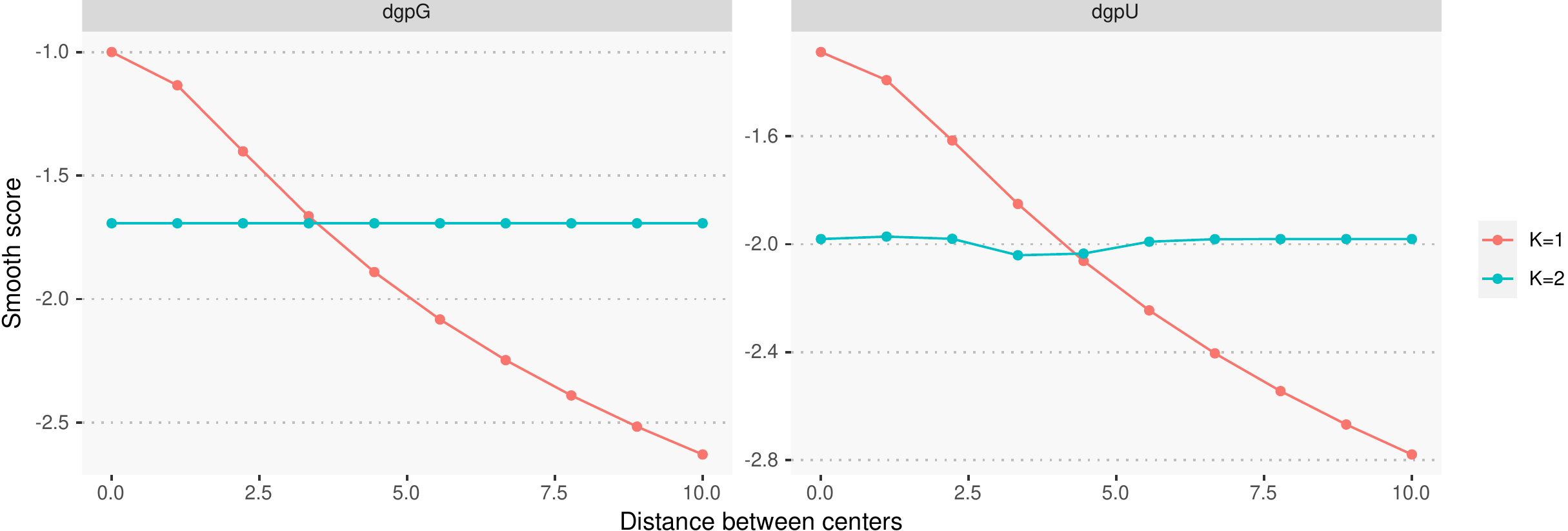}
}
\caption{Population score vs within-cluster distance for the dgpG and dgpU example.}
\label{fig:fusion}
\end{figure}
%Figures for simulated data results
\begin{figure}[!t]
    \centering
    \subfloat[dgpG; Hard Score\label{fig:fusion_graphs:GH}]{
        \includegraphics[width=0.9\textwidth, keepaspectratio]{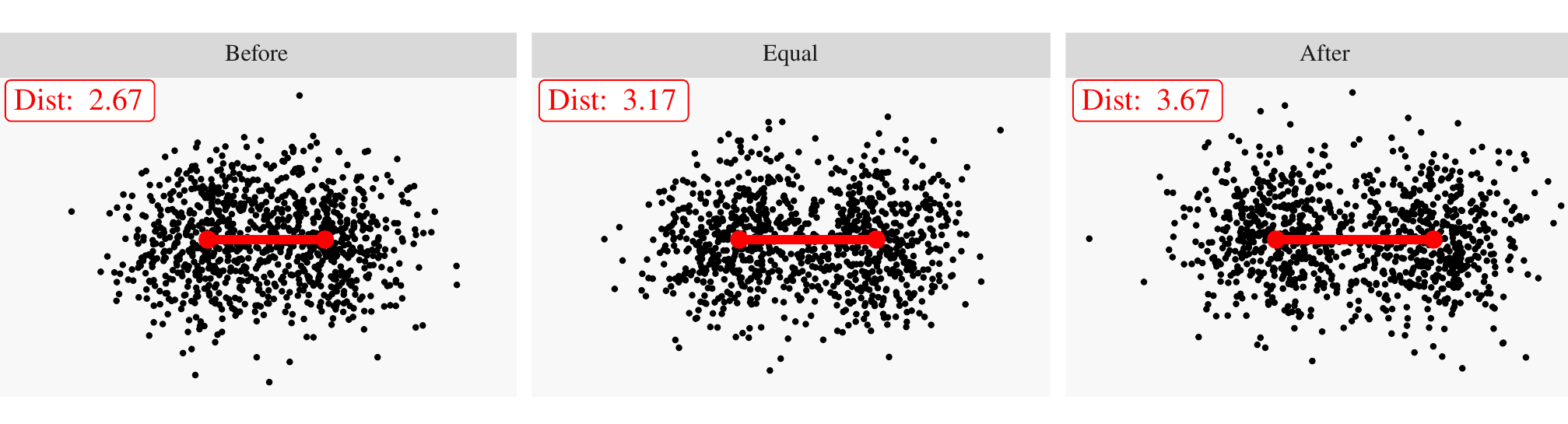}
    }

    \subfloat[dgpG; Smooth Score\label{fig:fusion_graphs:GS}]{
        \includegraphics[width=0.9\textwidth, keepaspectratio]{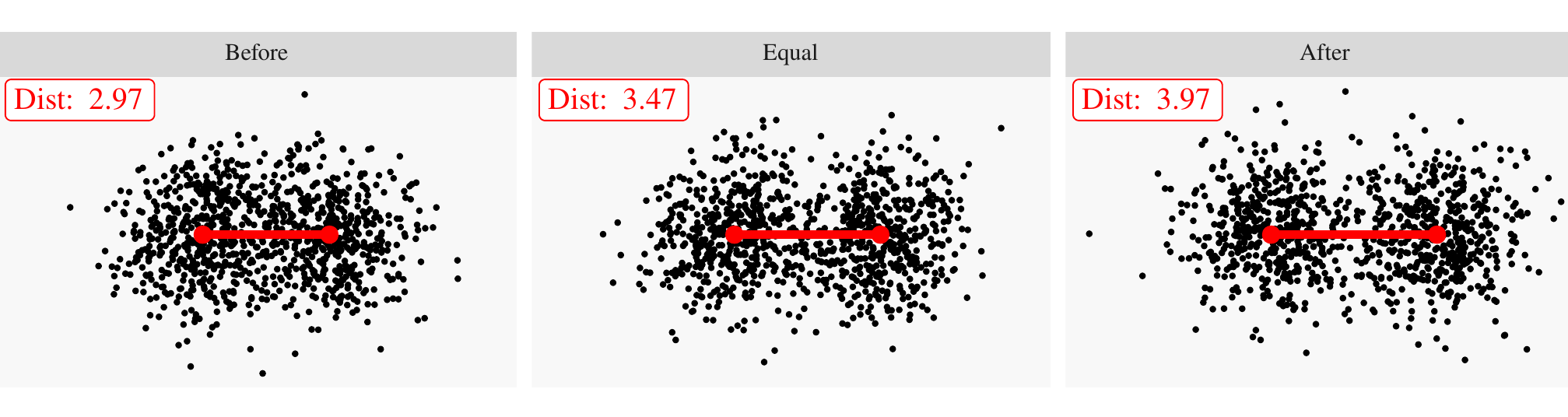}
    }

    \subfloat[dgpU; Hard Score\label{fig:fusion_graphs:UH}]{
        \includegraphics[width=0.9\textwidth, keepaspectratio]{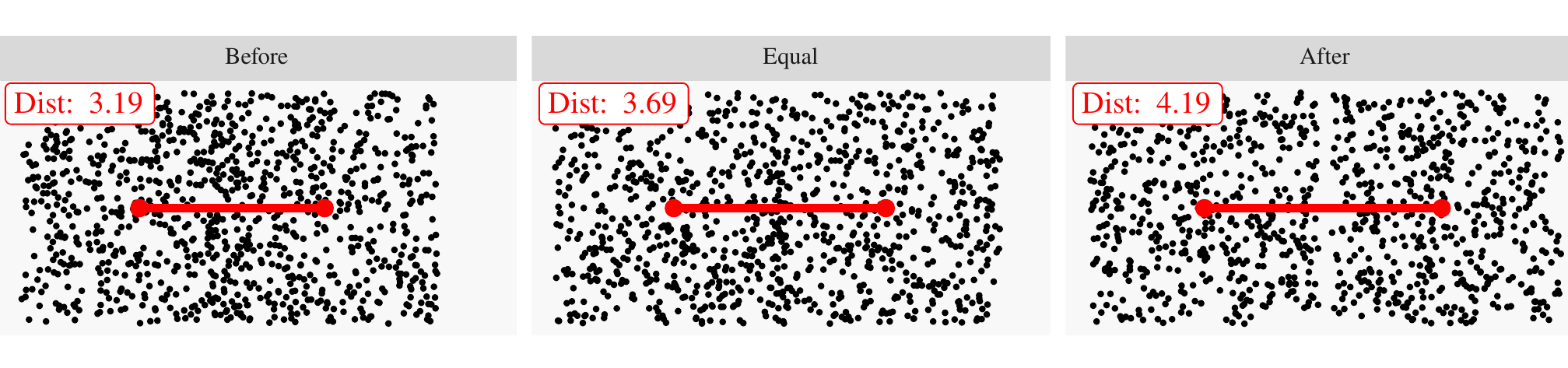}
    }

    \subfloat[dgpU; Smooth Score\label{fig:fusion_graphs:US}]{
        \includegraphics[width=0.9\textwidth, keepaspectratio]{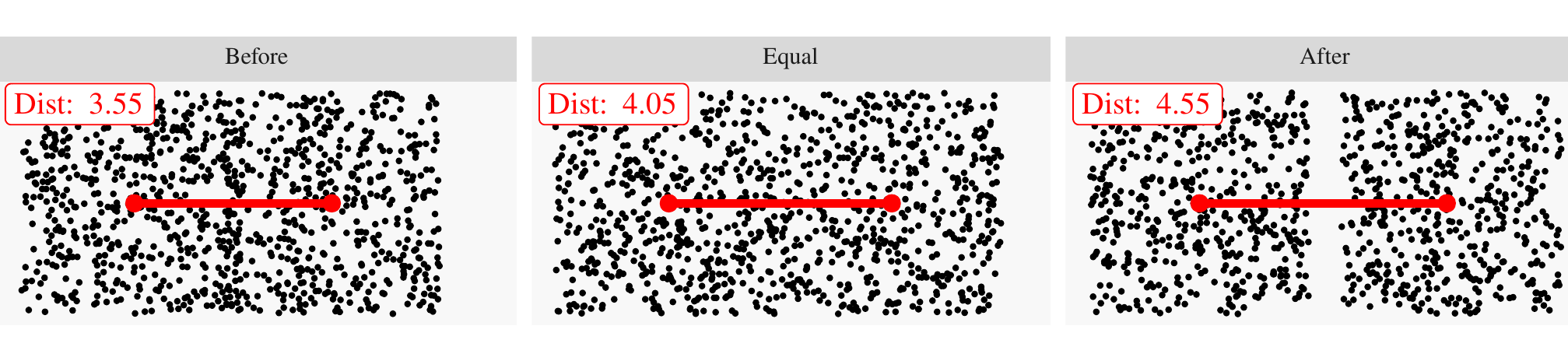}
        }
        \caption{Scatter plots of data sampled under the dgpG and dgpU sampling designs. In each of the four raw panels (a)--(d), we show the same sample design for three different values of $d$: the center plot refers to a value of $d$ such that the criterion ($H(\cdot)$ or $T(\cdot)$) values both cases (\ie $K=1$ and $K=2$) equally; the left plot refers to a value of $d$, where the criterion prefers $K=1$; the right plot refers to a value of $d$, where the criterion prefers $K=2$.
%Panels (a) and (b) represents the dgpG design, while panels (c) and (d) represent the dgpU design. The behavior of the hard scoring is shown in panels (a) and (c). Panels (b) and (d) refer to the smooth score criterion.
}\label{fig:fusion_graphs}
\end{figure}

\figref{fig:fusion_Hard} reports $\Hd$ \vs $d$. For both dgpG and dgpU, the hard score prefers a single cluster for low values of $d$.
The two clusters are split at $d=3.173$ for dgpG and $d=3.694$ for dgpU.
\figref{fig:fusion_graphs} shows examples of data produced by the two sampling designs around the point $d$ where $\Hd$ splits a single cluster into two clusters.
The general behavior of $\Hd$ is similar for both sampling designs.
Under $K=2$, for both dgpG and dgpU, there is evidence of a non-monotonic behavior of the criterion due to the hard weighting nature of $\Hd$.
Taking dgpG with $K=2$, $d$ only changes the position of the second group, and this is precisely reflected in the definition of $\btheta^{(2)}$.
We have the same quadratic regions accommodating data points in the same manner. The only difference introduced by $d$ is their location.
Therefore, one may expect a monotonic behavior of $\Hd$.
However, when $d$ decreases, overlapping the tail regions of the two distributions, both $Q_1(\btheta^{(2)})$ and $Q_2(\btheta^{(2)})$ start to lose tail points in favor of more central points where $\qs(\cdot)$ is larger.
Given the symmetric nature of the setup, for all larger values of $d$ both centers remain at an equal distance from the clusters' boundary.
Indeed, notice that the boundaries between $Q_1(\btheta^{(2)})$ and $Q_2(\btheta^{(2)})$ do not change at changing $d$, in this particular example.
This causes the tendency to split overlapped regions of points that one would not qualify as separate clusters.
This may be problematic in cases of strong overlap (as shown later, in \secref{sec:empirical_analysis}).
The behavior of $\Td$ is reported in \figref{fig:fusion_Smooth}.
For $K=2$, $\Td$ is flat for dgpG and almost flat for dgpU.
$\Td$ splits the two groups at slightly larger separation now: $d=3.47$ for dgpG and $d=4.05$ for dgpU.
$\Td$ does not attempt to split close clusters and is more appropriate to handle overlapped groups.
Scatter plots of data sets around the transition are shown in \figref{fig:fusion_graphs}.
Finally, we observe that  for all $d$, the true underlying model corresponds to $K=2$, but both scores will prefer $K=1$ for low values of $d$.
The latter implies that the maximum score can not identify the true underlying distribution even in the Gaussian case.

%%%EMACS  Local Variables:
%%%EMACS  ispell-local-dictionary: "american"
%%%EMACS  coding: utf-8
%%%EMACS  TeX-master: "../main.tex"   
%%%EMACS  End:

\section{Score selection via resampling}\label{sec:resampling}
The following discussion applies to both hard and smooth score criteria, therefore we unify the notation.
Rewrite both \eqref{eq:Hn} and \eqref{eq:Tn} as the average 
\begin{equation}\label{eq:Sn_and_s}
S_n(\btheta)= \frac{1}{n}\sum_{i=1}^{n} s(\bx_i; \btheta), %
\quad \text{where}\quad %
s(\bx; \btheta) := %
\sum_{k=1}^{K(\btheta)} w_k(\bx; \btheta) \qs(\bx; \btheta)
\end{equation}
is the cluster-weighted point-score.
With $w_k(\bx; \btheta) = \ind\set{\bx \in Q_k(\btheta)}$ we obtain the hard scoring, while $w_k(\bx; \btheta) = \tau_k(\bx; \btheta)$ returns the smooth score.
In Section~\ref{sec:qs}, we assumed a fixed list of candidate configurations, $\mathcal{M}$.
In practice, we work with a list of solutions obtained from applying different algorithms (and their various settings) to the only available data set $\mathbb{X}_n$.
Let $\hbtheta = \text{Clust}(\mathbb{X}_n) $ be a cluster configuration obtained by running a specific algorithm on $\mathbb{X}_n$; $\hbtheta$ reflects the sampling variability, the fitting method's variance and often an error equal to the difference between the method's true solution and its algorithmic approximation.
The clustering problem is affected by a mechanism similar to that of the bias-variance trade-off in predictive tasks.
Computing both $\hbtheta$ and $S_n(\cdot)$ using the same observed sample is not ideal because it will lead to an over-optimistic fitting: increasing the solution's complexity (\eg increasing $K$) improves the fit on the observed data, but does not necessarily guarantee a more coherent representation of the underlying clustering structure.
One way to overcome the previous issue is to make the fitting step independent of the validation step via resampling.
We explore two methodologies: cross-validation and bootstrap.

\begin{algorithm}[!th]
\caption{$k$-folds cross-validation of quadratic scores (CVQH, CVQS)}
\label{algo:cvscore}

\begin{algorithmic}[0]
    \State \textbf{input:} observed sample $\mathbb{X}_n$, clustering method $m\in \mathcal{M}$.
    \State \textbf{output:} $\widetilde{CV}^{(m)}$.
    \\
   \State (to ease notation, dependence on $m$ is dropped and reintroduced in step 3.1)
   \\
    \State (step 1)~~~randomly partition $\mathbb{X}_{n}$ into $k$ folds  $\set{\mathbb{X}^{(t)},\; t=1,\ldots, k}$, each with (approximately) $n/k$ data points.\\
    
    \For{$t = 1,\dots, k$}
    \State (step 2.1)~~~%
    $\hat \btheta^{(t)}  \gets \text{Clust}_{m}(\widehat{\mathbb{X}}),\quad $
    where  $\widehat{\mathbb{X}} \gets \bigcup_{j\neq t}\mathbb{X}^{(j)}$\\
    \State (step 2.2)~~~%%
    $S^{(t)}   \gets  \frac{1}{\# \mathbb{X}^{(t)}} %
                             \sum_{\by \in \mathbb{X}^{(t)}} s(\by ; \hat \btheta^{(t)})$
    \EndFor
    \\
    \State (step 3)~~~Compute:\qquad
    $\bar  S \gets \frac{1}{k}\sum_{t=1}^{k} S^{(t)}; \quad
    \hat \sigma_S \gets {\frac{1}{k-1}\sum_{t=1}^{k} \left(S^{(t)}  - \bar S\right)^{2}}$\\
    
    \State (step 3.1)~~~Compute:\qquad $\widetilde{CV}^{(m)}   \gets  \bar S  - \delta \,\frac{\hat \sigma_S}{\sqrt{k}}$ \\
   \hrulefill
\\
\State $CVQH = \argmax_{m\in \mathcal{M}}\left\{\widetilde{CV}^{(m)}\right\}$, when $s(\cdot)$ corresponds to  the hard quadratic score
\State $CVQS = \argmax_{m\in \mathcal{M}}\left\{\widetilde{CV}^{(m)}\right\}$, when $s(\cdot)$ corresponds to the smooth quadratic score
\end{algorithmic}
\end{algorithm}

\subsection{Cross-validation (CV)}\label{sec:resampling_cv}
CV is probably the most popular resampling method to perform model selection by separating the fitting and the testing step.
CV has been proposed to estimate $K$ in the MBC framework by \cite{smyth2000model}.
\cite{fu2020estimating} proposed the CV to select $K$ with the k-means algorithm.
The random CV method of \cite{smyth2000model} produces an estimate of expected Kullback-Leibler information loss under a reference mixture model over an independent test set.
Therefore, it is appropriate for tuning the mixture order for density approximation rather than clustering.
We consider estimating the expected score \eqref{eq:Sn_and_s} via the $k$-folds CV Algorithm \ref{algo:cvscore}. A clustering solution is selected by maximizing $\widetilde{CV}$; this defines the criteria CVQH and CVQS according to $s$ being the hard and smooth scores, respectively.
Rather than maximizing the average score criterion $\bar{S}$ computed in step 3, we look at the lower limit of an approximate confidence interval whose size depends on $\delta$.
Assuming the approximate normality of $\bar{S}$, $\delta=1.96$ would determine an approximate 95\% confidence interval. 
Although this may result in crude approximation due to the well-known difficulty to estimate the risk variance via CV \citep{bg04}, it allows to take into account the uncertainty about the estimated mean score, and it is rather popular in applications \citep[see ][]{hastie2009elements}.
In the numerical experiments, the selection based on the average criterion $\bar{S}$ led to inferior results compared with the approximate confidence interval rule of step 3.
Based on the experimental evidence we suggest $k=10$ folds and $\delta = 1.96$.  
The user may tune the constant $\delta$, but in our experiments, it produced relatively better results compared to the more common \emph{1-standard-error} rule, that is $\delta = 1$. Additional details about the CV-based selection methods (including the original proposal by \cite{smyth2000model}) are given in the \sisec{1}. 

Overall, CV-based methods did not perform well in the following experiments except for some specific cases. 
The latter is because the application of the CV framework to the clustering task is problematic.
CV is designed to estimate the prediction error of a model conditional on the training set, although \cite{bates2021crossvalidation} recently proved that CV does not achieve this goal in general.
However, clustering is not a prediction problem.
We want to assess how a certain $\bthetam$ describes the clustered structure produced by the underlying $F$.
Therefore, we need the fitted $\hbthetam$ and the sample on which the score is computed to convey the same information about the underlying $F$.
The CV aims to estimate a conditional prediction error, requiring that the train and the test set do not overlap, which often causes the two subsamples' structures to differ substantially in finite samples.
The latter is the primary motivation for introducing the following bootstrap method.

\begin{algorithm}[!t]
\caption{bootsrap scoring (BQH, BQS)}
\label{algo:bootscore}

\begin{algorithmic}[0]
    \State \textbf{input:} observed sample $\mathbb{X}_n$ (with ecdf $\mathbb{F}_{n}$), $\alpha \in (0,1)$; clustering method $m \in \mathcal{M}$.
   \State \textbf{output:} $\widetilde{W}_n$, $\widetilde{L}_n$, $\widetilde{U}_n$.
   \\
   \State (to ease notation, dependence on $m$ is dropped and reintroduced in step 3.1)
   \\
   \For{$b = 1,\dots, B$}
   \State (step 1.1)~~~$\mathbb{X}_n^{*(b)} \gets \set{x_i^{*(b)};\, i=1,2,\ldots, n} \iidsim \mathbb{F}_n$
   \State (step 1.2)~~~$\hbtheta^{*(b)} \gets \text{Clust}_{m}(\mathbb{X}_n^{*(b)})$ 
   \State (step 1.3)~~~$S^{*(b)}_{n}  \gets S_n(\hbtheta^{*(b)} ) = \inv{n} \sum_{i=1}^{n} s(x_i; \hbtheta^{*(b)})$
   \EndFor
   \State (step 2)~~~$\widetilde{W}_{n}   \gets  \frac{1}{B}\sum_{b=1}^B S^{*(b)}_{n}$
   \State (step 3)~~~Let $R_n^{*(b)} =  \sqrt{n} \left( S_n^{*(b)} -  \widetilde{W}_{n} \right)$
   \State (step 3.1)~~~Compute
   \begin{align*}
   &\widetilde{L}_n^{(m)} \gets\inf_{t}\left\{t:\frac{1}{B} \sum_{b=1}^B\ind\set{R_n^{*(b)}\leq
   t}\geq\frac{\alpha}{2}\right\};\quad
   &\widetilde{U}_n^{(m)} \gets\inf_{t}\left\{t:\frac{1}{B} \sum_{b=1}^B\ind\set{R_n^{*(b)}\leq
   t}\geq1-\frac{\alpha}{2}\right\}
   \end{align*}

    \\
    \hrulefill
    \\
\State $BQH = \argmax_{m\in\mathcal{M}}\left\{\widetilde{L}_{n}\right\}$ when $s(\cdot)$ corresponds to  the hard quadratic score
    \State $BQS = \argmax_{m\in\mathcal{M}}\left\{\widetilde{L}_{n}\right\}$ when $s(\cdot)$ corresponds to the smooth quadratic score
\end{algorithmic}
\end{algorithm}

\subsection{Bootstrap}\label{sec:resampling_boot}
Assume that $\hbtheta \sim G$, where $G$ reflects the randomness of the clustering output.
Assuming that $\hbtheta$ is independent of $X$, we want to construct a selection criterion that, at the population level, targets the quantity
$W = \oE_G[\oE_F[s(X; \hbtheta)]]$.
$W$ is the expectation over all possible realization of $\hbtheta$ of the expected cluster-weighted point score \eqref{eq:Sn_and_s}.
This approach is inspired by the seminal work of \cite{akaike1973information} on model selection.
%% Indeed, at the population level, AIC is defined as the expected Kullback-Leibler information loss over all possible estimates of the unknown parameter.
In practical situations, $G$ is not available, but the variations induced by $\hbtheta$ can be reproduced by repeating the clustering step on resampled versions of the data.
Let $\mathbb{F}_n$ be the ecdf of the sample; we propose to approximate $W$ using multiple independent samples obtained from $\mathbb{F}_n$.
The proposed estimation procedure is described in \algoref{algo:bootscore}, and it is based on the classical Efron's empirical bootstrap idea.
In steps (1.1)--(1.2) of \algoref{algo:bootscore}, independent bootstrap samples from the original data are used to reproduce the variations of $\hbtheta$.
In step (1.3), the original sample is used to compute the empirical approximation of the inner expectation of $W$ at the specific $\hbtheta^{*(b)}$.
Step (2) of \algoref{algo:bootscore} computes an estimate $\widetilde{W}_n$ of $W$ obtained as the expectation of the Monte Carlo approximation of the bootstrap distribution of $S_n^{*(b)}$.
Step (3) corresponds to the percentile method calculation of an approximate $(1-\alpha)$--confidence interval for $W$.
Calculation of a confidence interval for $W$ can be used to consider the uncertainty about $W$, reflecting both the sample variations and the variance of $\hbtheta$.
Let $W^{(m)}$ be the value of the expected score, $W$, produced by the $m$-th method/algorithm under comparison.
Rather than selecting the cluster configurations achieving the largest estimated $\widetilde{W}^{(m)}$, we propose to select the clustering corresponding to $\hbtheta^{(m^*)}$ where, for a fixed level of $\alpha \in (0,1)$, 
\begin{equation}\label{eq:L_selection}
    m^* = \argmax_{m \in \mathcal{M}} \; \widetilde{L}^{(m)}_n;
\end{equation}
this defines the BQH and BQS criteria when $s$ is the hard and smooth score, respectively.
In principle, one should fix $B$ large enough. The main drawback of \algoref{algo:bootscore} is that it requires refitting the clusters $B$ times for each clustering configuration $m \in \mathcal{M}$.
With the $k$-folds CV, one also needs to refit the clusters $k$ times, but the number of folds $k$ is usually much smaller than the number of required bootstrap data sets $B$.
In the large experiment shown in Section~\ref{sec:empirical_analysis}, we set $B=1000$.
In the \sisec{5.2}, we also provide evidence that even choosing $B=100$ did not change results substantially.

%% \begin{remark}[Bootstrap consistency]
%% We do not investigate the theoretical properties of the bootstrap procedure.
%% \algoref{algo:bootscore} is meant as a simple (and possibly brute force) approximation of the desired $W$.
%% However, the reader may wonder whether the bootstrap estimate $\widetilde{W}_n$ converges in some sense to the desired $W$ or whether the calculated confidence interval is valid.
%% It is well known that bootstrap theory can be technically involved depending on the statistical object that is resampled, that here is $\hbtheta$ \citep[see][and references therein]{efron2016computer}.
%% Formal guarantees for the \algoref{algo:bootscore} would essentially require certain strong regularity conditions on the mapping $\text{Clust} :\mathbb{X}_n^{*(b)} \to \hbtheta^{*(b)}$ involved in step (1.2).
%% It is undoubtedly interesting to check whether popular clustering algorithms lead to formal guarantees for the bootstrap.
%% However, this would involve extensive technical developments that are postponed to future research.
%% \end{remark}

%%%EMACS  Local Variables:
%%%EMACS  ispell-local-dictionary: "american"
%%%EMACS  coding: utf-8
%%%EMACS  TeX-master: "../main.tex"   
%%%EMACS  End:

\section{Experimental analysis}\label{sec:empirical_analysis}
In this section, we present an extensive experimental analysis of the selection problem.
The complexity of the following setting aims at offering a neutral comparison, where each competing method is expected to perform well in certain scenarios.
This is of utmost importance to achieve scientific progress in unsupervised learning, where global theoretical guarantees are rare, and most of the performances are shown via experimental studies \citep{mechelen2018benchmarking}.
Experiments are conducted on both real and simulated data. The latter are analyzed using Monte Carlo replicates, as explained later. \tabref{tab:data_designs} summarizes the different settings, giving a short description of the challenges that each setting poses for the selection problem.
A detailed discussion of the data is given in \sisec{4}.
In what follows, we describe the general aspects that are applied to all data sets in the experiments.

\begin{table}[t]
    \caption{Real data (top sub-table) and simulated designs (bottom sub-table). $n$: number of (simulated) points; $d$: dimensionality; $K$: true number of clusters.}\label{tab:data_designs}
    \scriptsize
    \centering
    \begin{tabular}{@{}l@{~}| l r r r l }
	&Data &n &d &K &Short Description\\
        \toprule
	\rowcolor{gray!15}
	\cellcolor{white} &Iris &$150$ &$4$ &$3$ &Measurements on Iris flowers; two classes show substantial overlap.\\
								     &Banknote  &$200$ &$6$ &$2$  &\makecell[l]{Measurements on original and counterfeit bills; the latter class is usually split\\in more groups due to the high variability of the measurements.}\\
	\rowcolor{gray!15}
	\cellcolor{white}					       &Olive  &$572$ &$8$ &$3$/$9$ &\makecell[l]{Olive oils' fatty acids. Features two different classifications; some classes scatters\\ are concentrated on lower dimensional hyperplanes and show substantial overlap.} \\
	\multirow{-6}{*}{\rotatebox[origin=c]{90}{Real}} &Wine &$178$ &$13$ &$3$ &\makecell[l]{Chemical analysis of wines grown from three different cultivars.\\High dimensions; balanced classes.} \\
        \midrule
        \midrule
							 &Pentagon5 &$300$ &$2$ &$5$ &\makecell[l]{Mixture of highly unbalanced Gaussian distributions;\\strong pairwise overlap of 4 of the 5 components.}\\
	\rowcolor{gray!15}
	\cellcolor{white} &T52D &$300$ &$2$ &$5$ &Mixture of 5 equal-proportions, well-separated Student-t components.\\
		       &T510D &$300$ &$10$ &$5$ &\makecell[l]{Adds 8 unclustered dimensions to T52D, increasing dimensionality without\\adding new clustering information.}\\
	\rowcolor{gray!15}
	\cellcolor{white} &Flower2 &$300$ &$2$ &$5$ &\makecell[l]{Mixture of 2 Student-t, 2 uniform and 1 spherical Gaussian;\\features regions of strong cluster overlaps.~~~~~~~~~~~~~~~~~~~~~~~~~~~~~~~~~~~~~~~~~~~~~~~~~~~~~~~~~~~~~~~~}\\
	 \multirow{-5}{*}{\rotatebox[origin=c]{90}{Simulated}}       &Uniform &$300$ &$2$ &$1$ &Uniform distribution; many criteria are not able to identify the unclustered case.\\
        \bottomrule
    \end{tabular}
\end{table}

\subsection{Clustering methods and algorithms}\label{sec:empirical_analysis:clustering_meth_algo} %%
For each data set, the set of candidate solutions for the selection problem is obtained by fitting a clustering method, $m\in\mathcal{M}$, to the data.
Each member $m\in\mathcal{M}$ is a solution obtained by an algorithm implementing a clustering method with a set of its specific hyper-parameters.
Hyper-parameters are the number of clusters $K\in\{1,\ldots,10\}$; often, restrictions and regularizers for clusters' covariance matrices (whenever possible); algorithmic initialization (for a subset of methods).
For each data set we consider $|\mathcal{M}|=440$ candidate solutions including: K-means and K-medoids partitions; ML for Gaussian mixtures with covariance matrix restrictions (as implemented in \pkg{mclust} software \citep{mclust2016paper}) or eigen-ratio regularization (as implemented in \pkg{otrimle} software \citep{r-otrimle,coretto2017consistency}); ML for Student-t and Skew Student-t mixtures (as implemented in \pkg{EMMIXskew} software \citep{r-emmix}).
Both Gaussian and Student-t based MBC methods are natural candidates to discover the cluster concept presented in \secref{sec:qs}.
On the other hand, we also consider Skewed Student-t models to assess the ability of the selection procedure to tame the overfitting issue usually arising with additional complexity.
In fact, the Skewed Student-t family contains both Gaussian and Student-t models as special cases.
In what follows, we occasionally refer to subsets of solutions in $\mathcal{M}$, named after the implementing software: \methodset{k-means}, \methodset{k-medoids}, \methodset{mclust}, \methodset{rimle} and \methodset{emmix}.
More details on the clustering methods are given in \sisec{1.1}.

\subsection{Selection methods}\label{sec:empirical_analysis:selection_meth}%%
We compare a large number of selection criteria over $\mathcal{M}$.
The list of existing criteria is vast.
Thus, we restrict the comparison to classical internal validation criteria routinely used by practitioners or those criteria rooted into the MBC literature that are more appropriate for pursuing the cluster notion of interest (for a detailed description see \sisec{1.2}).

\paragraph*{Method-independent criteria}
We consider the Caliński-Harabasz (CH; \cite{calinski1974dendrite}) and Average Silhouettes Width (ASW; \cite{kaufman1990partitioning}) indexes based on Euclidean distances.
While not designed to pursue the cluster's notion investigated in this paper, they are rather popular, and practitioners use them in various settings.
The bootstrap stability method of \cite{fang2012selection}, labeled as FW, is introduced in the comparison as another bootstrap-based alternative.
It pursues a stability notion rather than the validation philosophy developed here.

\paragraph*{Method-dependent criteria}
The strongest candidate to discover the cluster concept of interest are information criteria: AIC \cite{akaike1973information}, BIC \cite{schwarz1978estimating}, and ICL \cite{biernacki2000assessing}).
They can not be computed for members of $\mathcal{M}$  not derived from a probability model, or when the underlying model does not easily map into degrees-of-freedom (\eg \methodset{k-means}, \methodset{k-medoids} and \methodset{rimle}); this is summarized in \tabref{tab:algorithm_x_criteria}.
We also consider the methodology of \cite{smyth2000model}, labeled as CVLK, which minimizes a cross-validated risk based on the data likelihood.
CVLK also requires the definition of a models' likelihood function, therefore it can be applied to MBC methods only: \methodset{mclust}, \methodset{rimle} and \methodset{emmix}.

\begin{table}[t]
   \centering
   \caption{Possibility to compute clustering quality criteria (columns) for each type of configuration in $\mathcal{M}$.}\label{tab:algorithm_x_criteria}
   \tiny
   \begin{tabular}{lccccccccccccc}
       \toprule
       Configuration & AIC &BIC &ICL &ASW &CH &FW &CVLK &QH &QS &CVQH &CVQS &BQH &BQS \\
       \midrule
       \methodset{k-means} & & & &\checkmark &\checkmark &\checkmark & &\checkmark &\checkmark &\checkmark &\checkmark &\checkmark &\checkmark \\
       \methodset{k-medoids} & & & &\checkmark &\checkmark &\checkmark & &\checkmark &\checkmark &\checkmark &\checkmark &\checkmark &\checkmark \\
       \methodset{Mclust} &\checkmark &\checkmark &\checkmark &\checkmark &\checkmark &\checkmark &\checkmark &\checkmark &\checkmark &\checkmark &\checkmark &\checkmark &\checkmark \\
       \methodset{Rimle} & & & &\checkmark &\checkmark &\checkmark &\checkmark &\checkmark &\checkmark &\checkmark &\checkmark &\checkmark &\checkmark \\
       \methodset{Emmix} &\checkmark &\checkmark &\checkmark &\checkmark &\checkmark &\checkmark &\checkmark &\checkmark &\checkmark &\checkmark &\checkmark &\checkmark &\checkmark \\
       \bottomrule
   \end{tabular}
\end{table}

\paragraph*{Proposed selection criteria}
QH and QS select the clustering solution maximizing \eqref{eq:Hn} and \eqref{eq:Tn}, respectively.
They exploit in-sample information only, using the observed data both to estimate and score the solution; their results will motivate the need for resampling strategies as discussed above.
BQH and BQS methods are the bootstrapped version of the quadratic score method. They correspond to the maximization of \eqref{eq:L_selection} using hard and smooth scores, respectively.
For both BQH and BQS we set $B=1000$ for real data; this may be a demanding computing load for large data sets but, in practice, setting a much lower $B=100$ left the results almost unaltered (see \sisec{5}).
For the Monte Carlo experiments with simulated data, due to the higher computational load, we set $B=100$.

\subsection{Performance measures}\label{sec:empirical_analysis:performance_measure}
We measure the quality of the selected solutions in terms of: (\emph{i}) agreement with respect to true clusters' memberships; (\emph{ii}) the selected number of clusters compared to the ground truth.
Point (\emph{i}) captures similarity between true and fitted groups, and it is measured using the Adjusted Rand Index (ARI) of \cite{hubert1985comparing} and the Variation of Information Criterion (VIC) of \cite{meila2007comparing}.
$\text{ARI}\in[0,1]$, where ARI=1 means perfect agreement.
Originally, $\text{VIC} \in [0,\infty)$; however, we compute and report the negative of the VIC, so that a larger value means better agreement as for the ARI.
ARI and VIC are not only different in scales, but they capture the similarity differently.
The data sets present different challenges in retrieving the true classes.
We design situations where, even for some artificial data, the ``true clustering'' is not obvious and none of the 440 methods in $\mathcal{M}$ is able to reach near-to-perfect performances (\eg $\text{ARI}\approx 1$ and/or $\text{VIC}\approx 0$).
Nevertheless, here we do not compare clustering methods. In contrast, we study the problem of selecting the best available partition.
For this reason, besides comparing with the ground truth, we benchmark the 13 selection methods against the ``two best feasible partitions'', labeled as \textsc{best ari} and \textsc{best vic}.
These are obtained running the 440 methods' configurations on a data set and choosing the partitions achieving the best ARI and VIC, respectively.
Note that for some data sets, there are multiple members of $\mathcal{M}$ that give the same best feasible partition.

\section{Discussion of the results}\label{sec:result_discussion}
\tabref{tab:fullresults} summarizes the results on both real and simulated data, which are discussed in \ref{sec:result_discussion:real_data} and \ref{sec:result_discussion:simulated}, respectively.
Additional results and comments are given in \sisec{5}.

\begin{sidewaystable}
    \scriptsize
\centering
\caption{Experimental results. (Top sub-table) Selected solutions for real data. Cell value: selected solution's number of groups (\textsc{k}), ARI (\textsc{a}) and negative VIC (\textsc{v}). (Bottom sub-table) aggregated results for selected solutions on 100 MC replicates for each simulated design. Cell value: most frequently selected number of groups, with frequency in parentheses (\textsc{k}); ARI (\textsc{a}) and negative VIC (\textsc{v}), with standard errors in parentheses. Column \textsc{best} in both tables refer to the \textsc{best ari} (\textsc{a}) and \textsc{best vic} (\textsc{v}) solutions; when \textsc{best ari} and \textsc{best vic} select different number of groups or frequency (bottom sub-table), this is shown by different figures separated by ``-'', reporting \textsc{best ari} first. Best results are highlighted in bold.\label{tab:fullresults}}
\setlength{\tabcolsep}{2pt}
\begin{tabular}{l|c|l|lllllllllllll}
	 & & BEST    & AIC      & BIC     & ICL     & QH       & QS       & CH       & ASW     & FW      & CVLK   &CVQH &CVQS  &BQH     & BQS     \\
\toprule
%%Iris
\rowcolor{gray!15}
	  &\textsc{k}	  & 3 & 6  & 2 & 2 & 7  & 7  & 3  & 2 & 2 &4 &4 &4 & \textbf{3}  & \textbf{3}  \\
\rowcolor{gray!15}
     &\textsc{a} & 0.94 &0.57  &0.57 &0.57 &0.42  &0.42  &0.73  &0.57 &0.57 &0.81 &0.81 &0.81 & \textbf{0.9}  & \textbf{0.9}  \\
\rowcolor{gray!15}
\multirow{-3}*{Iris} &\textsc{v}  & -0.26 & -1.52 & -0.67 & -0.67 & -1.56 & -1.56 & -0.76 & -0.67 & -0.67 & -0.57 & -0.57 & -0.58 & \textbf{-0.32} & \textbf{-0.32} \\

%%Banknote
 &\textsc{k} & 2  & 6   & 3 & 3 & 10 & 10 & \textbf{2} & \textbf{2} &2 &3 &3 &3 &3 &3 \\
  &\textsc{a} & 1    & 0.6   &0.84 &0.84 &0.26 &0.26 &\textbf{1}  &\textbf{1}    &0.98 &0.85  &0.78 &0.78 &0.86 &0.86 \\
\multirow{-3}*{Banknote}  &\textsc{v} & 0 & -1.16 & -0.43 & -0.43 & -2.14 & -2.14 & \textbf{0} & \textbf{0} & -0.08 & -0.42 & -0.62 & -0.62 & -0.37 & -0.37 \\

%%Olive 3
\rowcolor{gray!15}
			  & \textsc{k} & 3 & 10 & 6 & 6 & 10 & 10 & 3 & 2 & \textbf{2} & 10 &7 &7 & 8 & 8 \\
\rowcolor{gray!15}
			  & \textsc{a} & 1    & 0.33 & 0.52 & 0.52 & 0.29 & 0.29 & 0.32  & 0.39 & \textbf{ 0.82} & 0.3  &0.28 &0.28 & 0.49 & 0.49 \\
\rowcolor{gray!15}
\multirow{-3}*{\makecell{Olive\\(K=3)}} &\textsc{v} &-0.03 &-1.74 &-1.42 &-1.42 &-1.84 &-1.84 &-1.88 &-1.28 &\textbf{-0.42} &-1.81 &-2.04 &-2.04 &-1.28 &-1.28\\

%%Olive 9
&\textsc{k} & 8 & 10 & 6 & 6 & 10 & 10 & 3 & 2 & 2 & 10 &7 &7 & \textbf{8} & \textbf{8} \\
&\textsc{a}& 0.88 &0.47 &0.76 & 0.76 &0.54 & 0.54 & 0.42  & 0.29 & 0.36 & 0.58   &0.44 &0.44 & \textbf{0.86} & \textbf{0.86} \\
\multirow{-3}*{\makecell{Olive\\(K=9)}} &\textsc{v} &-0.65 &-1.77 &-1.32 &-1.32 &-1.26 &-1.26 &-2.28 &-2.28 &-1.84 &-1.18 &-1.96 &-1.96 &-0.74 &-0.74\\

%%Wine
\rowcolor{gray!15}
&\textsc{k} & 3 & 3  & 3 & 3 & 8  & 8 & 10& 2 & 3 & 6  &5 &5 & \textbf{3}  & \textbf{3}  \\
\rowcolor{gray!15}
&\textsc{a}&0.98 & 0.44  & 0.84 & 0.84 & 0.46  & 0.46  & 0.15 & 0.37 & 0.87 & 0.59  &0.64 &0.64 & \textbf{ 0.9}  & \textbf{ 0.9}  \\
\rowcolor{gray!15}
\multirow{-3}*{Wine}      &\textsc{v} &-0.08 &-1.42 &-0.58 &-0.58 &-1.53 &-1.53 &-3.15 &-1.41 &-0.48 &-1.37 &-1.11 &-1.11 &-0.38 &-0.38\\
\midrule
\midrule
%% Pentagon5
&\textsc{k}  & 5 (83\%)-4 (52\%) & 5 (59\%) & \textbf{5 (45\%)} & 3 (88\%) & 10 (31\%) & 4 (39\%) & \textbf{3 (100\%)} & \textbf{3 (100\%)} & \textbf{3 (99\%)} & 6 (32\%) & 4 (42\%) & 3 (61\%) & 3 (90\%) & \textbf{3 (98\%)}\\
&\textsc{a}  &0.92 (0.02) &0.82 (0.11) & \textbf{0.88 (0.04)} &0.85 (0.03) &0.76 (0.12) &0.84 (0.07) & \textbf{0.84 (0.02)} & \textbf{0.84 (0.03)} & \textbf{0.84 (0.06)} &0.73 (0.13) &0.82 (0.09) &0.84 (0.05) &0.85 (0.03) & \textbf{0.84 (0.03)}\\
\multirow{-3}*{Pentagon5} &\textsc{v}  &-0.36 (0.07) &-0.66 (0.31) & \textbf{-0.43 (0.09)} &-0.42 (0.06) &-0.89 (0.36) &-0.57 (0.25) & \textbf{-0.43 (0.05)} & \textbf{-0.43 (0.06)} & \textbf{-0.43 (0.1)} &-0.85 (0.33) &-0.58 (0.25) &-0.48 (0.13) &-0.43 (0.06) & \textbf{-0.43 (0.06)}\\

%& T52D (place multirow at the end for color)
\rowcolor{gray!15}
 &\textsc{k} &5 (99\% - 97\%) &6 (21\%) &5 (90\%) &5 (95\%) &10 (41\%) &10 (21\%) &7 (40\%) &5 (86\%) &2 (81\%) &6 (41\%) &4 (43\%) &4 (54\%) & \textbf{5 (98\%)} & \textbf{5 (98\%)} \\
\rowcolor{gray!15}
 &\textsc{a}  &0.99 (0.01) &0.84 (0.13) &0.97 (0.04) &0.98 (0.01) &0.85 (0.1) &0.91 (0.08) &0.7 (0.1) &0.92 (0.15) &0.59 (0.18) &0.84 (0.13) &0.9 (0.11) &0.93 (0.08) & \textbf{0.99 (0.01)} & \textbf{0.99 (0.01)}\\
\rowcolor{gray!15}
\multirow{-3}{*}{T52D} &\textsc{v} &-0.06 (0.05) &-0.5 (0.32) &-0.12 (0.1) &-0.11 (0.08) &-0.55 (0.26) &-0.35 (0.25) &-0.7 (0.26) &-0.26 (0.32) &-0.93 (0.36) &-0.44 (0.27) &-0.32 (0.23) &-0.27 (0.19) & \textbf{-0.08 (0.06)} & \textbf{-0.08 (0.06) }\\

%%T510D
&\textsc{k}  &5 (99\%) &9 (24\%) &6 (50\%) & \textbf{5 (83\%)} &10 (98\%) &10 (98\%) &2 (100\%) &2 (100\%) &2 (94\%) &6 (45\%) &5 (36\%) &5 (38\%) &5 (69\%) & \textbf{5 (85\%)}\\
&\textsc{a}  &0.99 (0.01) &0.7 (0.13) &0.86 (0.1) & \textbf{0.94 (0.08)} &0.55 (0.08) &0.55 (0.07) &0.51 (0.03) &0.51 (0.03) &0.53 (0.11) &0.74 (0.13) &0.79 (0.13) &0.8 (0.13) &0.91 (0.12) & \textbf{0.94 (0.09)} \\
\multirow{-3}*{T510D} &\textsc{v} &-0.09 (0.06) &-1.06 (0.43) &-0.35 (0.18) & \textbf{-0.23 (0.15)} &-1.23 (0.32) &-1.2 (0.23) &-1.1 (0.05) &-1.1 (0.05) &-1.05 (0.21) &-0.75 (0.43) &-0.62 (0.33) &-0.61 (0.34) &-0.28 (0.26) & \textbf{-0.2 (0.18)}\\

%% Flower (place multirow at the end for color)
\rowcolor{gray!15}
&\textsc{k}  &5 (73\% - 77\%) &8 (24\%) &2 (58\%) &2 (65\%) &10 (85\%) &10 (59\%) &10 (87\%) & \textbf{5 (85\%)} &5 (86\%) &7 (43\%) &6 (27\%) &5 (23\%) & \textbf{5 (74\%)} & \textbf{5 (72\%)}\\
\rowcolor{gray!15}
&\textsc{a} &0.68 (0.06) &0.48 (0.1) &0.32 (0.1) &0.35 (0.12) &0.47 (0.07) &0.46 (0.08) &0.44 (0.04) & \textbf{0.45 (0.17)} &0.45 (0.1) &0.49 (0.11) &0.47 (0.13) &0.43 (0.14) & \textbf{0.53 (0.09)} & \textbf{0.46 (0.11)}\\
\rowcolor{gray!15}
\multirow{-3}*{Flower2}  &\textsc{v}  &-1.21 (0.17) &-1.91 (0.34) &-1.88 (0.22) &-1.8 (0.26) &-1.96 (0.21) &-1.88 (0.22) &-1.98 (0.17) & \textbf{-1.58 (0.29)} &-1.6 (0.21) &-1.8 (0.27) &-1.73 (0.27) &-1.74 (0.27) & \textbf{-1.51 (0.25)} & \textbf{-1.58 (0.24)}\\

%% Uniform
 &\textsc{k} &1 (100\%) &10 (51\%) &4 (65\%) &1 (77\%) &10 (90\%) &10 (71\%) &10 (46\%) &4 (74\%) &4 (64\%) &8 (30\%) &7 (22\%) &1 (85\%) &10 (82\%) & \textbf{1 (96\%)}\\
 &\textsc{a}  &1 (0) &0 (0) &0 (0) &0.77 (0.42) &0 (0) &0.16 (0.37) &0 (0) &0 (0) &0 (0) &0 (0) &0.06 (0.24) &0.85 (0.35) &0 (0) & \textbf{0.96 (0.19)}\\
\multirow{-3}*{Uniform} &\textsc{v}  &0 (0) &-3.01 (0.22) &-1.94 (0.4) &-0.22 (0.45) &-3.1 (0.11) &-2.61 (1.14) &-3.06 (0.41) &-2.1 (0.37) &-2.38 (0.59) &-2.84 (0.28) &-2.3 (0.81) &-0.25 (0.73) &-3.14 (0.31) & \textbf{-0.1 (0.54)}\\

\bottomrule
\end{tabular}
\end{sidewaystable}

\subsection{Results on real data sets}\label{sec:result_discussion:real_data}

The real data sets analyzed in this study are (\tabref{tab:data_designs}, top sub-table): the Iris data set of \cite{edgaranderson1936theiris, fisher1936theproblems}; the Banknote data set of \cite{flury1988multivariateapproach}; the Olive data set of \cite{forina1983classificationcomposition}, for which there are two possible true partitions (a coarser one with 3 classes corresponding to Italian geographical macro-regions, and a finer classification with 9 narrower geographical regions); the Wine data set of \cite{forina1988}.
Additional description and visualization is given in \sisec{4.1}.
Results presented in this section use $B=1000$ bootstrap resamples.
These are almost unaltered setting a much lower $B=100$ (see \sisec{5.2}).

\begin{figure}[H]
\centering
\includegraphics[width=\textwidth]{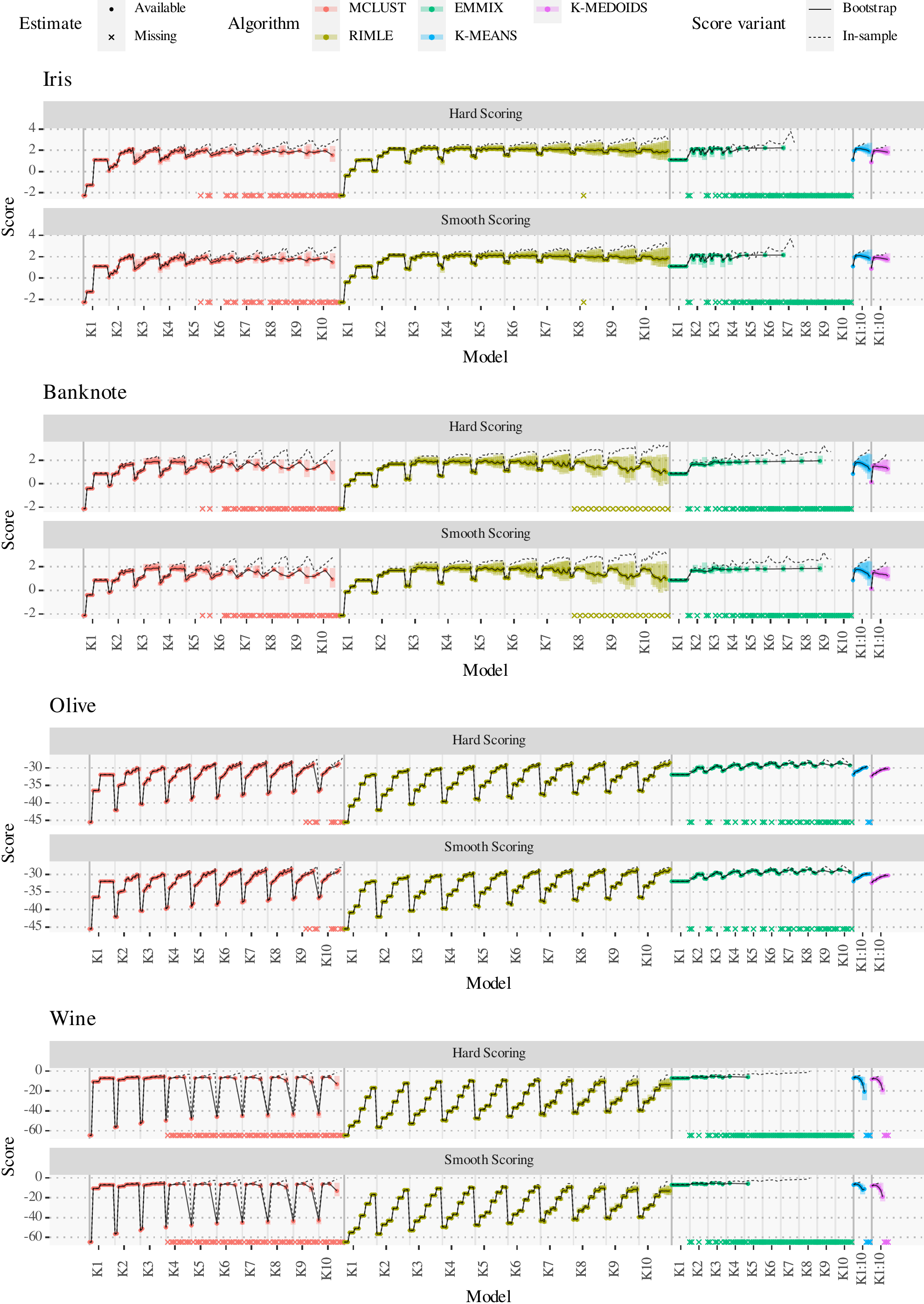}
\caption{Results for the Quadratic Smooth score criteria QS and BQS. Horizontal axes: the 440 $m\in\mathcal{M}$ are sorted by: clustering method (colors); increasing $K$ (axis ticks); increasing complexity (fewer restrictions on scatter matrices). Vertical axes: QH and QS (dashed lines); bootstrap estimated $\widetilde{W}_n$ (solid lines) with estimated confidence intervals at 95\% (shaded areas). Lower band corresponds to BQH and BQS. Missing solutions are reported with an $(\times)$-symbol (bottom of the plot).\label{fig:full_btss_1000b}}
\end{figure}

\figref{fig:full_btss_1000b} provides a graphical representation of the results for the proposed smooth score on the four data sets. Similar displays for the other data sets are shown in the \sisec{5.2}, using $B=100$.
For all the clustering methods, there is remarkable evidence that in-sample estimates of the score (QH and QS) become overly optimistic as the complexity of the clustering solutions increases.
Indeed, considering the Iris data, for $K>3$ (true number of groups) and increased model complexity, both QH and QS leave the scores' confidence intervals.
Moreover, as soon as $K(m)$ exceeds the true $K=3$, the more complex members of $\mathcal{M}$ also produce wider confidence bands, confirming the well-known pattern in the model selection that unnecessary additional model complexity introduces additional uncertainty.
An analogous pattern is found for the other data sets, although for Olive and Wine the vertical scale dominates the plots.
These results are robust to a lower $B=100$.

\tabref{tab:fullresults}, top sub-table, summarizes the selected solutions for all the clustering selection criteria on the four data sets (details of the selected solutions are shown in \sisec{5.2}).
First, note that the best feasible partitions available from $\mathcal{M}$ (\textsc{best}) do not always retrieve the underlying clusters perfectly, although they catch the true $K$ but for the Olive data with $9$ classes.
In this case, the best feasible solution corresponds to a configuration fitted by the \pkg{mclust} software with $K=8$ groups.

\begin{description}
    \item[Iris.] The selected solutions include partitions with a $K$ ranging from 2 to 6.  The true $K=3$ is detected by BQH, BQS and CH.
However, only BQH and BQS selected partition is very close to the best available.

    \item[Banknote.] The top performers are CH and ASW that discover the true partition exactly, with FW reporting a close performance. In this case, some methods, including BQH, BQS and ICL, provided a second-best performance fitting $K=3$ groups.  This is due to the heterogeneity of the ``counterfeit'' class, which a single ESD component can not adequately capture.

    \item[Olive.] Assuming 3-classes, only CH discovers 3 groups, but these are unrelated to the ground truth; FW reports the best, reasonably good ARI and VIC, with 2 groups mixing some of the underlying 3 classes. 
	Assuming $K=9$ classes, none of the selection methods discovers $9$ groups: BQH and BQS retrieve two partitions that are close to the best available in $\mathcal{M}$, while all other methods select solutions that are far away from the ground truth.

    \item[Wine.] Within the set of considered methods, it is almost possible to retrieve the true classes exactly. Nonetheless, all the methods show disappointing performances but for BIC, ICL, BQH and BQS. These four criteria select solutions with correct number of classes, but BQH and BQS outperform the other two, achieving better ARI and VIC, close to the optimal ones.
\end{description}

The overall conclusion are: (\emph{i}) BQH and BQS offer a similar performance, finding the best feasible partition or a partition close to it; (\emph{ii}) the in-sample versions of the quadratic score criteria, QH and QS, dramatically over-estimate $K$ in all situations; (\emph{iii}) all cross-validation alternatives showed a poor performance; (\emph{iv}) information-based criteria showed a mixed evidence.
AIC tends to select too complex solutions, while both BIC and the ICL select less complex solutions as expected.
BIC and ICL show a similar performance, selecting a reasonable partition in the case of the Banknote and Wine data.

\subsection{Monte Carlo experiments}\label{sec:result_discussion:simulated}

\begin{figure}[H]
\centering
\includegraphics[height=0.25\textheight, keepaspectratio]{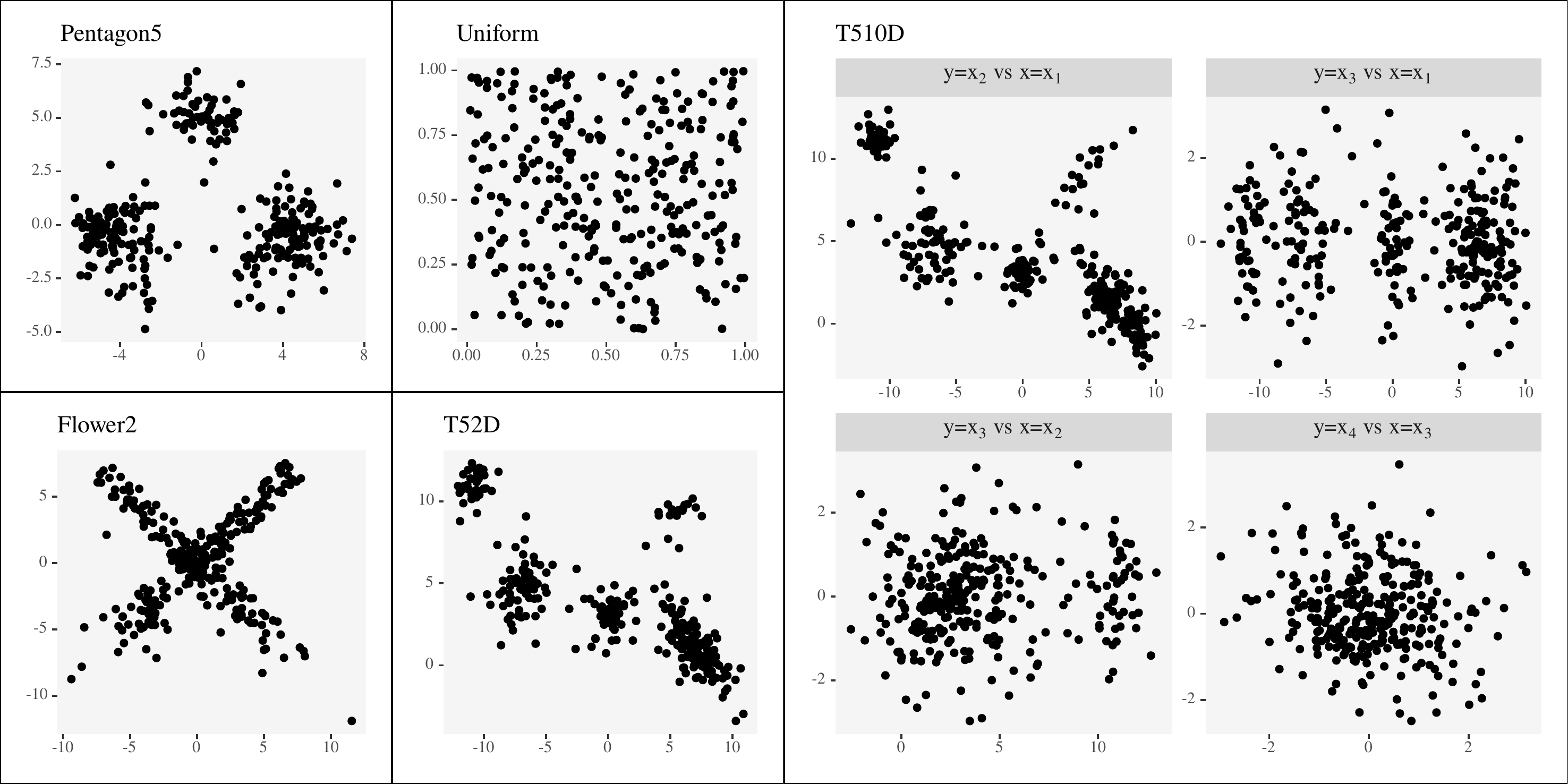}
\caption{Scatters produced by the 5 DGPs with $n=300$ in each case. For T510D (right) we plot the first two marginals ($x_{1}$ and $x_{2}$), a combination of them with an uninformative marginal ($x_{3}$) and two uninformative marginals ($x_{3}$ and $x_{4}$).}
\label{fig:simdata_example}
\end{figure}

In this section, we present experiments with data simulated from 5 different data generating processes (DGP), shown in \figref{fig:simdata_example}.
The DGPs are labeled as (\tabref{tab:data_designs}, bottom sub-table): \emph{Pentagon5}, \emph{T52D}, \emph{T510D}, \emph{Flower2} and \emph{Uniform}.
All DGPs produce data in dimension $p=2$ except for T10D, where $p=10$.
The Uniform design generates points drawn from a single 2-dimensional uniform distribution to test the behavior with unclustered data.
For all other DGPs, points are drawn from finite mixtures with 5 components. Pentagon5 generates points form Gaussian components, some of which are strongly overlapped and unbalanced. T52D generates points from reasonably separated Student-t components. T510D generates the same clusters as T52D on the first two coordinates while the remaining 8 dimensions are ``noisy features'' with a joint spherical distribution that does not carry any clustering information. Finally, Flower2 generates points from both uniform and ESD components.
A detailed description of the DGPs and additional data visualizations are available in \sisec{4.2}.
The ``true'' cluster membership of a point is identified with the corresponding mixture component generating it.
However, some DGPs produce situations that are not always in line with this ground truth definition.
For example, one may want to look for 3 clusters in Pentagon5, while 5 groups may not be necessarily the only appropriate description of Flower2's structure.
Some DGPs contain substantial departures from elliptic shapes, unbalanced groups and strong between-scatter discrepancies.
This is for testing the robustness of the proposed method in situations where the assumptions in Proposition~\ref{stm:coherence_qsp} are not exactly fulfilled.
Moreover, we fix $n=300$ for all simulated data sets.
The latter choice challenges some resampling criteria due to the strong stress it imposes on bootstrap resampling.
Indeed, empirical bootstrap may fail to replicate the distribution of small clusters when $n$ is small.

%% \textcolor{red}{Denoting with $B$ and $CV$ the number of bootstrap and cross-validation resamples (used for resample criteria) and with $MC$ the number of Monte Carlo replicates, the total number of estimated solutions in the simulation exercise is: $(B + CV +1) \times MC \times 5 \times 440$ (where 5 and 440 are the number of designs and clustering methods, respectively).
%% Given the massive simulation exercise, we set $B=100$ for a total of $24+$ million estimates.}

\begin{figure}[H]
    \centering
    \subfloat[Pentagon5\label{fig:simulated_results:pentagon5}]{
        \includegraphics[width=.30\textwidth, keepaspectratio]{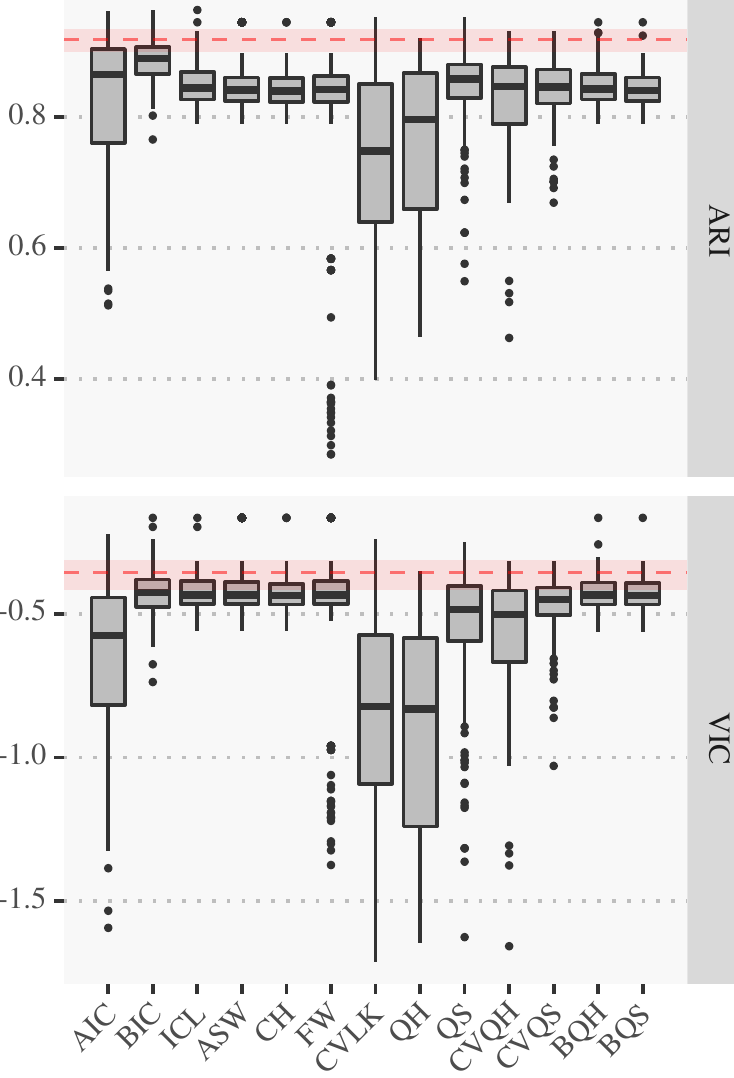}
    }
    \subfloat[T52D\label{fig:simulated_results:t52d}]{
        \includegraphics[width=.30\textwidth, keepaspectratio]{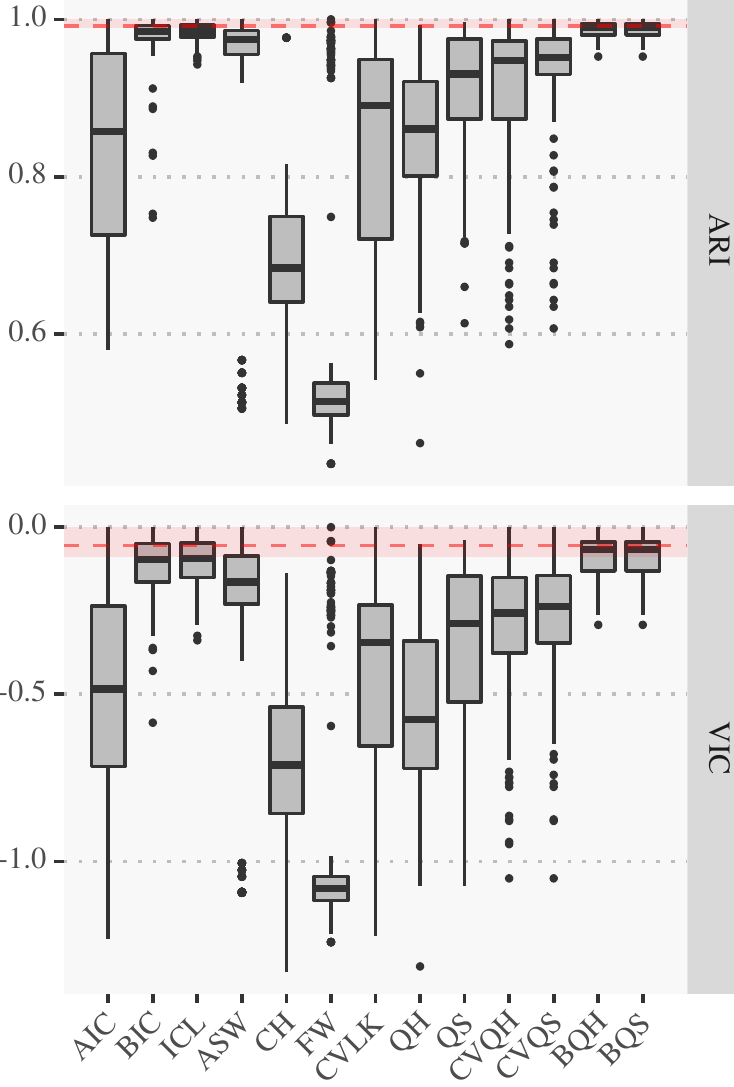}
    }
    \subfloat[T510D\label{fig:simulated_results:t510d}]{
        \includegraphics[width=.30\textwidth, keepaspectratio]{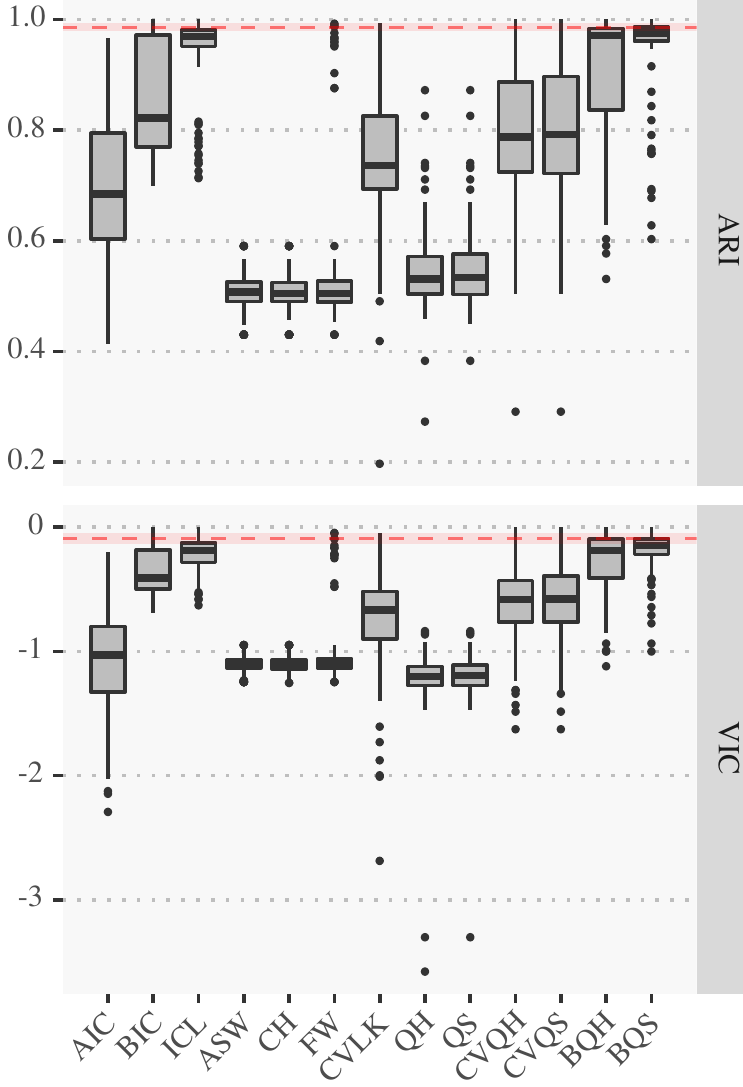}
        }

    \vspace{1ex}
    
    \subfloat[Flower2\label{fig:simulated_results:flower2}]{
        \includegraphics[width=.30\textwidth, keepaspectratio]{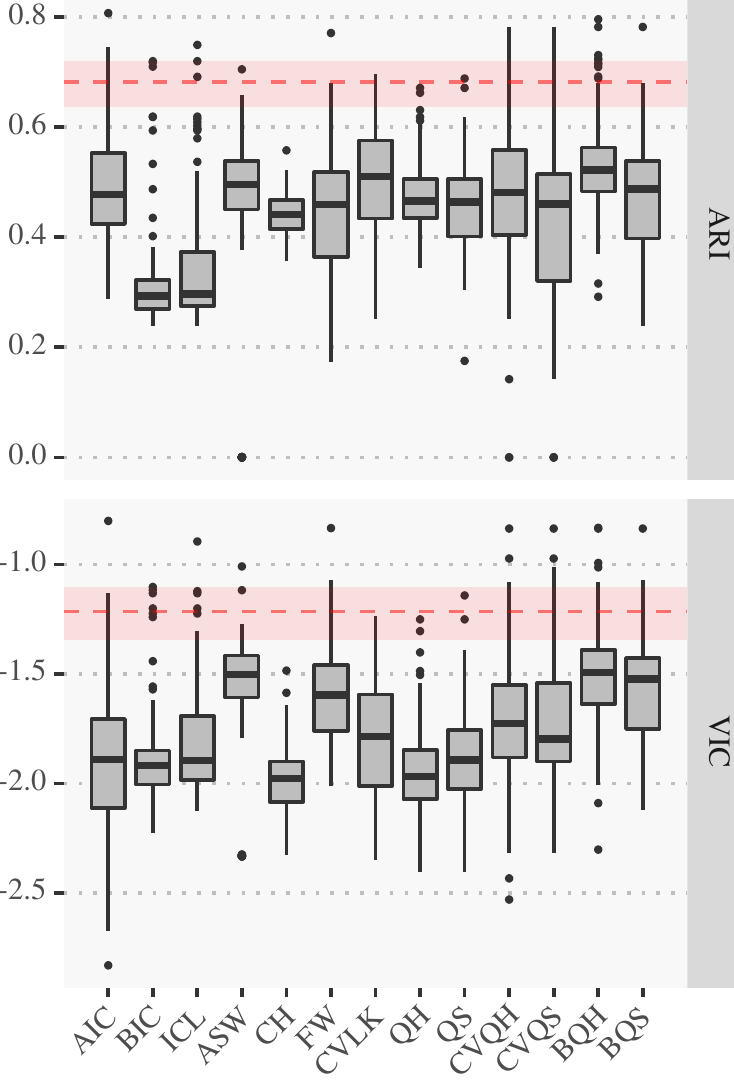}
        }
    \subfloat[Uniform\label{fig:simulated_results:uniform}]{
        \includegraphics[width=.30\textwidth, keepaspectratio]{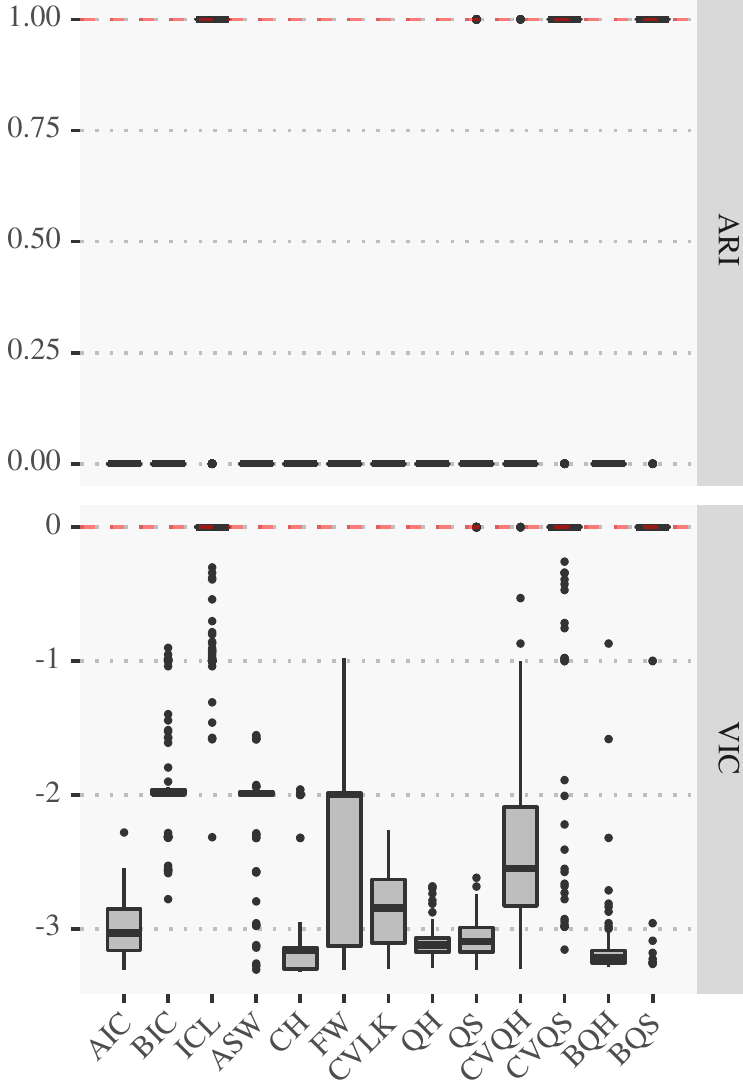}
    }
\caption{Boxplots of Monte Carlo distribution of ARI and VIC performance measures.  The dashed line in each plot indicates the average best feasible ARI and VIC available in $\mathcal{M}$ (achieved by the \textsc{best ari} and \textsc{best vic} solutions), together with bands (shaded area) ranging from the first to the third quartiles of the scores' empirical distributions.
}\label{fig:simulated_results}
\end{figure}

For each of the 5 simulated designs, we simulate 100 independent data sets from the DGP, and run the model selection experiment on each, in a Monte Carlo (MC) fashion.
In this section, we report results for the MC experiments, aggregated for each sample design.
Due to the computational complexity of this exercise, we limit the bootstrap replicate to $B=100$ for all the experiments.
Results for all the designs are summarized in \figref{fig:simulated_results}, showing boxplots of the Monte Carlo distribution of the ARI and the VIC, and \tabref{tab:fullresults}, bottom sub-table.
The ARI and the VIC compare the selected partition to the ground truth previously defined.

\begin{description}
  \item[Pentagon5.] All methods do well.
  The AIC and the BIC selected 5 groups in roughly 50\% of the experiments.
  This confirms the tendency of such criteria to recover the underlying true DGP rather than the clustering structure.
  In fact, for this DGP, 3 groups are what one would suggest by visual inspection of the scatter plot in \figref{fig:simdata_example}.
  The other well-performing criteria typically prefer the 3-clusters solution.
  BQH, BQS, ICL, ASW, CH and FW also excelled for the stability of the results.
  \item[T52D.] The top performers are BIC, ICL, ASW, BQH and BQS.
  All of these criteria fit 5 clusters on average, selecting partitions of $\mathcal{M}$ that are close to the best available in the set.
  This is not surprising given the strong between-cluster separation.
  BQH and BQS do marginally better, showing the most stable selection.
  It is worth noting that ASW does well, even if it is not specifically designed to handle DGPs of this type.
  Whenever clusters are well separated, the intuition is that a distance-based index like ASW can retrieve the true clusters if it uses an appropriate metric.
  
  \item[T510D.] The addition of uninformative noisy features in T510D changes the results dramatically: only ICL and BQS maintain excellent performances, with BQS doing slightly better overall in terms of ARI and VIC.
      In our experiments (see \sisec{5.3})  ICL never selects solutions having a number of groups extremely different from that of the ground truth partition, in contrasts with BQS, which selects $K>7$ groups in rare cases.
  However, it is worth noting that information-type criteria select over a smaller subset of $\mathcal{M}$, not including \methodset{k-means}, \methodset{k-medoids} and \methodset{rimle} solutions, which may produce less variability in the selection.
  
\item[Flower2.] This is probably the most challenging case.
  The best feasible solutions in $\mathcal{M}$ achieve modest levels of average ARI and VIC.
  A 5 cluster solution achieves the best ARI and VIC roughly 77\% of the time, and the methods identifying 5 clusters more often are ASW, FW, BQH, and BQS. The latter two more closely match the frequency with which \textsc{best ari} and \textsc{best vic} selects 5 groups.
  BQH does only marginally better than its competitors in terms of ARI and VIC, but we can see that the performance of BQS, ASW, and FW are equally good.
  
  \item[Uniform.] this sampling design is more of a clear-cut: ICL, BQS, and CVQS are all able to correctly identify no clustering structure.
  In this case, the clear winner is BQS, selecting a single cluster in 96\% of the replicates compared to the 85\% of CVQS and 77\% of ICL.
  All the other methods wrongly identify clustering structures in the data (note that FW can not be directly used to handle the unclustered case).
  Here, we can see that the AIC looks for the best distribution fit rather than accommodating clustered regions.
  Indeed, AIC prefers a large number of mixture components to fit the highly unstructured uniform scatter.
  The BIC mitigates this tendency, but it is not enough.
  It is also remarkable to see the difference between the top performer, BQS, and its close cousin BQH failing miserably.
  The explanation of such a bad performance is the tendency of the hard scoring approach to split close groups of points (as shown in \secref{sec:qs:cluster_boundaries}).
  In this case, with a small $n=300$, the uniform DGP (see \figref{fig:simdata_example}) creates many small groups of data points with minimal within-distance, which encourages the hard score to identify many groups.
\end{description}

Overall, experiments on simulated data confirms the analysis on real data.
BQS has shown a best or second-best performance in all cases, also yielding better results overall than BQH, proving to be more robust to diverse settings than the latter.
ICL is undoubtedly the strongest competitor, although its performance is far from optimal on some occasions.
Method-independent criteria like the ASW and the CH, routinely used by practitioners, sometimes completely miss the underlying structure.
However, they selected meaningful solutions occasionally, depending on the underlying clustering structure.
As already noted for the real data sets, the in-sample estimates QH and QS show a strong selection bias and variance for all data sets.
All the methods based on cross-validation exhibit disappointing performances.

%Figures for simulated data results

\section{Conclusions and final remarks}\label{sec:conclusions}
We introduce a unifying framework for treating the problem of cluster selection and validation in the context of clusters generated from elliptic-symmetric families.
Within this framework, we propose a novel method for selecting an appropriate clustering for a given data set over a set of candidate partitions (potentially obtained with any clustering method).
An extensive comparative experimental study shows that the proposed methodology improves upon popular existing alternatives.
In particular, the smooth score criterion with resampling (BQS) consistently provides the best or second-best results in all the considered settings and is thus the authors' advocated criterion.
Due to the resampling-refit strategy, the method can be computationally demanding in some circumstances, but this drawback is offset by improved performances and a visualization method that can be used to inspect for unnecessary complexity of the solutions.

%%%%%%%%%%%%%%%%%%%%%%%%%%%%%%%%%%%%%%%%%%%%%%
%% Single Appendix:                         %%
%%%%%%%%%%%%%%%%%%%%%%%%%%%%%%%%%%%%%%%%%%%%%%
%\begin{appendix}
%\section*{???}%% if no title is needed, leave empty \section*{}.
%\end{appendix}
%%%%%%%%%%%%%%%%%%%%%%%%%%%%%%%%%%%%%%%%%%%%%%
%% Multiple Appendixes:                     %%
%%%%%%%%%%%%%%%%%%%%%%%%%%%%%%%%%%%%%%%%%%%%%%
\begin{appendix}
     \section{Proofs of statements}\label{sec:proofs}

\paragraph*{Proof of Proposition~\ref{stm:coherence_qsp}}
The problem is the analogue of showing the optimality of the Bayes classifier. However, this is conceptually different due to the unsupervised nature of the clustering problem, where a natural notion of loss does not exist.
Consider any partition $\set{A_k,\, k=\otK}$, then
\begin{equation}
\begin{aligned}\label{eq:int_pi_fk}
\prob{\bigcup_{k=1}^K \set{Z_k=1 \cap X \in A_k}} %%
&= \sum_{k=1}^{K} \prob{Z_k=1}\prob{X \in A_k \given Z_k=1}, \\ %%
&= \sum_{k=1}^{K} \int_{A_k} \pi_k f(\bx ; \bmu_k, \bSigma_k)d\bx.
\end{aligned}
\end{equation}
In order to maximize \eqref{eq:int_pi_fk} it suffices to choose the partition $\set{A_k^*,\; k = \otK}$
\[
A^*_k=\set{\bx \in \setR^p :\; %
\pi_k f(\bx ; \bmu_k, \bSigma_k) = %
\max_{1 \leq j \leq K} \; \pi_j f(\bx ; \bmu_j, \bSigma_j)}.
\]
Under \cndGaussian, $\pi_kf(\bx; \bmu_k, \bSigma_k) = \pi_k\phi(\bx; \bmu_k, \bSigma_k)$, and it is immediate to see that $A^*_k$ coincides with $Q_k$, proving \eqref{eq:pr_best_partition}. %\eqref{eq:pr_best_partition}-(\mpf).
Denote $\delta_k =\tra{(\bx-\bmu_k)}\inv{\bSigma_k}(\bx-\bmu_k)$.
 Since both $g(t)$ and $\exp(-t/2)$ are monotonically decreasing for $t \in [0, +\infty)$, under \cndEqualPrecision, for any $\bx \in \setR^p$, %
\[
\begin{aligned}
\pi_k f(\bx ; \bmu_k, \bSigma_k) \geq \max_{1 \leq j \leq K} \pi_j f(\bx ; \bmu_j, \bSigma_j) %
&\iff g(\delta_k) \geq \max_{1 \leq j \leq K} \{ g(\delta_j)\}, \\
&\iff \exp(-\delta_k/2) \geq \max_{1 \leq j \leq K} \{ \exp(-\delta_j/2)\}, \\
&\iff \qs(\bx, \bthetam_{k}) \geq \max_{1 \leq j \leq K}\set{\qs(\bx, \bthetam_{j})}.
\end{aligned}
\]
Which means that $A^*_k = Q_k(\btheta) \in \mathcal{Q}(\btheta)$ for all $k \in \otK$.
 \qed

\paragraph*{Proof of Proposition~\ref{stm:H_vs_L}} %
First, note that 
\[
 \qs(\bx, \bthetam_{k}) = c + \log(\pi_k^{(m)} \phi(\bx; \bmu_k^{(m)}, \bSigma_k^{(m)})),
\]
where $c=p\log(\sqrt{2\pi})/2$, with $\pi$ here being the mathematical constant.
Since $\sum_{k=1}^{K(\bthetam)} \int_{Q_k(\bthetam)} \allowbreak c \; dF = c$, then 
\begin{equation}\label{eq:proof_H_decomp}
H(\bthetam) = %
 c
 + \sum_{k=1}^{K(\bthetam)} \int_{Q_k(\bthetam)} \log(\pi_k^{(m)})dF %
+ \sum_{k=1}^{K(\bthetam)} \int_{Q_k(\bthetam)} \log(\phi(\bx; \bmu_k^{(m)}, \bSigma_k^{(m)}))dF.
\end{equation}
Using the expression for $L(\btheta)$ from \eqref{eq:LH}, we can write 
\[
\sum_{k=1}^{K(\bthetam)} \int_{Q_k(\bthetam)} \log(\pi_k^{(m)})dF = %
L(\bthetam) %
- \sum_{k=1}^{K(\bthetam)} \int_{Q_k(\bthetam)} \log(f(\bx; \bmu_k^{(m)}, \bSigma_k^{(m)}))dF.
\]
Replace the right-hand side of the previous equation into \eqref{eq:proof_H_decomp} to obtain \eqref{eq:H_L_penalty}.
Under \cndInfDiffLoglik, for any choice of $\bthetam$ and $k$,
\[
\int_{Q_k(\bthetam)} %
\log \left( \frac{f(\bx; \bmu_k^{(m)}, \bSigma_k^{(m)})}{\phi(\bx; \bmu_k^{(m)}, \bSigma_k^{(m)})}
\right) dF \geq 0,
\]
which proves that $\Lambda(\bthetam) \geq 0$.
\qed

\paragraph*{Proof of Proposition~\ref{stm:T_vs_dkl_plus_ent}}
The posterior weights \eqref{eq:omega_posteriors} under the Gaussian group-conditional model coincide with the smooth score weights, in fact 
\[
\omega_{\phi,k}(\bx; \bthetam) %
\; = \; %
\frac{\pi_k^{(m)} \phi(\bx; \bmu_k^{(m)}, \bSigma_k^{(m)})}{\sum_{k=1}^{K(\bthetam)} \pi_k^{(m)} \phi(\bx; \bmu_k^{(m)}, \bSigma_k^{(m)})} %
\; = \; %
\tau_k(\bx; \bthetam)
\]
for all $k$.
Use the same arguments as in the proof of Proposition~\ref{stm:H_vs_L} and write 
\begin{equation}\label{eq:S_vs_gauss_1}
T(\bthetam) = c + \sum_{k=1}^{K(\bthetam)} \int \omega_{\phi,k}(\bx; \bthetam)\log(\pi_k^{(m)} \phi(\bx; \bmu_k^{(m)}, \bSigma_k^{(m)}))dF,
\end{equation}
for an appropriate constant $c$ that does not depend on $\bthetam$.
Since
$\sum_{k=1}^{K(\bthetam)} \allowbreak \omega_{\phi,k}(\bx; \bthetam) \allowbreak = 1$, the right-hand-side of \eqref{eq:S_vs_gauss_1}, neglecting the constant term, can be expressed as 
\begin{equation}\label{eq:S_vs_gauss_2} 
\sum_{k=1}^{K(\bthetam)} \int \omega_{\phi,k}(\bx; \bthetam)\log(\pi_k^{(m)} \phi(\bx; \bmu_k^{(m)}, \bSigma_k^{(m)}))dF = A(\bthetam) - B(\bthetam),\\
\end{equation}
where
\begin{equation}\label{eq:proof_A_theta}
A(\bthetam) = \int \log(\psi_{\phi}(\bx; \bthetam))dF = \int \log \left( \sum_{k=1}^{K(\bthetam)} \pi_k^{(m)} \phi(\bx; \bmu_k^{(m)}, \bSigma_k^{(m)})\right)dF,
\end{equation}
and
\begin{equation}\label{eq:proof_B_theta}
B(\bthetam) = -\sum_{k=1}^{K(\bthetam)} \int \omega_{\phi,k}(\bx; \bthetam) \log \omega_{\phi,k}(\bx; \bthetam)dF.
\end{equation}
The term $A(\bthetam)$ is the expected log-likelihood under the Gaussian mixture model.
Since $f_0$ by assumption is the density of $F$, then 
\[
A(\bthetam) = -\dkl(f_0 \,||\, \psi(\cdot; \bthetam)) + \int \log(f_0(\bx))dF ,
\]
where the last integral depends only on unknown population objects, and therefore does not depend on $\bthetam$.
\eqref{eq:proof_B_theta} is the expectation under $F$ of 
\begin{equation*}
\ent_\phi(Z \given X; \bthetam) = -\sum_{k=1}^{K(\bthetam)} \omega_{\phi,k}(X; \bthetam) \log\left( \omega_{\phi,k}(X; \bthetam) \right) .
\end{equation*}
We can now conclude that
\[
\begin{aligned}
\argmax_{1 \leq m \leq M} \; T(\bthetam)
 &= \argmax_{1 \leq m \leq M} \; A(\bthetam) - B(\bthetam), \\
 &= \argmin_{1 \leq m \leq M} \; \dkl(f_0 \,||\, \psi(\cdot; \bthetam)) + \E[F]{\ent_\phi(\bthetam)}.
 \end{aligned}
\]
The latter proves the desired result \eqref{eq:T_vs_dkl_entropy}.
\qed

\end{appendix}

%%%%%%%%%%%%%%%%%%%%%%%%%%%%%%%%%%%%%%%%%%%%%%
%% Support information, if any,             %%
%% should be provided in the                %%
%% Acknowledgements section.                %%
%%%%%%%%%%%%%%%%%%%%%%%%%%%%%%%%%%%%%%%%%%%%%%
%\begin{acks}[Acknowledgments]
% The authors would like to thank ...
%\end{acks}
%%%%%%%%%%%%%%%%%%%%%%%%%%%%%%%%%%%%%%%%%%%%%%
%% Funding information, if any,             %%
%% should be provided in the                %%
%% funding section.                         %%
%%%%%%%%%%%%%%%%%%%%%%%%%%%%%%%%%%%%%%%%%%%%%%
%\begin{funding}
% The first author was supported by ...
%
% The second author was supported in part by ...
%\end{funding}

%%%%%%%%%%%%%%%%%%%%%%%%%%%%%%%%%%%%%%%%%%%%%%
%% Supplementary Material, including data   %%
%% sets and code, should be provided in     %%
%% {supplement} environment with title      %%
%% and short description. It cannot be      %%
%% available exclusively as external link.  %%
%% All Supplementary Material must be       %%
%% available to the reader on Project       %%
%% Euclid with the published article.       %%
%%%%%%%%%%%%%%%%%%%%%%%%%%%%%%%%%%%%%%%%%%%%%%
\begin{supplement}
\stitle{Supplementary Material to ``Selecting the number of clusters, clustering models, and algorithms. A unifying approach based on the quadratic discriminant score''}
\sdescription{The Supplementary Material contains further motivation for the selection problem (Section~S2) and results that link the proposed quadratic scores to the sampling log-likelihood (Section~S3).
It also collects further descriptions for the experimental setup (Section~S1), details on real data and simulation designs (Section~S4), and additional experimental results (Section~S5).}
\end{supplement}

%%  imsart-nameyear.bst  will be used to                   %%
%%  MR numbers will be added by VTeX.                      %%
%%                                                         %%
%%  Use \cite{...} to cite references in text.             %%
%%                                                         %%
%%%%%%%%%%%%%%%%%%%%%%%%%%%%%%%%%%%%%%%%%%%%%%%%%%%%%%%%%%%%%

%% if your bibliography is in bibtex format, uncomment commands:
\bibliographystyle{imsart-nameyear} % Style BST file
\bibliography{main}       % Bibliography file (usually '*.bib')

%% or include bibliography directly:
% \begin{thebibliography}{}
% \bibitem[\protect\citeauthoryear{???}{???}]{b1}
% \end{thebibliography}

%% attach pdf 
%% ----------------------------------------------
\newpage


\section*{Supplementary material}

Supplementary Material:

# SELECTING THE NUMBER OF CLUSTERS, CLUSTERING MODELS, AND ALGORITHMS. A UNIFYING APPROACH BASED ON THE QUADRATIC DISCRIMINANT SCORE

Luca Coraggio*     Pietro Coretto†

This document presents additional discussions, technical details, proofs of formal statements and additional information concerning the numerical experiments. Other supplementary material includes data and software currently available at the following link: http://bit.do/coraggio-coretto-bqs

# Contents

---

*University of Naples "Federico II" (Italy) – E-mail: luca.coraggio@unina.it

†University of Salerno (Italy) – E-mail: pcoretto@unisa.it

# S1 Methods

In this section, we include additional details on clustering methods and cluster validation criteria.

## S1.1 Clustering methods

We aggregate the clustering methods based on the implementing software. In addition to the data set, each method requires additional inputs. These may correspond to the method's hyper-parameters, model restrictions, modeling alternatives, or even controls of the algorithm's implementation. The following list includes the main methods used in the following numerical experiments and examples.

K-MEANS
: classical partitioning method based on Euclidean distances that solves the "K-Means" optimization problem. Algorithm: Hartigan and Wong (1979). Software: R-base package of R Core Team (2021). Inputs: the number of clusters $K$.

K-MEDOIDS
: dissimilarity based partitioning method that solves the "K-medoid" problem introduced by Kaufman and Rousseeuw (1990). All the K-medoids solutions are obtained using the Euclidean dissimilarity. Algorithm: PAM by of Kaufman and Rousseeuw (1987). Software: cluster of Maechler et al. (2019). Inputs: the number of clusters $K$.

MCLUST
: model-based clustering (MBC) method that uses Maximum Likelihood (ML) estimation of finite Gaussian mixture models with covariance matrix restrictions. Algorithm: Expectation-Maximization (EM) algorithm (see Celeux and Govaert, 1995). Software: mclust of Scrucca et al. (2016). Inputs: the number of clusters $K$ and one of the 14 covariance matrix models (see details below).

RIMLE
: MBC method that uses ML estimation of finite Gaussian mixture models with covariance matrix regularization based on the so-called eigen-ratio constraint (Garcìa-Escudero et al., 2015; Coretto and Hennig, 2016). Algorithm: Expectation-Conditional-Maximization (ECM) algorithm of Coretto and Hennig (2017). Software: otrimle of Coretto and Hennig (2021). Inputs: the number of clusters $K$, the egein-ratio $\gamma = [1, +\infty)$ and the initialization strategy (see details below).

EMMIX
: MBC method that uses ML estimation of finite mixture of Student-t and Skew Student-t distributions (Peel and McLachlan, 2000; Wang et al., 2009) with covariance matrix restrictions. Algorithm: EM and ECM algorithms of Wang et al. (2009). Software: EMMIXskew of Wang et al. (2018). Inputs: the number of clusters $K$, the specific mixture component model (Student-t or Skew Student-t) and one of the 5 the covariance matrix models (see details below).

**Connections between partitioning and MBC methods**   The two partitioning methods chosen here are very similar and require minimal tuning. K-MEDOIDS with the Euclidean dissimilarity differs from K-MEANS only in the way it represents cluster centers. The two partitioning methods target similar cluster concepts that are the simplest in the list above. Although the K-MEANS method was not originally derived from model assumptions, the K-means solution coincides with the sample ML estimator under a fixed partition model where groups are generated by spherical Gaussian distributions (Pollard, 1981). Therefore, the two partitioning methods would be equivalent to some of the methods described below under strong restrictions on the parameter space.

**Connections and differences between MBC methods**   All the previous MBC methods rely on the ML estimation for mixture models of Gaussian or Student-t distributions. The Skew Student-t model contains the Student-t model as a special case, and the latter includes the Gaussian model. Even if some of the previous methods are derived from the same model (e.g. RIMLE and MCLUST), there are significant differences between them in terms of the required inputs. The main differences are mainly due to how the ML procedure takes into account two major problems: the unboundedness of the likelihood function that may be caused by degenerating mixture components' scatter matrices (Kiefer and Wolfowitz, 1956); the excessive number of free parameters caused by the large number of parameters needed to fill each of the components' scatter matrices.

Mclust software does not allow the user to control the regularization of the covariance parameters. The underlying EM algorithm includes a step where a lower bound for the minimum determinant across the covariance matrices is determined from the data using the Bayes regularization method proposed in Fraley and Raftery (2007). On the other hand, Mclust allows the user to specify a covariance matrix model from a list of 14 available models proposed in Celeux and Govaert (1995). The covariance matrix of the $k$-th mixture component is decomposed as

$$\boldsymbol{\Sigma}_k = \lambda_k \boldsymbol{D}_k \boldsymbol{A}_k \boldsymbol{D}_k^{\mathsf{T}}, \tag{S1}$$

where $\lambda_k = \det(\boldsymbol{\Sigma}_k)^{1/p}$, $\boldsymbol{A}_k$ is the diagonal matrix of the normalized eigenvalues sorted in decreasing order and $\boldsymbol{D}_k$ is the matrix of the corresponding eigenvectors. The factors of the decomposition control geometrical aspects of the $k$-th within-cluster scatter: volume, orientation and shape. The parametrizations available in mclust are reported in Table S1; they range from the strongest restriction $\boldsymbol{\Sigma}_k = \lambda \boldsymbol{I}$ for all $k = 1, 2, \ldots, K$, implying the same spherical scatter for all groups, to the unrestricted case where all members of the decomposition (S1) vary unrestricted across clusters. mclust does not allow to control the initial partition.

As an alternative way of restricting the scatter parameters of mixtures of Gaussians, we introduce solutions computed using the rimle function from the otrimle software of Coretto and Hennig (2021). The otrimle software performs robust cluster analysis for data sets affected by outliers and noise, and it implements the methods developed in Coretto and Hennig (2016) and Coretto and Hennig (2017). It can also be used to fit plain Gaussian mixture[1] where the unboundedness of the likelihood function is regularized based on the so-called eigen-ratio constraint (ERC). The ERC bounds the maximum spread between all eigenvalues of all mixture components' covariance matrices by a fixed constant $\gamma \geq 1$, that

---

[1] This can be done by setting the parameter $\delta = 0$ in Coretto and Hennig (2017) (corresponding to `logicd = -Inf` into the rimle software).

Table S1: Covariance matrix models available in Mclust. If a member of the right-hand side of (S1) has a subscript $k$, it means that it is allowed to vary across clusters. Elements of $\mathbf{\Sigma}_k$ not having a subscript $k$ are restricted to be equal across the $K$ groups. $p$ is data dimensionality. $\alpha$ counts the number of parameters other than those in covariance matrices, i.e. means $\mu$'s and mixing proportions $\pi$'s; $\alpha = Kp + K - 1$; if mixing proportions are restricted to be equal, $\alpha = Kp$. $\beta$ counts the number of parameters in a covariance matrix; $\beta = p(p+1)/2$.

| $\mathbf{\Sigma}_k$ | # parameters to estimate | Clusters geometry | Clusters Volume | Clusters Shape | Clusters Orientation | Config. Short name |
|---|---|---|---|---|---|---|
| $\lambda \mathbf{I}$ | $\alpha + 1$ | Spherical | Equal | Equal | – | EII |
| $\lambda_k \mathbf{I}$ | $\alpha + p$ | Spherical | Variable | Equal | – | VII |
| $\lambda \mathbf{A}$ | $\alpha + p$ | Diagonal | Equal | Equal | Coord. Axes | EEI |
| $\lambda_k \mathbf{A}$ | $\alpha + p + K - 1$ | Diagonal | Variable | Equal | Coord. Axes | VEI |
| $\lambda \mathbf{A}_k$ | $\alpha + Kp - K + 1$ | Diagonal | Equal | Variable | Coord. Axes | EVI |
| $\lambda_k \mathbf{A}_k$ | $\alpha + Kp$ | Diagonal | Variable | Variable | Coord. Axes | VVI |
| $\lambda \mathbf{D}\mathbf{A}\mathbf{D}^\mathsf{T}$ | $\alpha + \beta$ | Ellipsoidal | Equal | Equal | Equal | EEE |
| $\lambda \mathbf{D}\mathbf{A}_k\mathbf{D}^\mathsf{T}$ | $\alpha + \beta + K - 1$ | Ellipsoidal | Equal | Variable | Equal | EVE |
| $\lambda_k \mathbf{D}\mathbf{A}\mathbf{D}^\mathsf{T}$ | $\alpha + \beta + (K-1)(p-1)$ | Ellipsoidal | Variable | Equal | Equal | VEE |
| $\lambda_k \mathbf{D}\mathbf{A}_k\mathbf{D}^\mathsf{T}$ | $\alpha + \beta + (K-1)p$ | Ellipsoidal | Variable | Variable | Equal | VVE |
| $\lambda \mathbf{D}_k\mathbf{A}\mathbf{D}_k^\mathsf{T}$ | $\alpha + K\beta - (K-1)p$ | Ellipsoidal | Equal | Equal | Variable | EEV |
| $\lambda_k \mathbf{D}_k\mathbf{A}\mathbf{D}_k^\mathsf{T}$ | $\alpha + K\beta - (K-1)(p-1)$ | Ellipsoidal | Variable | Equal | Variable | VEV |
| $\lambda \mathbf{D}_k\mathbf{A}_k\mathbf{D}_k^\mathsf{T}$ | $\alpha + K\beta - (K-1)$ | Ellipsoidal | Equal | Variable | Variable | EVV |
| $\lambda_k \mathbf{D}_k\mathbf{A}_k\mathbf{D}_k^\mathsf{T}$ | $\alpha + K\beta$ | Ellipsoidal | Variable | Variable | Variable | VVV |

is

$$\max_{s,t=1,2,\ldots,K} \frac{\lambda_{\max}(\mathbf{\Sigma}_s)}{\lambda_{\min}(\mathbf{\Sigma}_t)} \leq \gamma, \tag{S2}$$

where $\lambda_{\max}(\mathbf{\Sigma}_k)$ and $\lambda_{\min}(\mathbf{\Sigma}_k)$ are the minimum and the maximum eigenvalue of $\mathbf{\Sigma}_k$, respectively. The idea of the ERC dates back to the seminal contribution of Hathaway (1985) (for a recent overview, see García-Escudero et al., 2017). For any $\gamma \in [1, +\infty)$ the resulting estimated covariance matrices are guaranteed to be positive definite whatever large are $p$ and $K$ compared to $n$. As found in Coretto and Hennig (2017), although the ERC fixes an otherwise ill-posed optimization problem, it strongly affects the complexity of the underlying model, significantly impacting the final clustering. $\gamma$ bounds the relative discrepancy between clusters' scatters: $\gamma$ close to one is the same as imposing spherical groups; $\gamma \to +\infty$ completely frees the clusters' shape, allowing to discover more complex structures. In some sense, the ERC tunes the complexity of the covariance structure as the 14 mclust parametrizations do. Still, the tuning provided by the ERC is *continuous* in the sense that increasing $\gamma$ enriches the model structure continuously. For any fixed $\gamma$, the number of parameters to be estimated does not change, but different values of $\gamma$ correspond to the clusters' different levels of complexity. Additional to $K$ and $\gamma$, the rimle implementation allows to control the initialization strategy. The three types of initial partitions are based on K-means, K-medoids partitioning, and agglomerative hierarchical clustering based on likelihood criteria of Fraley (1998).

The Student-t and Skew Student-t models at the basis of EMMIX are more flexible alternatives to the Gaussian prototype at the heart of both MCLUST and RIMLE. Therefore, EMMIX is introduced to assess whether the selection methods under study tempt to select configurations corresponding to models that have more chances to overfit the data because of their extra flexibility. Similar to Mclust, EMMIXskew allows choosing between a smaller set of 5 covariance models: (1) homoscedastic groups, that is $\mathbf{\Sigma}_k = \mathbf{\Sigma}$ for all $k = 1, 2, \ldots, K$; (2) homoscedastic diagonal model: $\mathbf{\Sigma}_k = \text{diag}(\sigma_1^2, \sigma_2^2, \ldots, \sigma_p^2)$ for all $k = 1, 2, \ldots,$

$K$; (3) fully heteroskedastic model (called general variance in EMMIXskew): $\Sigma_k$ are free to vary across groups; (4) heteroskedastic diagonal model: $\Sigma_k = \text{diag}(\sigma_{k1}^2, \sigma_{k2}^2, \ldots, \sigma_{kp}^2)$; (5) heteroskedastic spherical model: $\Sigma_k = \sigma_k^2 I_p$. For the skew version of the Student-t model, these covariances are combined with a skew parameter to obtain non-elliptical shapes. As for Mclust, EMMIXskew also applies covariance regularization to prevent spurious degenerate solutions, but this is done internally and the user cannot control it.

## S1.2 Selection methods

The selection methods under study are denoted as follows.

| | |
|---|---|
| ASW: | Average Silhouettes Width criterion; |
| CH: | Caliński-Harabasz criterion; |
| AIC: | Akaike Information Criterion; |
| BIC: | Bayesian Information Criterion; |
| ICL: | Integrated Complete-data Likelihood criterion; |
| FW: | bootstrap stability selection of Fang and Wang (2012); |
| CVLK: | cross-validation of likelihood risk of Smyth (2000); |
| QH: | in-sample Quadratic Hard Score criterion; |
| QS: | in-sample Quadratic Smooth Score criterion; |
| CVQH: | cross-validation with the Quadratic Hard Score criterion; |
| CVQS: | cross-validation with the Quadratic Smooth Score criterion; |
| BQH: | bootstrap selection with the Quadratic Hard Score criterion; |
| BQS: | bootstrap selection with the Quadratic Smooth Score criterion; |

In Table S2 we classify these selection methods into 4 groups. Each of the previous methods is set to maximize the corresponding criterion over the members of $\mathcal{M}$, where $\mathcal{M}$ is the set of candidate solutions. In case of ties, we apply random selection between equally good members of $\mathcal{M}$. We recall that not all selection methods in the list above can be computed for all members of $\mathcal{M}$: AIC, BIC, ICL can be computed only for MCLUST and EMMIX, and CVLK for MCLUST, EMMIX and RIMLE. Therefore, some of the selection methods are compared on a subset of the 440 members of $\mathcal{M}$.

Table S2: Selection methods classified as: *method-independent* vs. *method-dependent* criteria; methods using *in-sample* estimates vs. methods that use *resampling* of the original data.

| | Method-independent | Method-dependent |
|---|---|---|
| In-sample | CH, ASW, QH, QS | AIC, BIC, ICL |
| Resampling | FW, CVQH, CVQS BQH, BQS | CVLK |

**Method-independent criteria using in-sample information** Many validation methods look for partitions minimizing within-clusters dissimilarity assumed as the driving notion of internal homogeneity. Let $D(\boldsymbol{x}_i, \boldsymbol{x}_j)$ be the pairwise dissimilarity between units $i$ and $j$. Caliński and Harabasz (1974) proposed to measure the quality of clustering achieved

5

by a partition $\mathcal{G}_K$ in terms of the ratio

$$\text{CH}(\mathcal{G}_K) := \frac{\boldsymbol{B}(\mathcal{G}_K)(n-K)}{\boldsymbol{W}(\mathcal{G}_K)(K-1)} \tag{S3}$$

where $\boldsymbol{B}(\cdot)$ and $\boldsymbol{W}(\cdot)$ are respectively the between and within point scatter measured as

$$\boldsymbol{W}(\mathcal{G}_K) := \sum_{k=1}^{K} \frac{1}{\#\{G_k\}} \sum_{i,j \in G_k} D(\boldsymbol{x}_i, \boldsymbol{x}_j), \quad \boldsymbol{B}(\mathcal{G}_K) := \frac{1}{n} \sum_{i,j=1}^{n} D(\boldsymbol{x}_i, \boldsymbol{x}_j) - \boldsymbol{W}(\mathcal{G}_K).$$

Calculation of $\text{CH}(\cdot)$ only requires the clusters' labels and dissimilarities computed on the observed data set. Originally, (S3) was proposed for $D(\cdot)$ being the squared Euclidean distance within the k-means framework, but later Milligan and Cooper (1985) showed that the $\text{CH}(\cdot)$ criteria performed well for more general dissimilarity measures. It is preferred a clustering with large $\text{CH}(\cdot)$. The ratio $\boldsymbol{W}(\mathcal{G}_K)/\boldsymbol{B}(\mathcal{G}_K)$ can be improved simply by increasing $K$: this would fit the data more locally, reducing the within dissimilarity. The previous mechanism is tamed by introducing the correction factor $(n-K)/(K-1)$.

The *Average Silhouette Width* criteria (ASW) of Rousseeuw and Kaufman (1990) is another popular index. Let $i \in G_k$ and define

$$a(i) := \frac{1}{\#\{G_k\}-1} \sum_{j \in G_k \setminus \{i\}} D(i,j), \quad b(i) := \min_{G_k: i \notin G_k} \left\{ \frac{1}{\#\{G_k\}} \sum_{j \in G_k} D(i,j) \right\}.$$

The ASW index averages the point's *silhouette* $s(i)$ as follows

$$\text{ASW}(\mathcal{G}_K) = \frac{1}{n} \sum_{i=1}^{n} s(i), \quad \text{where } s(i) = \frac{b(i) - a(i)}{\max\{a(i), b(i)\}}. \tag{S4}$$

The silhouette $s(\cdot)$ measures how tightly a point is connected to its cluster compared to the remaining clusters. One desires partitions with large $\text{ASW}(\cdot)$. As for (S3), the computation of (S4) only requires clusters' label and a reference dissimilarity notion.

We use ASW and CH as in-sample criteria by estimating the clustering solution $\mathcal{G}_k$ on the full sample, and re-using the same sample to compute both criteria; we set $D$ to be the Euclidean distance.

**Method-independent criteria using resampling** The stability criterion proposed by Fang and Wang (2012) (FW) performs method-independent selection based on bootstrap resampling. FW looks for stable clustering solutions, i.e. partitions that are robust against changes due to the randomness in the sample, and this is assessed using bootstrap resampling. Multiple independent pairs of bootstrap samples are drawn. For each pair, the given clustering method, $m$, is fit on both resamples, obtaining two clustering solutions; then, a clustering distance is computed using the assignment of points from the original sample into $K(m)$ clusters, made by the two solutions. The distance is averaged across the pairs to obtain the clustering instability. Algorithm S1 shows the exact implementation. FW can not be used for configurations with $K(m) = 1$ because, in this case, it would always obtain a trivial instability score of 0. Therefore, FW is not computed for members of $m \in \mathcal{M}$ with $K(m) = 1$. For the sake of consistency with the other selection methods, we report

the negative clustering instability so that we maximize $-FW$; however, to ease notation, we drop the minus sign.

---

**Algorithm S1** FW: Stability criterion of Fang and Wang (2012)

---

**input:** observed sample $\mathbb{X}_n$, integer $B$.
**output:** $FW$.

**for** $b = 1, \ldots, B$ **do**

(step 1) $(\mathbb{X}_n^*, \mathbb{Y}_n^*) \leftarrow$ couple of independent, non-parametric bootstrap resamples form $\mathbb{X}_n$.

(step 2) Fit clustering on bootstrap resamples, obtaining mapping functions ($\psi_{\mathbb{X}} : \mathcal{X} \to \{1, \ldots, K(m)\}$):

$$\psi_{\mathbb{X}_n^*} \leftarrow \text{Clust}(\mathbb{X}_n^*); \qquad \psi_{\mathbb{Y}_n^*} \leftarrow \text{Clust}(\mathbb{Y}_n^*)$$

(step 3) Compute clustering distance (on original sample $\mathbb{X}_n$):

$$d_b \leftarrow \frac{1}{n^2} \sum_{i=1}^{n} \sum_{j=1}^{n} \left| \mathbb{I}\left\{\psi_{\mathbb{X}_n^*}(x_i) = \psi_{\mathbb{X}_n^*}(x_j)\right\} - \mathbb{I}\left\{\psi_{\mathbb{Y}_n^*}(x_i) = \psi_{\mathbb{Y}_n^*}(x_j)\right\} \right|$$

**end for**

(step 4) Obtain clustering instability

$$FW \leftarrow \frac{1}{B} \sum_{b=1}^{B} d_b$$

---

**Method-dependent criteria using in-sample information** The first group of method-dependent criteria are those based on information-type criteria derived within the MBC setting. These would be better called "model-dependent" methods, because in reality they are strictly connected to an underlying model rather than a fitting method. In the following definitions, $\boldsymbol{\theta}$ is meant to be the unknown parameter vector of a mixture model; $\text{lik}_{f,n}(\cdot)$ and $\text{clik}_{f,n}(\cdot)$ are the log-likelihood and the so-called complete data log-likelihood functions under a certain mixture model, as defined later in (S6).

The classical estimators of the AIC (Akaike, 1973) and the BIC (Schwarz, 1978) are defined as

$$\text{AIC} = 2\,\text{lik}_{f,n}(\boldsymbol{\theta}_n^{\text{ml}}) - 2\nu(\boldsymbol{\theta}) \quad \text{and} \quad \text{BIC} = 2\,\text{lik}_{f,n}(\boldsymbol{\theta}_n^{\text{ml}}) - \log(n)\nu(\boldsymbol{\theta}),$$

where $\nu(\boldsymbol{\theta})$ is the number of free parameters of the mixture distribution (S7) and $\boldsymbol{\theta}_n^{\text{ml}}$ is the corresponding ML estimate. The *Integrated Complete-data Likelihood* (ICL) of Biernacki et al. (2000) is an alternative information-type criteria that focuses on the clustering structure rather than the distribution fit. The definition of the ICL replaces the log-likelihood function with the conditional expected complete log-likelihood within the BIC's construction:

$$\text{ICL} = 2\text{E}\left[\text{clik}_{f,n}(\boldsymbol{\theta}_n^{\text{ml}}) \mid \mathbb{X}_n\right] - \log(n)\nu(\boldsymbol{\theta}). \tag{S5}$$

It turns out that $\text{ICL} = \text{BIC} - \overline{\text{ent}}_n(\boldsymbol{\theta}_n^{\text{ml}})$, where $\overline{\text{ent}}_n(\boldsymbol{\theta}_n^{\text{ml}})$ is the empirical average of the entropy of the MAP assignment computed at the ML estimate $\boldsymbol{\theta}_n^{\text{ml}}$. Thus, the ICL puts an extra penalty on the BIC for the uncertainty of the overall clustering, quantified by

$\overline{\text{ent}}_n(\boldsymbol{\theta}_n^{\text{ml}})$. The ICL recently gained a strong popularity, and it is now included in most software packages. Baudry (2015) investigates the theoretical foundations of the ICL.

The calculations of the AIC, BIC and ICL require:

(*i*) the definition of a clustering probability model so that the likelihood term can be defined and calculated;

(*ii*) that the parameter vector is fitted by ML;

(*iii*) that the effective degrees of freedom of the competing models can be approximated in terms of model's dimension $\nu(\boldsymbol{\theta})$.

The requirements (*i*)-(*ii*) do not allow comparing clustering solutions obtained with methods that, despite being able to retrieve the clusters of interest, are not based on a generating probability model. For example, it is not hard to construct examples where the K-Medoids method based on the Euclidean distance discovers clusters similar to those found with Gaussian mixtures. However, it is not clear how to compute the BIC-type index for a K-Medoid solution. Also, (*iii*) is not always possible in the model-based framework. An example is the subset of RIMLE configurations in $\mathcal{M}$: any $1 \leq \gamma < +\infty$ changes the model complexity without changing the number of estimated parameters. Therefore, in the case of ML for Gaussian mixture models with ERC constraints, the penalty term of both the BIC and the ICL would not reflect the change in the model complexity introduced by different specifications of $\gamma$. The previous issue also affects other types of scale constraints that have been proposed to ensure the existence of the ML for mixtures of more general location-scale families.

---

**Algorithm S2** 10-Fold cross-validation of the likelihood risk (CVLK)

---

**input:** observed sample $\mathbb{X}_n$
**output:** $\widetilde{CVLK}$.

(step 1) randomly partition $\mathbb{X}_n$ into 10 folds $\{\mathbb{X}^{(t)}, t = 1, \ldots, 10\}$, each with (approximately) $n/10$ data points.
**for** $t = 1, \ldots, 10$ **do**
    (step 2.1)  $\widehat{\mathbb{X}} \leftarrow \bigcup_{j \neq t} \mathbb{X}^{(j)}$,
    (step 2.2)  $\boldsymbol{\theta}_{(t)}^{\text{ml}} \leftarrow \arg\max_{\boldsymbol{\theta}} \sum_{\boldsymbol{y} \in \widehat{\mathbb{X}}} \log(\psi_f(\boldsymbol{y}; \boldsymbol{\theta}))$
    (step 2.3)  $\ell_{(t)} \leftarrow \frac{1}{\#\mathbb{X}^{(t)}} \sum_{\boldsymbol{y} \in \mathbb{X}^{(t)}} \log(\psi_f(\boldsymbol{y}; \boldsymbol{\theta}_{(h)}^{\text{ml}}))$
**end for**
(step 3)    Obtain
$$\widetilde{CVLK} \leftarrow \frac{1}{10} \sum_{t=1}^{10} \ell_{(t)}$$

---

**Method-dependent criteria using resampling** In the work of Smyth (2000) the selection problem is solved by using cross-validation and defining the risk as minus the average log-likelihood function of an underlying mixture model; here, it is implemented as the CVLK method. The risk function defined by Smyth (2000) coincides with a quantity that is proportional to the expected Kullback-Leibler information loss over a test set when the true underlying generating distribution is approximated by the mixture model. To

8

some extent, the method mimics the rationale of the AIC criteria and therefore looks for models that best fit the data distribution on unseen data. The precise implementation of the CVLK method is given in Algorithm S2, formulated for the mixture model (S7). In our experiments, CVLK selects a cluster configuration that maximizes $\widetilde{CVLK}$. Algorithm S2 differs from the original proposal for two reasons:

1. Smyth (2000) defined the risk as "minus the average log-likelihood", minimizing this criterion. In step 2.1 we compute "average log-likelihood" values over the test sets to maximize the final criterion, consistently with all its competitors.

2. Smyth (2000) proposed to use random cross-validation with both training and testing data sets containing half of the data points. After having implemented the original proposal, we compared it with alternative cross-validation sampling schemes, discovering that 10-fold cross-validation does a better job. Therefore, in the final experiments, we only considered the 10-fold cross-validation.

The application of the CVLK method requires a probability model to define the likelihood risk. For this reason, we classified it as method-dependent but, in fact, it would be better qualified as "model-dependent" as for the AIC, BIC and ICL. The difference with AIC, BIC and ICL is that, in this case, an approximation of degrees of freedom in terms of the number of free parameters to be estimated is not necessary. For this reason, we apply the CVLK to RIMLE configurations also, for which information criteria do not make much sense.

## S2 Beyond the choice of $K$: an example with the Iris data

The well-known Iris data set, first studied in Anderson (1936) and Fisher (1936), contains 50 samples from each of three speces of Iris flowers, for a total of 150 samples; 4 numerical (continuous) features were measured on each sample. Figure S1 shows the observed features and "true clusters" defined in terms of the Iris species. In Figure S1b, true clusters are adequately captured by normal ellipsoids. The true groups can be captured by different methods, where not all of them are genuinely designed for discovering this type of cluster. In order to show that the number of groups, $K$, is not the only relevant choice to make, we set $K = 3$ (true value) for all of the following methods.

Additional to the K-MEANS and K-MEDOIDS we apply the MBC methods defined in Section S1 as follows.

GM-S      RIMLE solution with $\gamma = 1$ corresponding to the simplest Gaussian mixture model in the rimle package.

GM-C      same as GM-S, but $\gamma = 10^6$. Such a large value of $\gamma$ produces little regularization allowing the covariance matrix parameters of the model to vary almost freely. This extremely large value of $\gamma$ determines a model that is more flexible than GM-S; however, it exposes the algorithm to the risk of finding spurious clusters that are close to having a singular covariance matrix.

TM-S      EMMIX with Student-t mixture models. The clusters' covariance matrices are restricted to be equal and spherical.

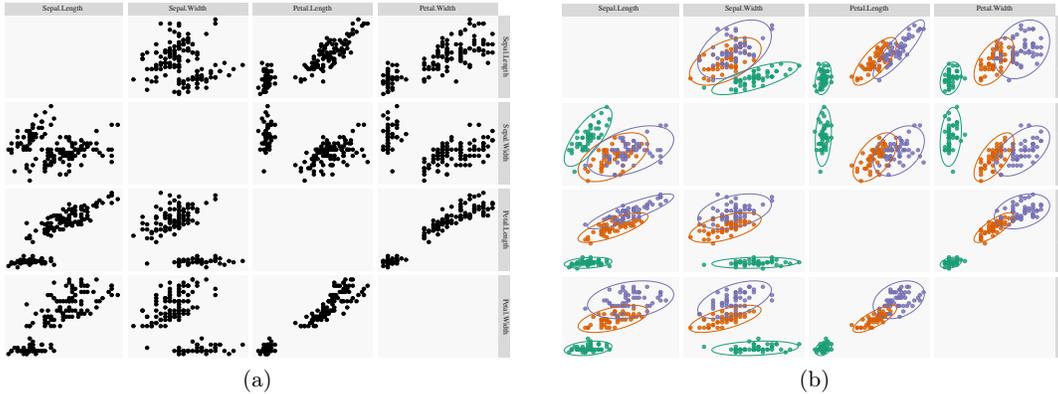

Figure S1: pairwise scatter plots of the Iris data. Panel (a): raw data scatters that does not use groups' information. Panel (b): scatters with colors and 95%-normal ellipses identifying "true clusters" coinciding with the 3 Iris species. Normal ellipses are computed based on the within-group empirical mean and covariance matrix.

TM-C     same as TM-S, except that covariance matrices are not restricted to any model. As in the case of GM-C, this corresponds to a flexible parametrization that causes the risk of finding spurious singular groups.

In this example we also consider two density based clustering methods that are a strong departure from those described in Section S1.1.

CFSFDP     denote a solution obtained with the "clustering by fast search and find of density peaks" (CFSFDP) algorithm of Rodriguez and Laio (2014) implemented into the densityClust software of Pedersen et al. (2017). The CFSFDP is a density-based algorithm. It requires two hyper-parameters used by the algorithm to determine $K$; we followed the guidelines provided in Rodriguez and Laio (2014), setting the hyper-parameters to obtain three groups.

DBSCAN:     this solution is obtained with the dbscan package of Hahsler et al. (2019) implementing the popular dbscan clustering algorithm of Ester et al. (1996). As in for the CFSFDP, dbscan also requires choosing two bandwidth-type hyper-parameters and a kernel function. The number of clusters $K$ is determined by the algorithm which additionally may identify some data-points as noise. As for the CFSFDP, we set DBSCAN to obtain the best solution with $K = 3$.

The comparison involves 6 algorithms with a total of 8 different settings. Table S3 reports the agreement between all pairs of partitions including the true one. K-MEAN, K-MEDOIDS, GM-S, and TM-S produce the same partition shown in Figure S2a. Another partition that is rather close to the previous one is that produced by the CFSFDP method (see Figure S2b). The two partitions achieve the best agreement with the true partition (CFSFDP does marginally better).

Although GM-C and TM-C are methods that are specifically designed to capture elliptically-shaped clusters, their flexible covariance parametrization drives the algorithms towards excessive fitting in some regions of the data, completely missing the overall structure. The

Table S3: Agreement between pairs of partitions produced in the Iris example. The agreement is expressed as the percentage of data points assigned to the same group by both the row and the column method. For each pair, the agreement is computed with respect to the permutation of cluster labels resulting in the best match. The *Truth* column/row represents the partition corresponding to the true Iris species.

|  | Truth | K-MEANS | K-MEDOIDS | GM-S | GM-C | TM-S | TM-C | CFSFDP | DBSCAN |
|---|---|---|---|---|---|---|---|---|---|
| Truth |  | 89.3% | 89.3% | 89.3% | 66.7% | 89.3% | 66.7% | 90.7% | 68.7% |
| K-MEANS | 89.3% |  | 100.0% | 100.0% | 74.7% | 100.0% | 74.7% | 98.7% | 75.3% |
| K-MEDOIDS | 89.3% | 100.0% |  | 100.0% | 74.7% | 100.0% | 74.7% | 98.7% | 75.3% |
| GM-S | 89.3% | 100.0% | 100.0% |  | 74.7% | 100.0% | 74.7% | 98.7% | 75.3% |
| GM-C | 66.7% | 74.7% | 74.7% | 74.7% |  | 74.7% | 86.0% | 76.0% | 86.0% |
| TM-S | 89.3% | 100.0% | 100.0% | 100.0% | 74.7% |  | 74.7% | 98.7% | 75.3% |
| TM-C | 66.7% | 74.7% | 74.7% | 74.7% | 86.0% | 74.7% |  | 76.0% | 88.7% |
| CFSFDP | 90.7% | 98.7% | 98.7% | 98.7% | 76.0% | 98.7% | 76.0% |  | 76.7% |
| DBSCAN | 68.7% | 75.3% | 75.3% | 75.3% | 86.0% | 75.3% | 88.7% | 76.7% |  |

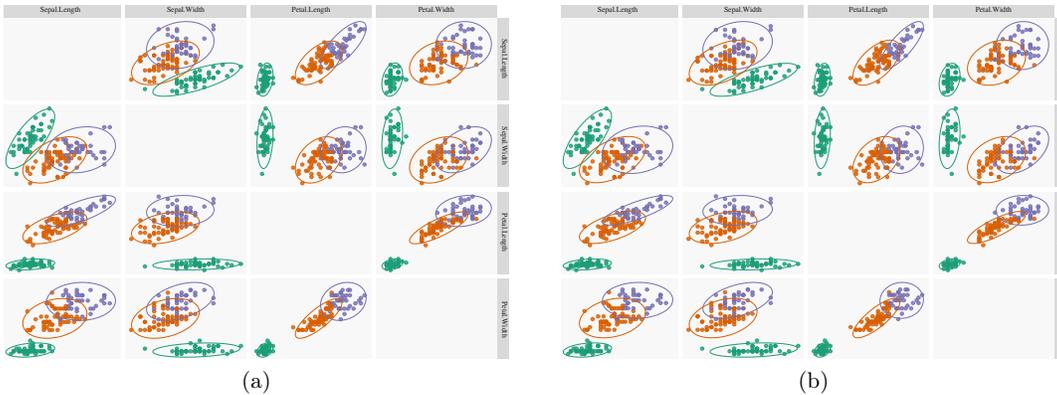

(a)       (b)

Figure S2: Iris data clustering. Panel (a): clustering solutions obtained from K-MEAN, K-MEDOIDS, GM-S, and TM-S methods. Panel (b): clustering solution obtained from the CFSFDP method.

solution corresponding to TM-C can be seen in Figure S3a: one of the mixture components is almost singular, representing a single point so that the software ignores it. GM-C in Figure S3b does a better job because it returns three groups, but one of the mixture components fits a concentrated portion of points in the center of one of the original clusters, splitting it into two clusters, and merges the remaining two original clusters. The DBSCAN solution finds a large proportion of noise, messing all the clusters.

We draw two important conclusions from this example.

- Different algorithms can lead to the same partition. This is not surprising because algorithms built around different principles can target similar cluster notions. For example, K-MEANS and K-MEDOIDS tempt to purse balanced spherical clusters. Since the Student-t model includes the Gaussian model, TM-S can be seen as more general version of GM-S. Both TM-S and GM-S purse spherical groups, although not necessarily balanced groups. In general, what K-MEANS, K-MEDOIDS, TM-S and GM-S have in common is that they are able to find groups that can be separated by linear shapes. Nevertheless, these groups form dense regions that can also be found with the completely different CFSFDP method.

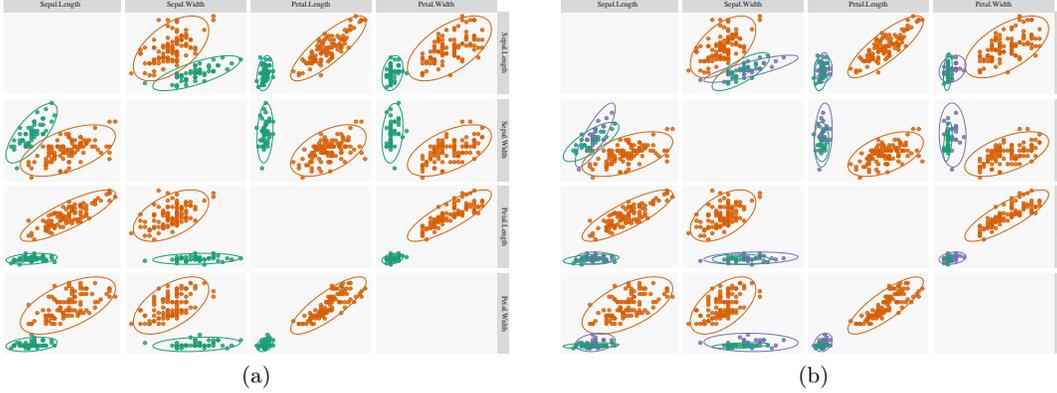

Figure S3: Iris data clustering. Panel (a): clustering solutions obtained from TM-C method. Panel (b): clustering solution obtained from the GM-C method.

- The second conclusion is that methods' flexibility in representing the data structure is a key factor. In principle, GM-C and TM-C are the strongest candidates to capture the true clusters in the example: as shown in Figure S1b, true clusters are well described by elliptical regions, and both GM-C and TM-C are specifically designed to capture these kind of structures. However, the high flexibility of the clusters' covariance matrix models in GM-C and TM-C produced an excessive fitting in some regions of the data. For some of these methods, there is a clear notion of flexibility because this is mapped into the underlying model complexity. However, for methods that are not genuinely formalized in terms of a generating model, it is not straightforward to think about the complexity of the data structures that a certain method can handle. The complexity is usually tuned by the methods' hyper-parameters. This happens for example for the two density-based algorithms considered here, namely CFSFDP and DBSCAN. In clustering problems, $K$ is certainly the major source of the complexity level at which we are willing to describe the data; however, most methods have additional hyper-parameters that can strongly interact with $K$ for obtaining a certain result.

## S3   Further connections with the mixture model likelihood

Under the Gaussian assumption, there is a further connection between the sample scores $H_n(\cdot)$ and $T_n(\cdot)$ and what is called observed *complete data log-likelihood*. In the MBC framework, there are two approaches to fit the unknown mixture parameters of a model like. These approaches correspond to the maximization of one of the following log-likelihood functions:

$$\text{lik}_{f,n}(\boldsymbol{\theta}) = \sum_{i=1}^{n} \log\left(\psi_f(\boldsymbol{x}_i; \boldsymbol{\theta})\right), \qquad \text{clik}_{f,n}(\boldsymbol{\theta}) = \sum_{i=1}^{n}\sum_{k=1}^{K} Z_{ik} \log\left(\pi_k f(\boldsymbol{x}_i; \boldsymbol{\theta}_k)\right), \qquad (S6)$$

with

$$\psi_f(\boldsymbol{x}; \boldsymbol{\theta}) := \sum_{k=1}^{K(\boldsymbol{\theta})} \pi_k f(\boldsymbol{x}; \boldsymbol{\mu}_k, \boldsymbol{\Sigma}_k). \qquad (S7)$$

12

$\text{lik}_{f,n}(\cdot)$ is the log-likelihood function for the mixture model $\psi_f(\cdot)$; $\text{clik}_{f,n}(\cdot)$ is the so-called *complete-data* log-likelihood function where, in practice, the memberships indicators $\{Z_{ik}\}$ are treated as missing data and replaced with their MAP assignment counterpart. Maximization of $\text{clik}_{f,n}(\cdot)$ leads to inconsistent mixture parameters estimates but, as expected, produces more separated groups (see Frühwirth-Schnatter et al., 2019, and references therein). The following statement clarifies the relationships between the previous likelihood quantities and the sample versions of the scores in the case of Gaussian clusters.

**Proposition 1.** *Let $\text{lik}_{\phi,n}(\cdot)$ and $\text{clik}_{\phi,n}(\cdot)$ be respectively the log-likelihood and the complete-data log-likelihood function under the Gaussian group-conditional model. The following hold*

$$\underset{1 \leq m \leq M}{\arg\max} \ H_n(\boldsymbol{\theta}^{(m)}) = \underset{1 \leq m \leq M}{\arg\max} \ \underset{\{Z_{ik}\}}{\max} \ \text{clik}_{\phi,n}(\boldsymbol{\theta}^{(m)}), \tag{S8}$$

$$\underset{1 \leq m \leq M}{\arg\max} \ T_n(\boldsymbol{\theta}^{(m)}) = \underset{1 \leq m \leq M}{\arg\max} \ \text{E}\left[\text{clik}_{\phi,n}(\boldsymbol{\theta}^{(m)}) \mid \mathbb{X}_n\right], \tag{S9}$$

*Proof.* The optimal replacement of $\{Z_{ik}\}$ is given by the MAP assignment

$$\hat{z}_k(\boldsymbol{x}_i; \boldsymbol{\theta}) = \mathbb{I}\left\{k = \underset{1 \leq j \leq K(\boldsymbol{\theta})}{\arg\max} \ \omega_{f,k}(\boldsymbol{x}_i; \boldsymbol{\theta})\right\},$$

that is

$$\underset{\{Z_{ik}\}}{\max} \ \text{clik}_{\phi,n}(\boldsymbol{\theta}^{(m)}) = \sum_{i=1}^{n} \sum_{k=1}^{K(\boldsymbol{\theta}^{(m)})} \hat{z}_k(\boldsymbol{x}_i; \boldsymbol{\theta}^{(m)}) \log\left(\pi_k^{(m)} \phi(\boldsymbol{x}_i; \boldsymbol{\mu}_k^{(m)}, \boldsymbol{\Sigma}_k^{(m)})\right);$$

the latter is at the basis of the EM algorithm for the maximization of the classification likelihood (McLachlan and Krishnan, 2007). Under the Gaussian group-conditional model, $\hat{z}_k(\boldsymbol{x}_i; \boldsymbol{\theta}^{(m)}) = \mathbb{I}\{\boldsymbol{x}_i \in Q_k(\boldsymbol{\theta})\}$. Using the fact that $\text{qs}(\boldsymbol{x}_i; \boldsymbol{\theta}_k^{(m)}) = c + \log(\pi_k \phi(\boldsymbol{x}_i; \boldsymbol{\mu}_k, \boldsymbol{\Sigma}_k))$ with $c = p \log(\sqrt{2\pi})/2$, we obtain

$$\underset{\{Z_{ik}\}}{\max} \ \text{clik}_{\phi,n}(\boldsymbol{\theta}^{(m)}) = \sum_{i=1}^{n} \sum_{k=1}^{K(\boldsymbol{\theta}^{(m)})} \hat{z}_k(\boldsymbol{x}_i; \boldsymbol{\theta}^{(m)}) \log\left(\pi_k^{(m)} \phi(\boldsymbol{x}_i; \boldsymbol{\mu}_k^{(m)}, \boldsymbol{\Sigma}_k^{(m)})\right),$$

$$= \sum_{i=1}^{n} \sum_{k=1}^{K(\boldsymbol{\theta}^{(m)})} \mathbb{I}\{\boldsymbol{x}_i \in Q_k(\boldsymbol{\theta})\} \left(\text{qs}(\boldsymbol{x}_i; \boldsymbol{\theta}_k^{(m)}) - c\right),$$

$$= H_n(\boldsymbol{\theta}^{(m)}) - nc.$$

The previous expression proves (S8). Now observe that under the Gaussian assumption

13

the posterior weights, $\omega(\cdot)$, would coincide with the $\tau(\cdot)$ weights, therefore

$$\begin{aligned}
\mathrm{E}\left[\mathrm{clik}_{\phi,n}(\boldsymbol{\theta}^{(m)}) \mid \mathbb{X}_n\right] &= \sum_{i=1}^{n} \sum_{k=1}^{K(\boldsymbol{\theta}^{(m)})} \mathrm{E}[Z_{ik}|\mathbb{X}_n] \log\left(\pi_k^{(m)} \phi(\boldsymbol{x}_i; \boldsymbol{\mu}_k^{(m)}, \boldsymbol{\Sigma}_k^{(m)})\right), \\
&= \sum_{i=1}^{n} \sum_{k=1}^{K(\boldsymbol{\theta}^{(m)})} \tau_k(\boldsymbol{x}_i; \boldsymbol{\theta}^{(m)}) \log\left(\pi_k^{(m)} \phi(\boldsymbol{x}_i; \boldsymbol{\mu}_k^{(m)}, \boldsymbol{\Sigma}_k^{(m)})\right), \\
&= \sum_{i=1}^{n} \sum_{k=1}^{K(\boldsymbol{\theta}^{(m)})} \tau_k(\boldsymbol{x}_i; \boldsymbol{\theta}^{(m)}) \left(\mathrm{qs}(\boldsymbol{x}_i; \boldsymbol{\theta}_k^{(m)}) - c\right) \\
&= T_n(\boldsymbol{\theta}^{(m)}) - nc,
\end{aligned}$$

which proves (S9). □

## S4 Data description

In this section, we provide more information on the data sets used for the experimental analysis, both for real data and simulation designs (figures are provided at the end of the section).

### S4.1 Real data sets

We use a total of four different real data sets. They are described next, together with an indication of the total number of points ($n$), number of dimensions ($d$), and number of classes $K$.

**Iris** ($n = 150$, $d = 4$, $K = 3$)
    The famous Iris data set was introduced by Anderson (1936) and Fisher (1936). It collects equally split measurements on three Iris flower species: *Versicolor, Virginica* and *Setosa* (see Figure S4, left panel). Two of the classes, Versicolor and Virginica, show a substantial overlap, making it generally difficult to tell them apart for clustering algorithms.

**Banknote** ($n = 200$, $d = 6$, $K = 2$)
    The Banknote data set was introduced in Flury and Riedwyl (1988) and collects observations on Swiss 1000-franc banknotes, evenly split between two classes: *counterfeit* and *genuine* notes. It collects measurements on the bill: width of the note measured along the left and right sides; bottom and top margins' width; diagonal image length; bill's length. Figure S4, right panel shows data pairs plot. The two classes show a moderate overlap; the higher variability of the characteristics of the counterfeit notes may induce algorithms to split the class into two different groups.

**Olive** ($n = 572$, $d = 8$, $K = 3$ or $K = 9$)
    The Olive data set, first appeared in Forina and Tiscornia (1982) and Forina et al. (1983), collects measurements on olive oils' fatty acids. The data features two levels of classification, based on oils' origins: a coarser one consisting of 3 macro-regions and

14

a finer one identifying 9 regions. Figure S5 shows pairs plot for both classification levels. While some areas are better separated (e.g. West Liguria, Costal Sardinia and Inland Sardinia), others show larger similarities (e.g. Sicily, Calabria and South Apulia). As can be noted from the graphs, data comes with a high amount of discreteness. Moreover, for some areas, the scatters are highly concentrated along certain directions. These aspects make it particularly difficult to retrieve the true partitions.

**Wine ($n = 178$, $d = 13$, $K = 3$)**

The Wine data set (Forina et al., 1988) contains results of chemical analysis of wines grown in the same region in Italy, but from three different cultivars. Thus, the number of classes is identified with $K = 3$. The three classes are (almost) equally represented in the data. In a classification context, the classes are separable, and the data set does not appear too challenging for many classification methods. Things are different for cluster analysis: the high-dimensional setting and the classes structure appear to pose a serious challenge to some of the validation criteria, with model-based validation indexes expected to perform better. Figure S6 shows the pairs plot for the Wine data. The classes are well separated along some dimensions, while others appear less informative.

## S4.2 Simulated data sets

In this section, we present a detailed description of the data generating process for the simulated data sets. The reader can refer to Figure S7 for a visual description of the simulated data, both with and without groups membership.

**Pentagon5** The DGP is a mixture of 5 spherical Gaussian components in a two-dimensional space. The Gaussian components are centered along the sides of a pentagon centered at the origin. The underlying density function is

$$f(\bm{x}) = \sum_{k=1}^{5} \pi_k \phi(\bm{x}; \bm{\mu}_k; \bm{\Sigma}_k)$$

where $\phi(\cdot; \bm{\mu}, \bm{\Sigma})$ is the Gaussian density with parameters $\bm{\mu}$ and $\bm{\Sigma}$. Components parameters are set at the following values:

- $\pi_1 = 0.2$; $\pi_2 = 0.35$; $\pi_3 = 0.35$; $\pi_4 = 0.05$; $\pi_5 = 0.05$;
- $\bm{\mu}_1 = (0, 5)^\mathsf{T}$; $\bm{\mu}_2 = (-4.5, -0.5)^\mathsf{T}$; $\bm{\mu}_3 = (4.5, -0.5)^\mathsf{T}$; $\bm{\mu}_4 = (3, -2.5)^\mathsf{T}$; $\bm{\mu}_5 = (-3, -2.5)^\mathsf{T}$;
- $\bm{\Sigma}_k = \begin{bmatrix} 1 & 0 \\ 0 & 1 \end{bmatrix}$, $k = 1 \ldots 5$.

Assuming the definition of ground truth that each mixture component generates a cluster, here we have $K = 5$. However, due to the highly unbalanced components' proportions, a 3 cluster solution may also be reasonable (see Figure S7). If within-group homogeneity is pursued, with small sample sizes (such as the one we consider), it may be preferable to have 3 groups rather than 5.

15

**T52D** This is a two-dimensional design, where the DGP is a mixture of Student-t distributions, with density

$$f(x) = \sum_{k=1}^{5} \pi_k t\left(\boldsymbol{x}; \nu_k, \boldsymbol{\mu}_k, \boldsymbol{\Sigma}_k\right);$$

$t(\cdot)$ is the multivariate Student-t density. For the $k$-th component, $\nu_k$ are degrees of freedom, $\boldsymbol{\mu}_k$ is the mean vector and $\boldsymbol{\Sigma}_k$ is covariance matrix. Note that for a Student-t distribution with $\nu_k > 2$, $\boldsymbol{\Sigma}_k = (\nu_k - 2)/\nu_k \boldsymbol{V_k}$, where $\boldsymbol{V_k}$ is a symmetric positive-definite scatter matrix. For convenience, we parameterize the Student-t distribution in terms of the covariance matrices since we always consider $\nu_k > 2$. Components' parameters are set at the following values:

- $\pi_1 = 0.15$; $\pi_2 = 0.4$; $\pi_3 = 0.05$; $\pi_4 = 0.15$; $\pi_5 = 0.25$;
- $\nu_1 = 10$, $\nu_2 = 12$, $\nu_3 = 14$, $\nu_4 = 16$, $\nu_5 = 18$.
- $\boldsymbol{\mu}_1 = (0,3)^\mathsf{T}$; $\boldsymbol{\mu}_2 = (7,1)^\mathsf{T}$; $\boldsymbol{\mu}_3 = (5,9)^\mathsf{T}$; $\boldsymbol{\mu}_4 = (-11,11)^\mathsf{T}$; $\boldsymbol{\mu}_5 = (-7,5)^\mathsf{T}$;
- $\boldsymbol{\Sigma}_1 = \begin{bmatrix} 1 & 0.5 \\ 0.5 & 1 \end{bmatrix}$, $\boldsymbol{\Sigma}_2 = \begin{bmatrix} 2 & -1.5 \\ -1.5 & 2 \end{bmatrix}$, $\boldsymbol{\Sigma}_3 = \begin{bmatrix} 2 & 1.3 \\ 1.3 & 2 \end{bmatrix}$, $\boldsymbol{\Sigma}_4 = \begin{bmatrix} 0.5 & 0 \\ 0 & 0.5 \end{bmatrix}$, $\boldsymbol{\Sigma}_5 = \begin{bmatrix} 2.5 & 0 \\ 0 & 2.5 \end{bmatrix}$.

Given the strong separation, it makes sense to assume that each mixture component generates a cluster. Therefore, $K = 5$ in this case. However, with few data, the difficulty arises because of the clusters' varying shape, which causes problems for solutions with fixed-shape clusters. Moreover, with a small $n = 300$, this data set may cause trouble to the selection methods based on resampling. Indeed, the third component generates on average 15 points and may easily disappear in bootstrap samples or cross-validation folds.

**T510D** The DGP is the same as T52D, except that we add 8 uninformative spherically distributed marginals. Let $\dot{\boldsymbol{\mu}}_k$ and $\dot{\boldsymbol{\Sigma}}_k$ the mean and covariance matrix parameters for the $k$-th component of the T510D sample design. Define

- $\dot{\boldsymbol{\mu}}_k = (\boldsymbol{\mu}_k^\mathsf{T}, 0, 0, 0, 0, 0, 0, 0, 0)^\mathsf{T}$;
- $\dot{\boldsymbol{\Sigma}}_k = \begin{bmatrix} \boldsymbol{\Sigma}_k & \mathbf{0} \\ \mathbf{0} & \boldsymbol{I}_8 \end{bmatrix}$;

where $\boldsymbol{\mu}_k$ and $\boldsymbol{\Sigma}_k$ are the mean and covariance matrix parameters in the previous T52D model. The addition of the 8 uninformative features do not provide clustering information, which is only defined along the first two dimensions. This DGP adds noise to the previous DGP because it introduces some interaction between informative marginals and uninformative ones. Along some pair-wise scatter plot (see Figure S7), it is possible to visualize 3 or 4 clusters rather than the true defined $K = 5$. With $n = 300$, the difficulty is also due to the small ratio $n_k/p$, where $n_k = \pi_k n$ is the expected number of points belonging to the $k$-th group. For example, for the third group, we expect 15 data points into $p = 10$ dimensions, which makes particularly difficult to jointly estimate the mean and the covariance parameter.

**Flower2** The DGP is a mixture of 5 components with equal proportions. The density function is

$$f(\boldsymbol{x}) = \sum_{i=1}^{2} \pi_i u_i(\boldsymbol{x}) + \sum_{l=3}^{4} \pi_l t_l(\boldsymbol{x}; \boldsymbol{\mu}_t, \boldsymbol{\Sigma}_t) + \pi_5 \phi(\boldsymbol{x}; \boldsymbol{\mu}_\phi, \boldsymbol{\Sigma}_\phi);$$

where $u(\cdot)$, $t(\cdot)$ and $\phi(\cdot)$ are uniform, Gaussian and Student-t densities respectively. The distributions used are obtained as follows:

- $\pi_i = 0.2$ for $i = 1, \ldots, 5$;

- $u_1$ is a uniform distribution on the two-dimensional rectangle with vertices in $(-1, 1)$, $(-1, 10)$, $(1, 1)$, $(1, 10)$, rotated 45 degrees clockwise;

- $u_2$ is defined as $u_1$, except that the rotation occurs counter-clockwise;

- $t_3$ is a Student-t distribution with 9 degrees of freedom, centered on $\boldsymbol{\mu}_t = (0, 5)^\mathsf{T}$, with covariance

$$\boldsymbol{\Sigma}_t = \begin{bmatrix} 1 & 0 \\ 0 & 10 \end{bmatrix}$$

  and the overall scatter is rotated 135 degrees clockwise.

- $t_4$ is defined as $t_3$, except that the rotation occurs counter-clockwise;

- $\phi_5$ is a standard Gaussian distribution in dimension 2, that is $\boldsymbol{\mu}_\phi = (0, 0)^\mathsf{T}$ with covariance matrix $\boldsymbol{\Sigma}_\phi = \boldsymbol{I}_2$.

It may be challenging to identify 5 groups here because the DGP mixes elliptical shapes (Gaussian and Student-t) with non-elliptical ones (uniform distributions), and features strong overlap between the components.

**Uniform** Provides a case of unclustered data in two dimensions. Points are sampled from a uniform distribution over the unit square with vertices on $(0, 0)$, $(0, 1)$, $(1, 0)$ and $(1, 1)$. Not all validation criteria are able to deal with the unclustered case: some criteria can not by construction (e.g. FW); others (e.g. AIC and BIC) tend to select overly-complex solutions trying to better accommodate the data.

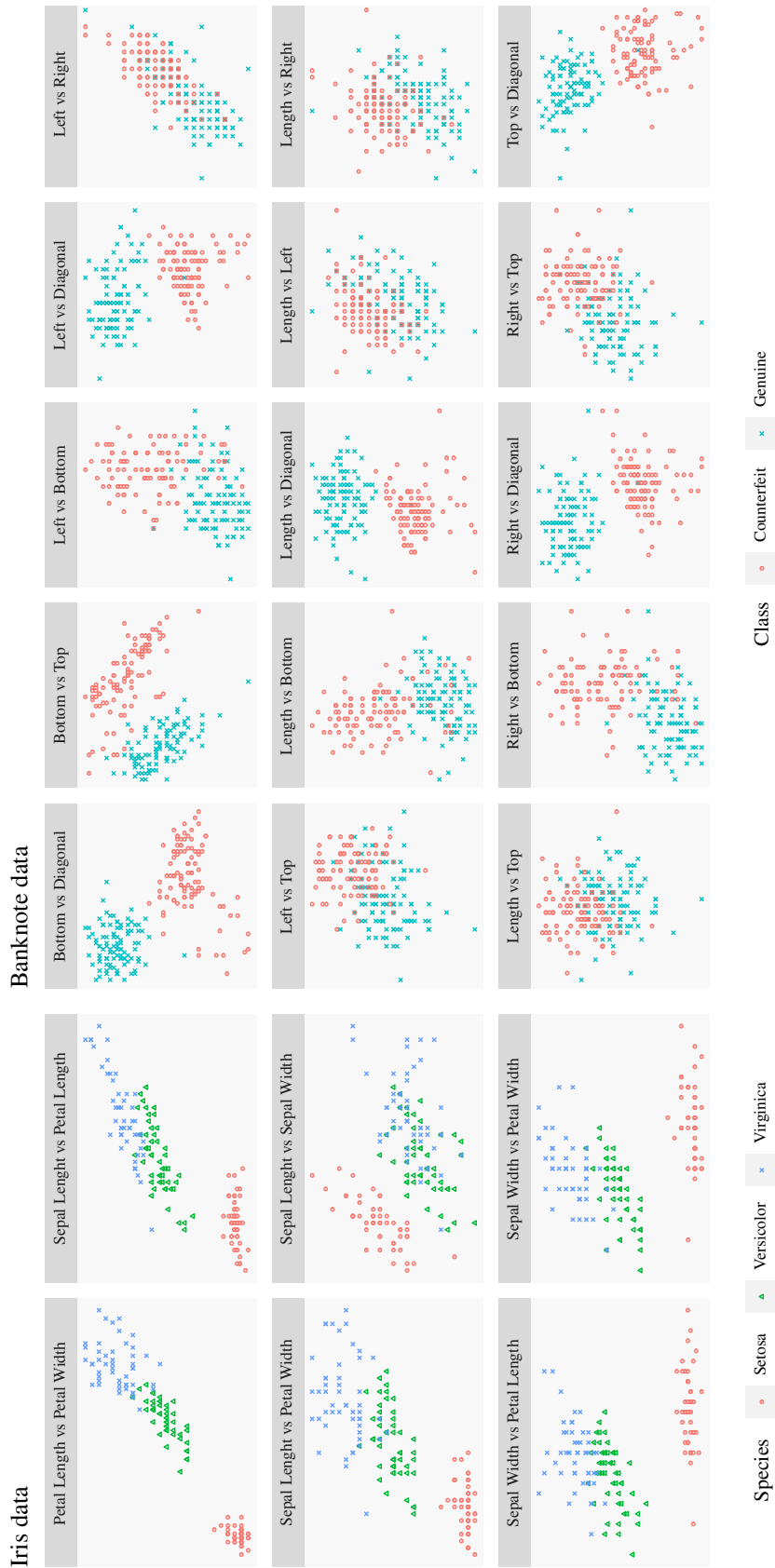

Figure S4: Pairs plots for the Iris data (left) and Banknote data (right).

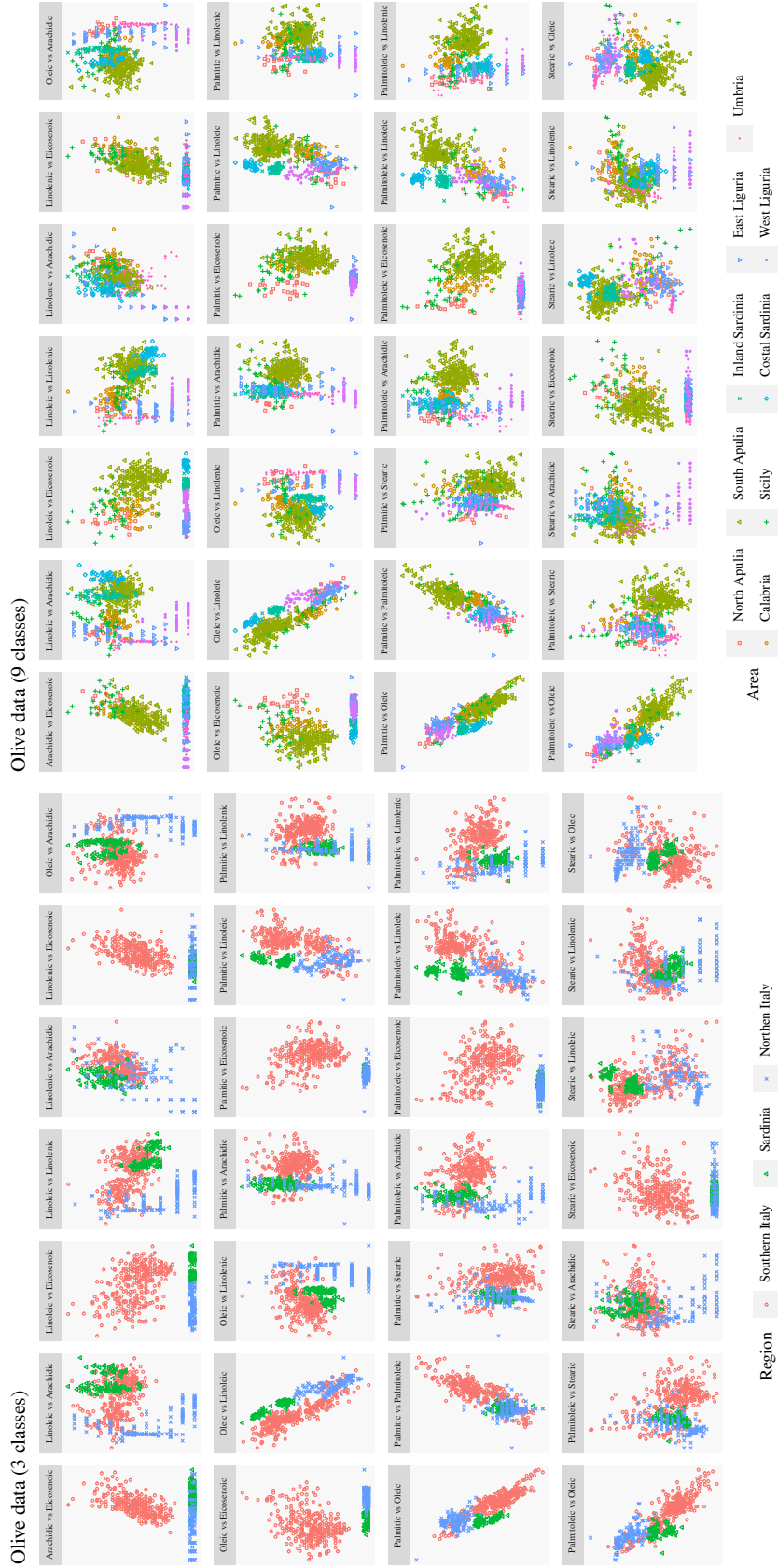

Figure S5: Pairs plot for Olive data, with coarser (left) and finer (right) classifications.

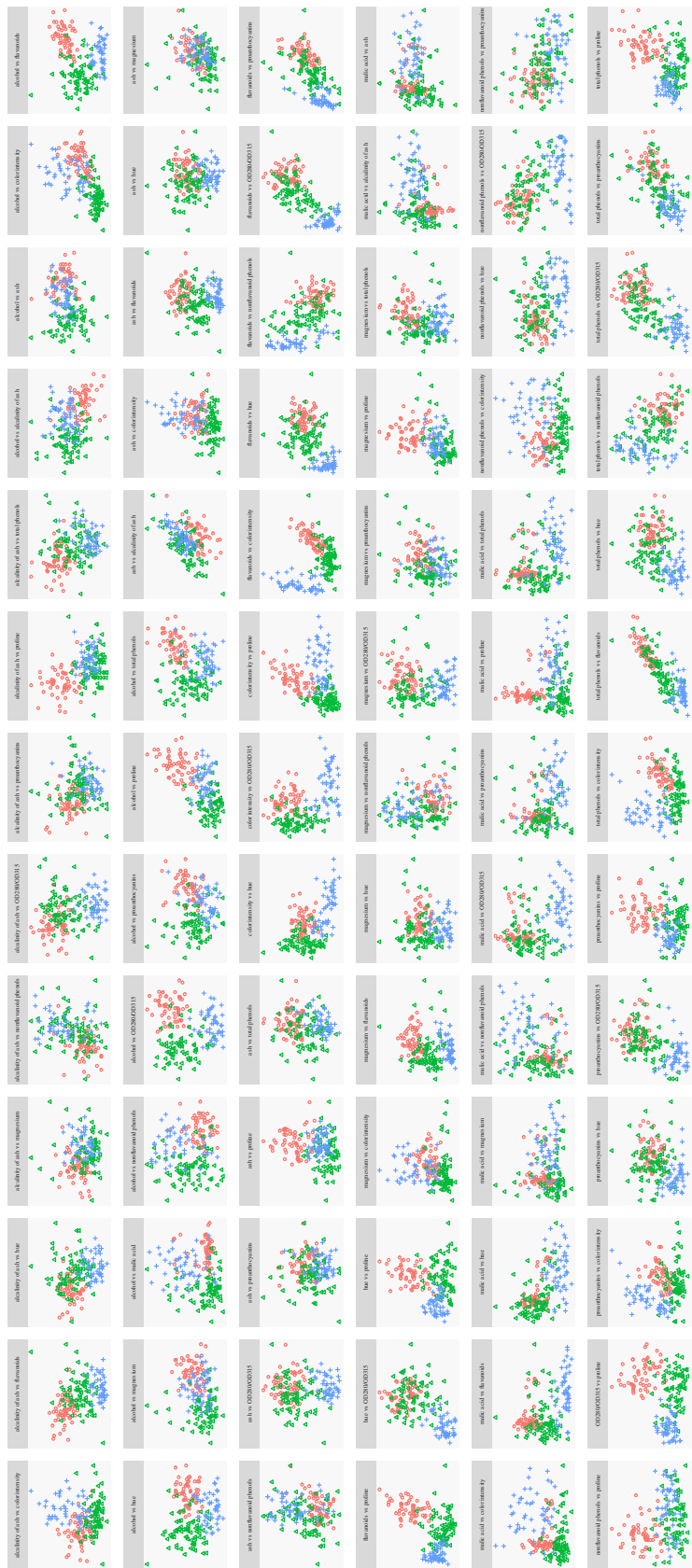

Figure S6: Pairs plots for Wine data set.

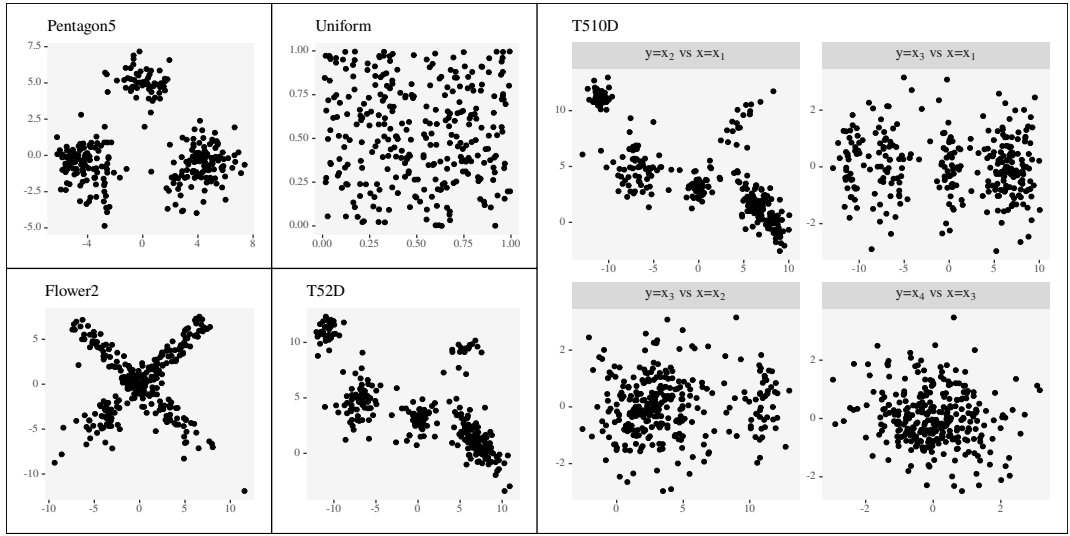

(a) Example scatter plots ($n = 300$ sample points each) from the 5 simulated data sets' DGPs, without group membership.

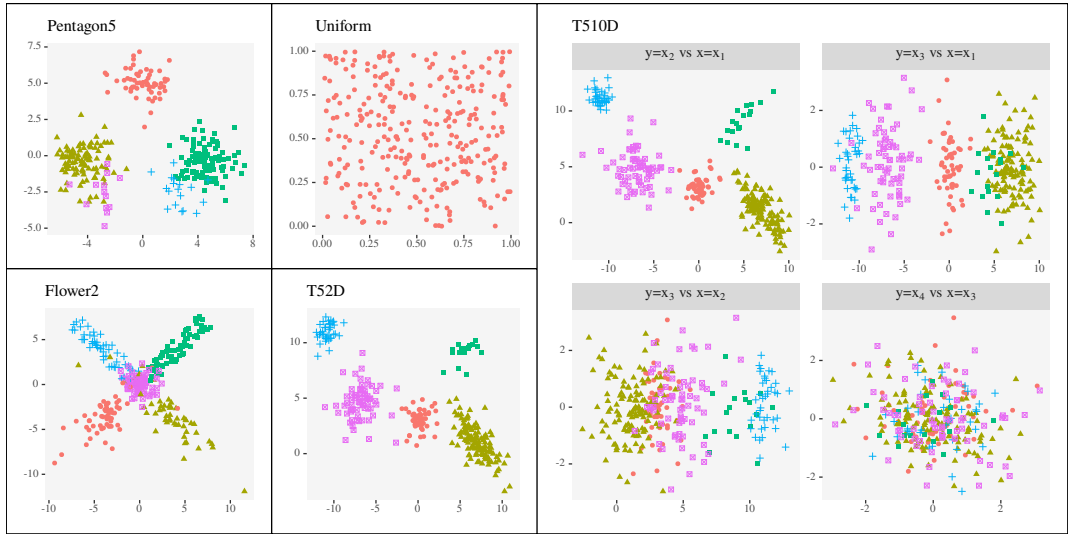

(b) Example scatter plots ($n = 300$ sample points each) from the 5 simulated data sets' DGPs. Group membership is highlighted by colors and points' shape.

Figure S7: Simulated data sets. For T510D (rightmost panel) we plot the first two marginals ($x_1$ and $x_2$), a combination of the them with an uninformative marginal ($x_3$) and two uninformative marginals ($x_3$ and $x_4$).

# S5 Additional details on the experimental analysis and discussion of the results

## S5.1 Experimental setup

**Clustering methods and algorithms** For each data set, we obtain several clustering solutions, indexed with $m \in \mathcal{M}$. In practice, each member $m \in \mathcal{M}$ is the output of an algorithm that implements a specific method, configured with a certain number of clusters $K(m)$ plus other hyper-parameters. $\mathcal{M}$ denotes the set of all cluster configurations obtained from a fixed input data set, and we want to select one of them. We consider a set of 440 cluster configurations, i.e. $|\mathcal{M}| = 440$, obtained with the algorithms listed in Section S1.1 and the following hyper-parameters (total number of methods from each algorithm in parentheses):

- (10) K-MEANS with $K \in \{1, \ldots, 10\}$;
- (10) K-MEDOIDS with $K \in \{1, \ldots, 10\}$;
- (140) MCLUST with $K \in \{1, \ldots, 10\}$ and 14 covariance parametrization (see Table S1);
- (180) RIMLE with $K \in \{1, \ldots, 10\}$, eigen-ratio constraint $\gamma \in \{1, 5, 10, 100, 1000, 10000\}$, and three initialization methods;[2]
- (100) EMMIX with with $K \in \{1, \ldots, 10\}$, the 5 parametrization described in Section S1.1, and both skewed and symmetric Student-t distributions.

**Selection methods** For each data set, we select the desired solution from $\mathcal{M}$ using the 13 different selection methods in Section S1.2. We remind that not all the validation criteria can be computed for all members of $\mathcal{M}$ (Section S1.2).

In-sample criteria (Table S2) are straightforward to compute: all the 440 methods are fitted once to the full data and the obtained solution is used to evaluate the criterion (whenever possible). On the other hand, resample criteria require multiple solutions to be computed. This is accomplished as follows: each of the 440 method is fitted multiple times on (sub-samples of) the data as prescribed by the validation criterion; solutions so obtained are used to compute the validation criterion. In our experiments

- Cross-validation criteria use the 10-fold cross-validation scheme for all data sets (implying that each of the 440 method is fitted 10 times);
- Bootstrap criteria use $B = 1000$ replicates for all the real data sets; $B = 100$ replicates for all the simulated data sets.

For real data sets, as a robustness check, we also show results obtained considering only the first $B = 100$ replicates for the bootstrap criteria.

---

[2]These are `InitClust`, `pam` and `k-means`; see `otrimle` package documentation for detailed description.

For resampling criteria, methods producing at least 25% of invalid solutions (e.g. singular mixture component; algorithm not converging) across resamples are discarded. That is, they are not considered for selection when using the criterion for which they were discarded.

For all criteria, it might happen that different methods achieve the same highest validation score; to split the ties, we randomly select a unique solution among them. Thus, each validation criterion selects a unique solution.

**Performance measures** For each data set and method, we measure: (*i*) agreement with respect to true clusters' memberships; (*ii*) selection of the number of clusters compared to the ground truth. Point (*i*) is measured using the Adjusted Rand Index (ARI) of Hubert and Arabie (1985) and the Variation of Information Criterion (VIC) of Meilă (2007) These performance measures are computed for all the 440 methods fitted on the full sample. Both in-sample and resample criteria select only one of the 440 candidate solutions. Note that in-sample validation criteria select among solutions fitted on the full data and the performance measures for the selected solutions are reported. Resample criteria use multiple re-fits of the same clustering method to select a solution. Here, we report the performance measure as calculated on the selected clustering method fitted on the full data. Both for the 13 selection criteria and for BEST ARI and BEST VIC, it might happen that multiple methods $m \in \mathcal{M}$ achieve the same, optimal partition. In these cases, ties are broken at random and the selected solution is reported.

## S5.2 Results for real data sets

Table S4 shows detailed results for the selection criteria on real data sets, obtained with both $B = 100$ and $B = 1000$; this affects FW, BQH and BQS criteria only, and is presented as a robustness check, allowing to assess how a smaller $B$ affects the quality of the Monte Carlo approximation of the Bootstrap distribution. This is important because, when the computational cost of refitting the clusters $B$ times is large, one may consider a low value of $B$. Results are qualitatively the same but in three occasions: *(i)* FW presents much better results with $B = 1000$ for the Wine data (Table S4i, Table S4j); *(ii)* for BQH and BQS on the Banknote data, results improve by a good margin with $B = 1000$ (Table S4c, Table S4d); *(iii)* for BQH on the Olive data (Table S4g, Table S4h), using an higher $B = 1000$ gives better results. Hence, BQS seems slightly more robust to a lower choice of $B$ with respect to BQH.

Figure S8 show a complete visualization of the proposed methodology for real data sets. The figure shows results for both the hard score and smooth score versions, contrasting the in-sample criteria, QH and QS, with their resample counterparts, BQH and BQS. Each panel refers to one real data set, and has two plots: the top-plot shows results obtained with $B = 100$ bootstrap resamples; the bottom-plot shows results using $B = 1000$ resamples. On the horizontal axis of each plot we list all the 440 members of $\mathcal{M}$, hierarchically sorted on: the clustering algorithm; increasing $K$ (reported in the plot's labels); the increasing complexity of the underlying clustering model. The method behind each member of $\mathcal{M}$ is denoted with colors. For some of the algorithms, some settings produced missing solutions for the in-sample estimates and/or an excessive amount of missing solutions across bootstrap repetitions; these cases are discarded ($\times$ symbol).

Table S4: Clustering solutions selected by the 13 selection criteria under comparison on real data sets. First two rows in each table indicate the two best feasible solutions: the BEST ARI and the BEST VIC partitions. The column "K" reports the number of groups discovered by the selected solution. The "Algorithm" and "Configuration" columns identify the specific member of $\mathcal{M}$ that is selected. ARI and the VIC columns measure the agreement of the selected partition with respect to the ground truth. Best ARI and VIC are highlighted in bold.

(a) Iris data set ($B = 100$)

| Criterion | K | Algorithm | Configuration | ARI | VIC |
|---|---|---|---|---|---|
| BEST ARI | 3 | Mclust | EEE | 0.94 | −0.26 |
| BEST VIC | 3 | Mclust | EEE | 0.94 | −0.26 |
| AIC | 6 | Emmix | mvt; ncov=3 | 0.57 | −1.52 |
| BIC | 2 | Mclust | VEV | 0.57 | −0.67 |
| ICL | 2 | Mclust | VEV | 0.57 | −0.67 |
| QH | 7 | Emmix | mvt; ncov=4 | 0.42 | −1.56 |
| QS | 7 | Emmix | mvt; ncov=4 | 0.42 | −1.56 |
| CH | 3 | K-Median | | 0.73 | −0.76 |
| ASW | 2 | Rimle | $\gamma = 10000$; InitClust | 0.57 | −0.67 |
| FW | 2 | Mclust | EEE | 0.57 | −0.67 |
| CVLK | 4 | Rimle | $\gamma = 10000$; InitClust | 0.81 | −0.57 |
| CVQH | 4 | Rimle | $\gamma = 10000$; InitClust | 0.81 | −0.57 |
| CVQS | 4 | Rimle | $\gamma = 1000$; InitClust | 0.81 | −0.58 |
| BQH | 3 | Rimle | $\gamma = 100$; pam | **0.9** | **−0.32** |
| BQS | 3 | Rimle | $\gamma = 100$; pam | **0.9** | **−0.32** |

(b) Iris data set ($B = 1000$)

| Criterion | K | Algorithm | Configuration | ARI | VIC |
|---|---|---|---|---|---|
| BEST ARI | 3 | Mclust | EEE | 0.94 | −0.26 |
| BEST VIC | 3 | Mclust | EEE | 0.94 | −0.26 |
| AIC | 6 | Emmix | mvt; ncov=3 | 0.57 | −1.52 |
| BIC | 2 | Mclust | VEV | 0.57 | −0.67 |
| ICL | 2 | Mclust | VEV | 0.57 | −0.67 |
| QH | 7 | Emmix | mvt; ncov=4 | 0.42 | −1.56 |
| QS | 7 | Emmix | mvt; ncov=4 | 0.42 | −1.56 |
| CH | 3 | K-Median | | 0.73 | −0.76 |
| ASW | 2 | Rimle | $\gamma = 10000$; InitClust | 0.57 | −0.67 |
| FW | 2 | Mclust | EVE | 0.57 | −0.67 |
| CVLK | 4 | Rimle | $\gamma = 10000$; InitClust | 0.81 | −0.57 |
| CVQH | 4 | Rimle | $\gamma = 10000$; InitClust | 0.81 | −0.57 |
| CVQS | 4 | Rimle | $\gamma = 1000$; InitClust | 0.81 | −0.58 |
| BQH | 3 | Rimle | $\gamma = 100$; pam | **0.9** | **−0.32** |
| BQS | 3 | Rimle | $\gamma = 100$; pam | **0.9** | **−0.32** |

(c) Banknote data set ($B = 100$)

| Criterion | K | Algorithm | Configuration | ARI | VIC |
|---|---|---|---|---|---|
| BEST ARI | 2 | K-Means | | **1** | **0** |
| BEST VIC | 2 | K-Means | | **1** | **0** |
| AIC | 6 | Emmix | mvt; ncov=3 | 0.6 | −1.16 |
| BIC | 3 | Mclust | VVE | 0.84 | −0.43 |
| ICL | 3 | Mclust | VVE | 0.84 | −0.43 |
| QH | 10 | Rimle | $\gamma = 10000$; pam | 0.26 | −2.14 |
| QS | 10 | Rimle | $\gamma = 10000$; pam | 0.26 | −2.14 |
| CH | 2 | Rimle | $\gamma = 1$; pam | **1** | **0** |
| ASW | 2 | Mclust | EII | **1** | **0** |
| FW | 2 | Rimle | $\gamma = 10000$; kmeans | 0.98 | −0.08 |
| CVLK | 3 | Rimle | $\gamma = 10$; InitClust | 0.85 | −0.42 |
| CVQH | 3 | Mclust | VEV | 0.78 | −0.62 |
| CVQS | 3 | Mclust | VEV | 0.78 | −0.62 |
| BQH | 4 | Rimle | $\gamma = 10$; InitClust | 0.67 | −0.84 |
| BQS | 4 | Rimle | $\gamma = 10$; InitClust | 0.67 | −0.84 |

(d) Banknote data set ($B = 1000$)

| Criterion | K | Algorithm | Configuration | ARI | VIC |
|---|---|---|---|---|---|
| BEST ARI | 2 | K-Means | | **1** | **0** |
| BEST VIC | 2 | K-Means | | **1** | **0** |
| AIC | 6 | Emmix | mvt; ncov=3 | 0.6 | −1.16 |
| BIC | 3 | Mclust | VVE | 0.84 | −0.43 |
| ICL | 3 | Mclust | VVE | 0.84 | −0.43 |
| QH | 10 | Rimle | $\gamma = 10000$; pam | 0.26 | −2.14 |
| QS | 10 | Rimle | $\gamma = 10000$; pam | 0.26 | −2.14 |
| CH | 2 | Rimle | $\gamma = 1$; pam | **1** | **0** |
| ASW | 2 | Mclust | EII | **1** | **0** |
| FW | 2 | Rimle | $\gamma = 1000$; kmeans | 0.98 | −0.08 |
| CVLK | 3 | Rimle | $\gamma = 10$; InitClust | 0.85 | −0.42 |
| CVQH | 3 | Mclust | VEV | 0.78 | −0.62 |
| CVQS | 3 | Mclust | VEV | 0.78 | −0.62 |
| BQH | 3 | Rimle | $\gamma = 10$; kmeans | 0.86 | −0.37 |
| BQS | 3 | Rimle | $\gamma = 10$; kmeans | 0.86 | −0.37 |

The difference between BQH and BQS is not clear-cut. The overfitting behavior described by dashed-lines leaving the bootstrap confidence bands is present in all plots, although it is less evident in Figure S7c and Figure S7d, whose plots are dominated by the wider scale in score values. There is no appreciable difference comparing the 100 and 1000 bootstrap resamples, which indicates that the methodology is robust to a lower choice of $B$.

Table S4

(e) Olive data set (3 classes, $B = 100$)

| Criterion | K | Algorithm | Configuration | ARI | VIC |
|---|---|---|---|---|---|
| BEST ARI | 3 | Emmix | mvt; ncov=3 | 1 | −0.03 |
| BEST VIC | 3 | Emmix | mvt; ncov=3 | 1 | −0.03 |
| AIC | 10 | Emmix | mvt; ncov=3 | 0.33 | −1.74 |
| BIC | 6 | Emmix | mvt; ncov=3 | 0.52 | −1.42 |
| ICL | 6 | Emmix | mvt; ncov=3 | 0.52 | −1.42 |
| QH | 10 | Mclust | VVV | 0.29 | −1.84 |
| QS | 10 | Mclust | VVV | 0.29 | −1.84 |
| CH | 3 | K-Means | | 0.32 | −1.88 |
| ASW | 2 | Emmix | mvt; ncov=5 | 0.39 | −1.28 |
| FW | 2 | Emmix | mvt; ncov=3 | **0.82** | **−0.42** |
| CVLK | 10 | Rimle | $\gamma = 1000$; InitClust | 0.3 | −1.81 |
| CVQH | 7 | Emmix | mvt; ncov=3 | 0.28 | −2.04 |
| CVQS | 7 | Emmix | mvt; ncov=3 | 0.28 | −2.04 |
| BQH | 8 | Emmix | mvt; ncov=4 | 0.43 | −1.32 |
| BQS | 8 | Mclust | VVV | 0.49 | −1.28 |

(f) Olive dataset (3 classes, $B = 1000$)

| Criterion | K | Algorithm | Configuration | ARI | VIC |
|---|---|---|---|---|---|
| BEST ARI | 3 | Emmix | mvt; ncov=3 | 1 | −0.03 |
| BEST VIC | 3 | Emmix | mvt; ncov=3 | 1 | −0.03 |
| AIC | 10 | Emmix | mvt; ncov=3 | 0.33 | −1.74 |
| BIC | 6 | Emmix | mvt; ncov=3 | 0.52 | −1.42 |
| ICL | 6 | Emmix | mvt; ncov=3 | 0.52 | −1.42 |
| QH | 10 | Mclust | VVV | 0.29 | −1.84 |
| QS | 10 | Mclust | VVV | 0.29 | −1.84 |
| CH | 3 | K-Means | | 0.32 | −1.88 |
| ASW | 2 | Emmix | mvt; ncov=5 | 0.39 | −1.28 |
| FW | 2 | Emmix | mvt; ncov=3 | **0.82** | **−0.42** |
| CVLK | 10 | Rimle | $\gamma = 1000$; InitClust | 0.3 | −1.81 |
| CVQH | 7 | Emmix | mvt; ncov=3 | 0.28 | −2.04 |
| CVQS | 7 | Emmix | mvt; ncov=3 | 0.28 | −2.04 |
| BQH | 8 | Mclust | VVV | 0.49 | −1.28 |
| BQS | 8 | Mclust | VVV | 0.49 | −1.28 |

(g) Olive data set (9 classes, $B = 100$)

| Criterion | K | Algorithm | Configuration | ARI | VIC |
|---|---|---|---|---|---|
| BEST ARI | 8 | Mclust | EVE | 0.88 | −0.65 |
| BEST VIC | 8 | Mclust | EVE | 0.88 | −0.65 |
| AIC | 10 | Emmix | mvt; ncov=3 | 0.47 | −1.77 |
| BIC | 6 | Emmix | mvt; ncov=3 | 0.76 | −1.32 |
| ICL | 6 | Emmix | mvt; ncov=3 | 0.76 | −1.32 |
| QH | 10 | Mclust | VVV | 0.54 | −1.26 |
| QS | 10 | Mclust | VVV | 0.54 | −1.26 |
| CH | 3 | K-Means | | 0.42 | −2.28 |
| ASW | 2 | Emmix | mvt; ncov=5 | 0.29 | −2.28 |
| FW | 2 | Emmix | mvt; ncov=3 | 0.36 | −1.84 |
| CVLK | 10 | Rimle | $\gamma = 1000$; InitClust | 0.58 | −1.18 |
| CVQH | 7 | Emmix | mvt; ncov=3 | 0.44 | −1.96 |
| CVQS | 7 | Emmix | mvt; ncov=3 | 0.44 | −1.96 |
| BQH | 8 | Emmix | mvt; ncov=4 | 0.62 | −1.28 |
| BQS | 8 | Mclust | VVV | **0.86** | **−0.74** |

(h) Olive dataset (9 classes, $B = 1000$)

| Criterion | K | Algorithm | Configuration | ARI | VIC |
|---|---|---|---|---|---|
| BEST ARI | 8 | Mclust | EVE | 0.88 | −0.65 |
| BEST VIC | 8 | Mclust | EVE | 0.88 | −0.65 |
| AIC | 10 | Emmix | mvt; ncov=3 | 0.47 | −1.77 |
| BIC | 6 | Emmix | mvt; ncov=3 | 0.76 | −1.32 |
| ICL | 6 | Emmix | mvt; ncov=3 | 0.76 | −1.32 |
| QH | 10 | Mclust | VVV | 0.54 | −1.26 |
| QS | 10 | Mclust | VVV | 0.54 | −1.26 |
| CH | 3 | K-Means | | 0.42 | −2.28 |
| ASW | 2 | Emmix | mvt; ncov=5 | 0.29 | −2.28 |
| FW | 2 | Emmix | mvt; ncov=3 | 0.36 | −1.84 |
| CVLK | 10 | Rimle | $\gamma = 1000$; InitClust | 0.58 | −1.18 |
| CVQH | 7 | Emmix | mvt; ncov=3 | 0.44 | −1.96 |
| CVQS | 7 | Emmix | mvt; ncov=3 | 0.44 | −1.96 |
| BQH | 8 | Mclust | VVV | **0.86** | **−0.74** |
| BQS | 8 | Mclust | VVV | **0.86** | **−0.74** |

(i) Wine data set ($B = 100$)

| Criterion | K | Algorithm | Configuration | ARI | VIC |
|---|---|---|---|---|---|
| BEST ARI | 3 | Mclust | EEE | 0.98 | −0.08 |
| BEST VIC | 3 | Mclust | EEE | 0.98 | −0.08 |
| AIC | 3 | Emmix | mvt; ncov=3 | 0.44 | −1.42 |
| BIC | 3 | Emmix | mst; ncov=4 | 0.84 | −0.58 |
| ICL | 3 | Emmix | mst; ncov=4 | 0.84 | −0.58 |
| QH | 8 | Emmix | mvt; ncov=4 | 0.46 | −1.53 |
| QS | 8 | Emmix | mvt; ncov=4 | 0.46 | −1.53 |
| CH | 10 | Rimle | $\gamma = 100$; kmeans | 0.15 | −3.15 |
| ASW | 2 | Rimle | $\gamma = 100$; kmeans | 0.37 | −1.41 |
| FW | 3 | Rimle | $\gamma = 1000$; InitClust | 0.36 | −1.8 |
| CVLK | 6 | Mclust | EVI | 0.59 | −1.37 |
| CVQH | 5 | Mclust | EEI | 0.64 | −1.11 |
| CVQS | 5 | Mclust | EEI | 0.64 | −1.11 |
| BQH | 3 | Emmix | mvt; ncov=2 | **0.9** | **−0.38** |
| BQS | 3 | Emmix | mvt; ncov=2 | **0.9** | **−0.38** |

(j) Wine data set ($B = 1000$)

| Criterion | K | Algorithm | Configuration | ARI | VIC |
|---|---|---|---|---|---|
| BEST ARI | 3 | Mclust | EEE | 0.98 | −0.08 |
| BEST VIC | 3 | Mclust | EEE | 0.98 | −0.08 |
| AIC | 3 | Emmix | mvt; ncov=3 | 0.44 | −1.42 |
| BIC | 3 | Emmix | mst; ncov=4 | 0.84 | −0.58 |
| ICL | 3 | Emmix | mst; ncov=4 | 0.84 | −0.58 |
| QH | 8 | Emmix | mvt; ncov=4 | 0.46 | −1.53 |
| QS | 8 | Emmix | mvt; ncov=4 | 0.46 | −1.53 |
| CH | 10 | Rimle | $\gamma = 100$; kmeans | 0.15 | −3.15 |
| ASW | 2 | Rimle | $\gamma = 100$; kmeans | 0.37 | −1.41 |
| FW | 3 | Mclust | VEI | 0.87 | −0.48 |
| CVLK | 6 | Mclust | EVI | 0.59 | −1.37 |
| CVQH | 5 | Mclust | EEI | 0.64 | −1.11 |
| CVQS | 5 | Mclust | EEI | 0.64 | −1.11 |
| BQH | 3 | Emmix | mvt; ncov=2 | **0.9** | **−0.38** |
| BQS | 3 | Emmix | mvt; ncov=2 | **0.9** | **−0.38** |

Figure S8: In-sample and bootstrap versions of the Hard Scoring (top plots in each panel) and Smooth Scoring (bottom plots in each panel); bootstrap results are shown for 100 and 1000 bootstrap resamples. In each plot: QH and QS criteria (dashed line); estimates $\widetilde{W}_n$ (solid line). Bootstrap confidence bounds (at 95%-level) around $\widetilde{W}_n$ are represented by shaded areas; lower confidence bounds correspond to BQH and BQS criteria.

(a) Iris data

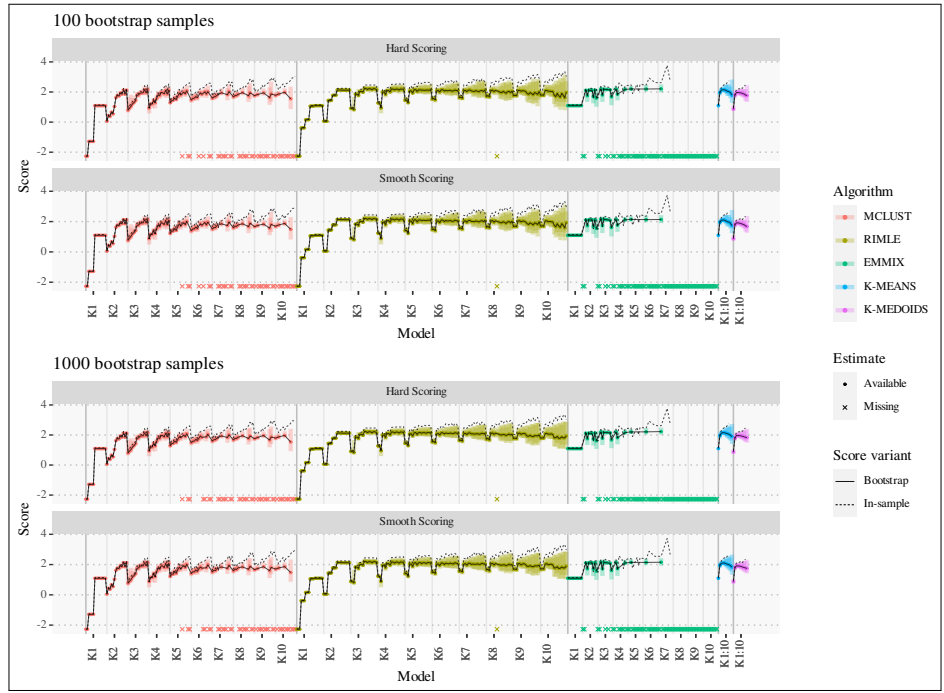

(b) Banknote data

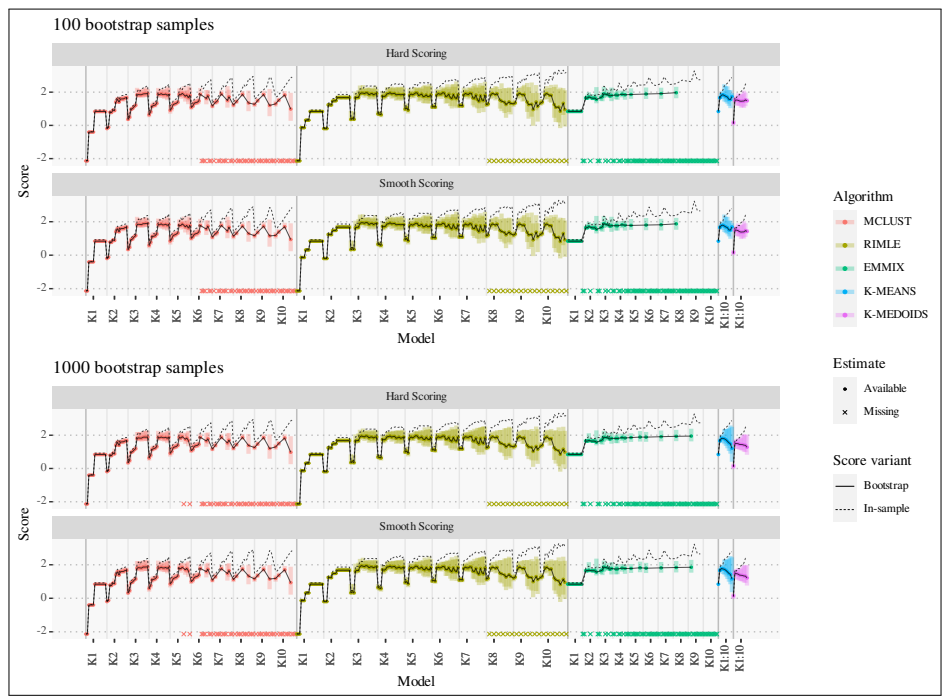

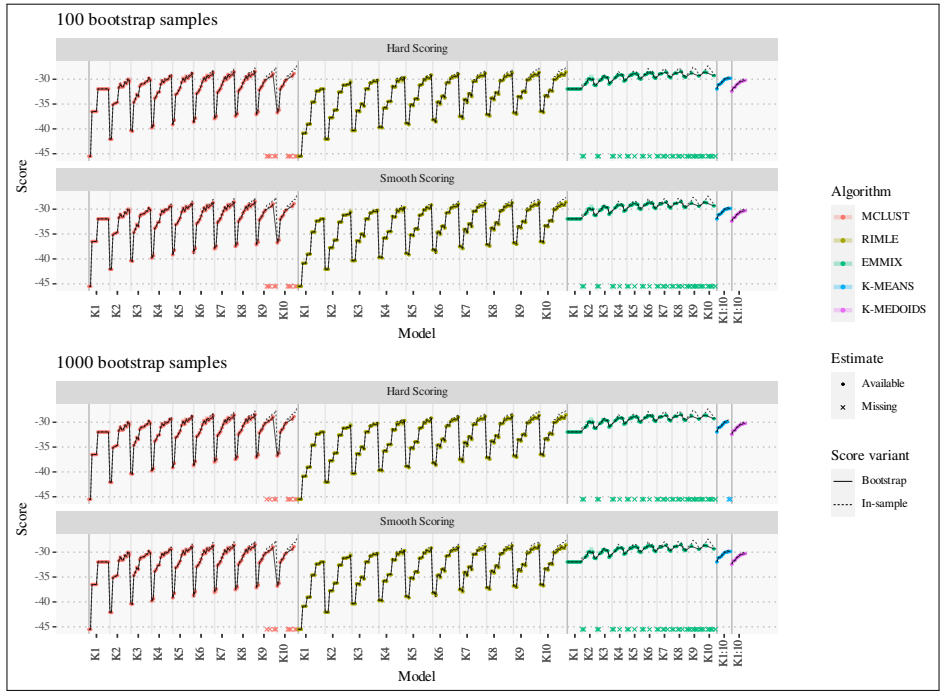
(c) Olive data

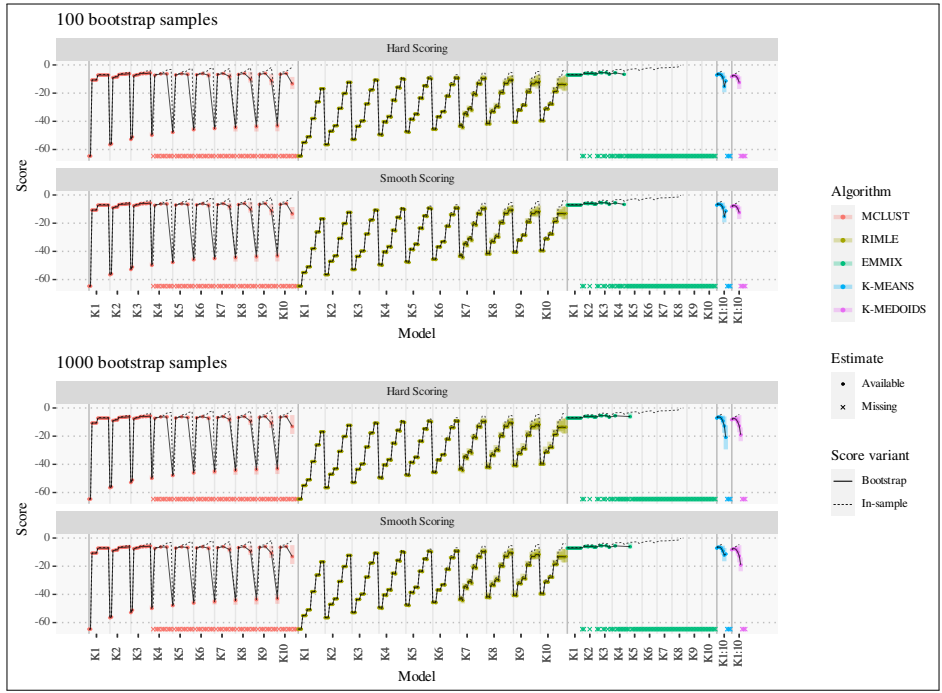
(d) Wine data

Table S5: Summary results for the Monte Carlo experiments. Mean and median ARI and VIC are reported (standard errors in parentheses). Last two columns indicate the most and second most frequently selected number of clusters (frequency in parentheses). First two rows show the best feasible solutions with respect to the ARI and the VIC criteria (BEST ARI and BEST VIC).

(a) Pentagon5

| Criterion | Mean ARI | Mean VIC | Med. ARI | Med. VIC | 1st K | 2nd K |
|---|---|---|---|---|---|---|
| BEST ARI | $0.92 \pm (0.02)$ | | 0.92 | | 5 (0.83) | 4 (0.1) |
| BEST VIC | | $-0.36 \pm (0.07)$ | | $-0.36$ | 4 (0.52) | 5 (0.34) |
| AIC | $0.82 \pm (0.11)$ | $-0.66 \pm (0.31)$ | 0.87 | $-0.57$ | 5 (0.59) | 6 (0.11) |
| BIC | $\mathbf{0.88 \pm (0.04)}$ | $-0.43 \pm (0.09)$ | $\mathbf{0.89}$ | $-0.43$ | 5 (0.45) | 4 (0.3) |
| ICL | $0.85 \pm (0.03)$ | $\mathbf{-0.42 \pm (0.06)}$ | 0.84 | $\mathbf{-0.43}$ | 3 (0.88) | 4 (0.08) |
| QH | $0.76 \pm (0.12)$ | $-0.89 \pm (0.36)$ | 0.80 | $-0.83$ | 10 (0.31) | 7 (0.16) |
| QS | $0.84 \pm (0.07)$ | $-0.57 \pm (0.25)$ | 0.86 | $-0.48$ | 4 (0.39) | 3 (0.16) |
| CH | $0.84 \pm (0.02)$ | $-0.43 \pm (0.05)$ | 0.84 | $-0.44$ | 3 (1) | |
| ASW | $0.84 \pm (0.03)$ | $-0.43 \pm (0.06)$ | 0.84 | $\mathbf{-0.43}$ | 3 (1) | 4 (0) |
| FW | $0.84 \pm (0.06)$ | $-0.43 \pm (0.1)$ | 0.84 | $\mathbf{-0.43}$ | 3 (0.99) | 2 (0.01) |
| CVLK | $0.73 \pm (0.13)$ | $-0.85 \pm (0.33)$ | 0.75 | $-0.82$ | 6 (0.32) | 5 (0.23) |
| CVQH | $0.82 \pm (0.09)$ | $-0.58 \pm (0.25)$ | 0.85 | $-0.50$ | 4 (0.42) | 3 (0.32) |
| CVQS | $0.84 \pm (0.05)$ | $-0.48 \pm (0.13)$ | 0.85 | $-0.45$ | 3 (0.61) | 4 (0.31) |
| BQH | $0.85 \pm (0.03)$ | $-0.43 \pm (0.06)$ | 0.84 | $-0.44$ | 3 (0.9) | 4 (0.08) |
| BQS | $0.84 \pm (0.03)$ | $-0.43 \pm (0.06)$ | 0.84 | $-0.44$ | 3 (0.98) | 4 (0.02) |

(b) T52D

| Criterion | Mean ARI | Mean VIC | Med. ARI | Med. VIC | 1st K | 2nd K |
|---|---|---|---|---|---|---|
| BEST ARI | $0.99 \pm (0.01)$ | | 0.99 | | 5 (0.99) | 4 (0.01) |
| BEST VIC | | $-0.06 \pm (0.05)$ | | $-0.05$ | 5 (0.97) | 4 (0.01) |
| AIC | $0.84 \pm (0.13)$ | $-0.5 \pm (0.32)$ | 0.86 | $-0.48$ | 6 (0.21) | 9 (0.2) |
| BIC | $0.97 \pm (0.04)$ | $-0.12 \pm (0.1)$ | 0.98 | $-0.10$ | 5 (0.9) | 6 (0.09) |
| ICL | $0.98 \pm (0.01)$ | $-0.11 \pm (0.08)$ | $\mathbf{0.99}$ | $-0.09$ | 5 (0.95) | 4 (0.04) |
| QH | $0.85 \pm (0.1)$ | $-0.55 \pm (0.26)$ | 0.86 | $-0.58$ | 10 (0.41) | 9 (0.19) |
| QS | $0.91 \pm (0.08)$ | $-0.35 \pm (0.25)$ | 0.93 | $-0.29$ | 10 (0.21) | 5 (0.2) |
| CH | $0.7 \pm (0.1)$ | $-0.7 \pm (0.26)$ | 0.68 | $-0.71$ | 7 (0.4) | 6 (0.31) |
| ASW | $0.92 \pm (0.15)$ | $-0.26 \pm (0.32)$ | 0.97 | $-0.16$ | 5 (0.86) | 2 (0.13) |
| FW | $0.59 \pm (0.18)$ | $-0.93 \pm (0.36)$ | 0.51 | $-1.08$ | 2 (0.81) | 4 (0.13) |
| CVLK | $0.84 \pm (0.13)$ | $-0.44 \pm (0.27)$ | 0.89 | $-0.34$ | 6 (0.41) | 7 (0.24) |
| CVQH | $0.9 \pm (0.11)$ | $-0.32 \pm (0.23)$ | 0.95 | $-0.26$ | 4 (0.43) | 5 (0.31) |
| CVQS | $0.93 \pm (0.08)$ | $-0.27 \pm (0.19)$ | 0.95 | $-0.24$ | 4 (0.54) | 5 (0.31) |
| BQH | $\mathbf{0.99 \pm (0.01)}$ | $\mathbf{-0.08 \pm (0.06)}$ | $\mathbf{0.99}$ | $\mathbf{-0.07}$ | 5 (0.98) | 4 (0.01) |
| BQS | $\mathbf{0.99 \pm (0.01)}$ | $\mathbf{-0.08 \pm (0.06)}$ | $\mathbf{0.99}$ | $\mathbf{-0.07}$ | 5 (0.98) | 4 (0.01) |

## S5.3 Results on simulated data

In this section, we present more detailed results on simulated data. Table S5 reports the average ARI and VIC ± one Monte Carlo standard error in parentheses. We also report median values and the most and second most frequently selected number of clusters with their frequencies in parentheses. Note that we report the negative of VIC so that higher values correspond to better clusterings. The first two rows show the same results for the BEST ARI and BEST VIC selected solutions. The best performances, excluding that of BEST ARI and BEST VIC, are highlighted in bold.

Figure S8 shows frequencies plots, on the left, for the selected number of clusters $K = 1, 2, \ldots, 10$: in each MC replicate, each criterion selects the optimal clustering solution, of which we track the number of clusters; frequencies across the 100 MC are shown in this plot. On the right, the figure shows boxplots of the MC distribution of the ARI and the VIC (top and bottom panel of each subplot) for the selected solutions, for all the criteria. The dashed line in each plot indicates the MC average best feasible ARI and VIC available in $\mathcal{M}$ (achieved by the BEST ARI and BEST VIC solutions), together with bands (shaded area) ranging from the first to the third quartiles of the indexes' empirical distributions.

Table S5

(c) T510D

| Criterion | Mean ARI | Mean VIC | Med. ARI | Med. VIC | 1st K | 2nd K |
|---|---|---|---|---|---|---|
| BEST ARI | $0.99 \pm (0.01)$ | | 0.99 | | 5 (0.99) | 4 (0.01) |
| BEST VIC | | $-0.09 \pm (0.06)$ | | $-0.09$ | 5 (0.99) | 4 (0.01) |
| AIC | $0.7 \pm (0.13)$ | $-1.06 \pm (0.43)$ | 0.68 | $-1.03$ | 9 (0.24) | 10 (0.24) |
| BIC | $0.86 \pm (0.1)$ | $-0.35 \pm (0.18)$ | 0.82 | $-0.41$ | 6 (0.5) | 5 (0.49) |
| ICL | $0.94 \pm (0.08)$ | $-0.23 \pm (0.15)$ | 0.97 | $-0.19$ | 5 (0.83) | 6 (0.16) |
| QH | $0.55 \pm (0.08)$ | $-1.23 \pm (0.32)$ | 0.53 | $-1.20$ | 10 (0.98) | 9 (0.02) |
| QS | $0.55 \pm (0.07)$ | $-1.2 \pm (0.23)$ | 0.53 | $-1.19$ | 10 (0.98) | 9 (0.02) |
| CH | $0.51 \pm (0.03)$ | $-1.1 \pm (0.05)$ | 0.50 | $-1.11$ | 2 (1) | |
| ASW | $0.51 \pm (0.03)$ | $-1.1 \pm (0.05)$ | 0.51 | $-1.11$ | 2 (1) | |
| FW | $0.53 \pm (0.11)$ | $-1.05 \pm (0.21)$ | 0.50 | $-1.10$ | 2 (0.94) | 5 (0.04) |
| CVLK | $0.74 \pm (0.13)$ | $-0.75 \pm (0.43)$ | 0.74 | $-0.67$ | 6 (0.45) | 7 (0.38) |
| CVQH | $0.79 \pm (0.13)$ | $-0.62 \pm (0.33)$ | 0.79 | $-0.58$ | 5 (0.36) | 6 (0.35) |
| CVQS | $0.8 \pm (0.13)$ | $-0.61 \pm (0.34)$ | 0.79 | $-0.58$ | 5 (0.38) | 6 (0.29) |
| BQH | $0.91 \pm (0.12)$ | $-0.28 \pm (0.26)$ | 0.97 | $-0.19$ | 5 (0.69) | 6 (0.19) |
| BQS | $0.94 \pm (0.09)$ | $-0.2 \pm (0.18)$ | 0.98 | $-0.15$ | 5 (0.85) | 6 (0.1) |

(d) Flower2

| Criterion | Mean ARI | Mean VIC | Med. ARI | Med. VIC | 1st K | 2nd K |
|---|---|---|---|---|---|---|
| BEST ARI | $0.68 \pm (0.06)$ | | 0.69 | | 5 (0.73) | 6 (0.11) |
| BEST VIC | | $-1.21 \pm (0.17)$ | | $-1.21$ | 5 (0.77) | 4 (0.18) |
| AIC | $0.48 \pm (0.1)$ | $-1.91 \pm (0.34)$ | 0.48 | $-1.89$ | 8 (0.24) | 9 (0.22) |
| BIC | $0.32 \pm (0.1)$ | $-1.88 \pm (0.22)$ | 0.29 | $-1.92$ | 2 (0.58) | 3 (0.33) |
| ICL | $0.35 \pm (0.12)$ | $-1.8 \pm (0.26)$ | 0.30 | $-1.90$ | 2 (0.65) | 3 (0.2) |
| QH | $0.47 \pm (0.07)$ | $-1.96 \pm (0.21)$ | 0.47 | $-1.97$ | 10 (0.85) | 9 (0.13) |
| QS | $0.46 \pm (0.08)$ | $-1.88 \pm (0.22)$ | 0.46 | $-1.89$ | 10 (0.59) | 9 (0.21) |
| CH | $0.44 \pm (0.04)$ | $-1.98 \pm (0.17)$ | 0.44 | $-1.98$ | 10 (0.87) | 9 (0.08) |
| ASW | $0.45 \pm (0.17)$ | $-1.58 \pm (0.29)$ | 0.50 | $-1.50$ | 5 (0.85) | 2 (0.1) |
| FW | $0.45 \pm (0.1)$ | $-1.6 \pm (0.21)$ | 0.46 | $-1.60$ | 5 (0.86) | 2 (0.09) |
| CVLK | $0.49 \pm (0.11)$ | $-1.8 \pm (0.27)$ | 0.51 | $-1.79$ | 7 (0.43) | 8 (0.23) |
| CVQH | $0.47 \pm (0.13)$ | $-1.73 \pm (0.27)$ | 0.48 | $-1.73$ | 6 (0.27) | 7 (0.27) |
| CVQS | $0.43 \pm (0.14)$ | $-1.74 \pm (0.27)$ | 0.46 | $-1.80$ | 5 (0.23) | 6 (0.22) |
| BQH | $0.53 \pm (0.09)$ | $-1.51 \pm (0.25)$ | 0.52 | $-1.49$ | 5 (0.74) | 9 (0.08) |
| BQS | $0.46 \pm (0.11)$ | $-1.58 \pm (0.24)$ | 0.49 | $-1.52$ | 5 (0.72) | 2 (0.23) |

(e) Uniform

| Criterion | Mean ARI | Mean VIC | Med. ARI | Med. VIC | 1st K | 2nd K |
|---|---|---|---|---|---|---|
| BEST ARI | $1 \pm (0)$ | | 1 | | 1 (1) | |
| BEST VIC | | $0 \pm (0)$ | | 0 | 1 (1) | |
| AIC | $0 \pm (0)$ | $-3.01 \pm (0.22)$ | 0 | $-3.03$ | 10 (0.51) | 9 (0.27) |
| BIC | $0 \pm (0)$ | $-1.94 \pm (0.4)$ | 0 | $-1.98$ | 4 (0.65) | 5 (0.13) |
| ICL | $0.77 \pm (0.42)$ | $-0.22 \pm (0.45)$ | 1 | 0.00 | 1 (0.77) | 2 (0.18) |
| QH | $0 \pm (0)$ | $-3.1 \pm (0.11)$ | 0 | $-3.12$ | 10 (0.9) | 9 (0.08) |
| QS | $0.16 \pm (0.37)$ | $-2.61 \pm (1.14)$ | 0 | $-3.10$ | 10 (0.71) | 1 (0.16) |
| CH | $0 \pm (0)$ | $-3.06 \pm (0.41)$ | 0 | $-3.16$ | 10 (0.46) | 9 (0.38) |
| ASW | $0 \pm (0)$ | $-2.1 \pm (0.37)$ | 0 | $-1.99$ | 4 (0.74) | 3 (0.07) |
| FW | $0 \pm (0)$ | $-2.38 \pm (0.59)$ | 0 | $-1.99$ | 4 (0.64) | 10 (0.18) |
| CVLK | $0 \pm (0)$ | $-2.84 \pm (0.28)$ | 0 | $-2.84$ | 8 (0.3) | 9 (0.24) |
| CVQH | $0.06 \pm (0.24)$ | $-2.3 \pm (0.81)$ | 0 | $-2.55$ | 7 (0.22) | 6 (0.19) |
| CVQS | $0.85 \pm (0.35)$ | $-0.25 \pm (0.73)$ | 1 | 0.00 | 1 (0.85) | 2 (0.07) |
| BQH | $0 \pm (0)$ | $-3.14 \pm (0.31)$ | 0 | $-3.21$ | 10 (0.82) | 9 (0.1) |
| BQS | $0.96 \pm (0.19)$ | $-0.1 \pm (0.54)$ | 1 | 0.00 | 1 (0.96) | 10 (0.03) |

Figure S8: Results on artificial data sets. In each panel: (left) frequencies for the selected number of clusters for $K = 1, 2, \ldots 10$; (right) boxplots of the Monte Carlo distribution of ARI and BIC.

(a) Pentagon5

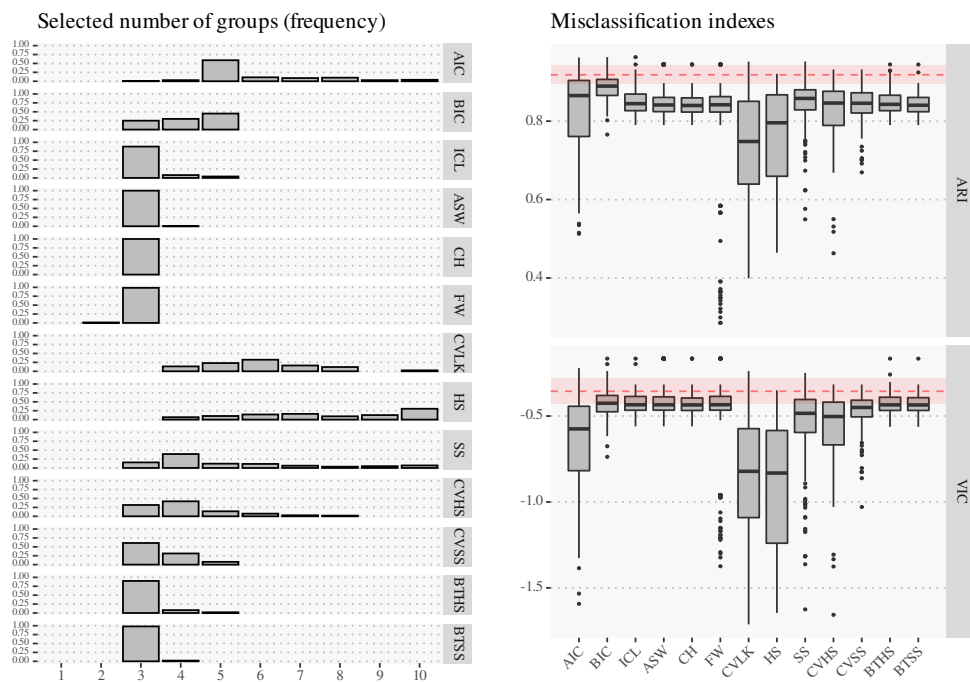

(b) T52D

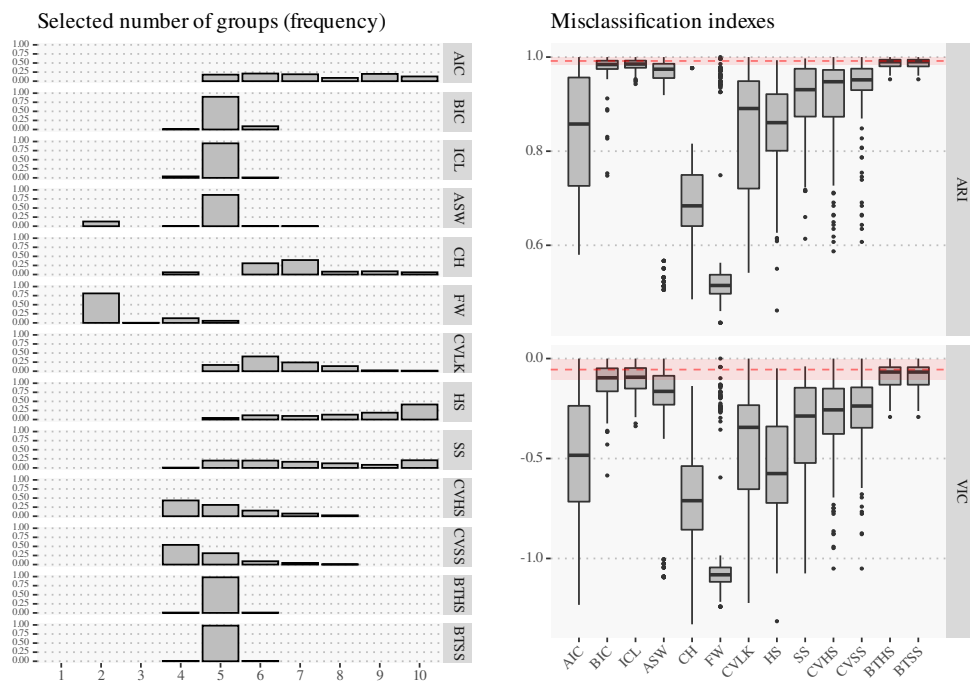

Figure S8

(c) T510D

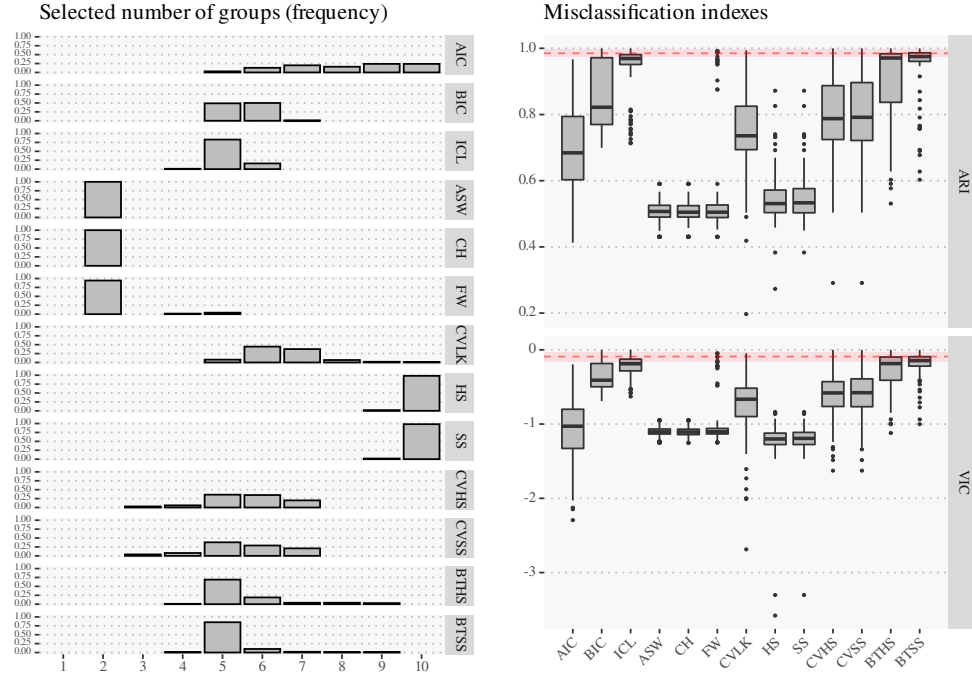

(d) Flower2

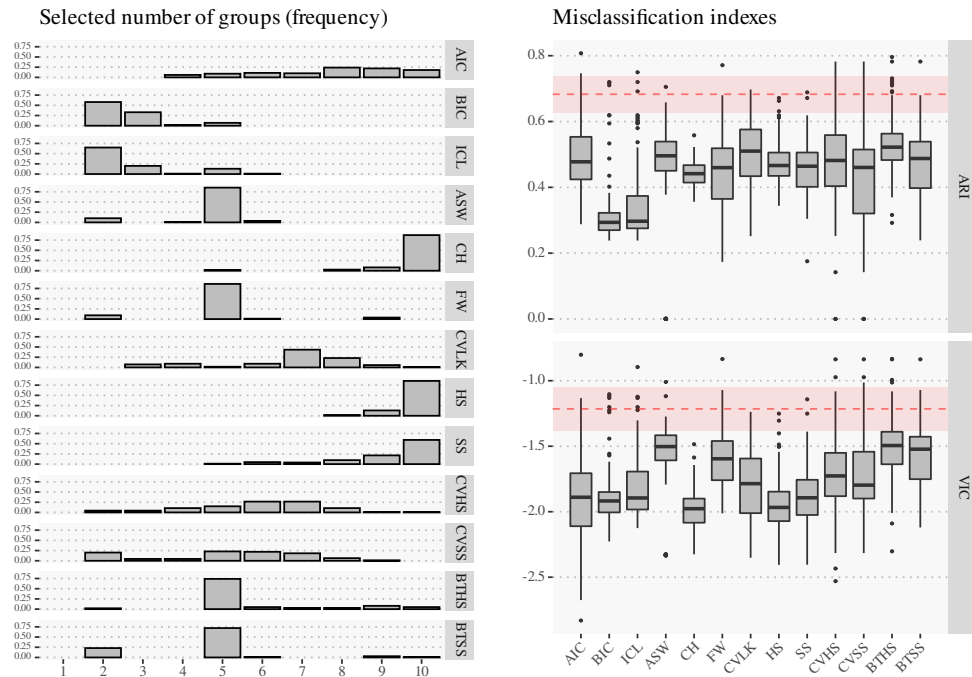

31

Figure S8

(e) Uniform

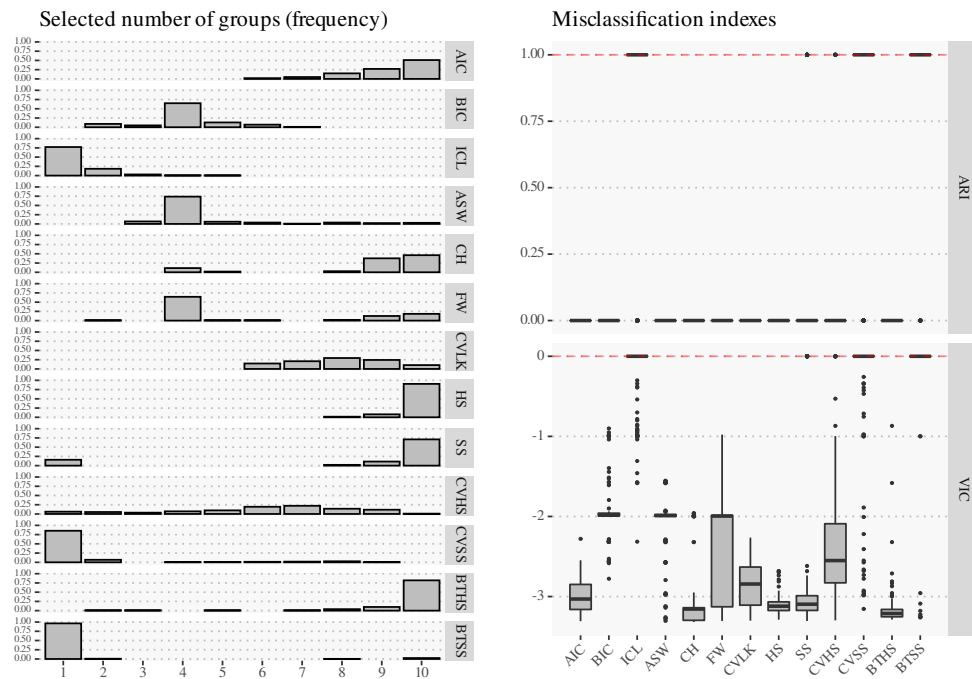



\end{document}